\documentclass[letterpaper,10pt,journal]{IEEEtran}

\usepackage[percent]{overpic}
\usepackage{cite}
\usepackage{amsmath,amssymb,amsfonts}
\usepackage{algorithm}
\usepackage{algpseudocode}
\usepackage{graphicx}
\usepackage{booktabs}
\usepackage{multirow}
\usepackage{makecell}
\usepackage[normalem]{ulem}
\usepackage{float}
\usepackage{array}
\usepackage{adjustbox}
\usepackage{stfloats}
\usepackage{xcolor}
\usepackage{colortbl}
\usepackage{microtype}
\usepackage{capt-of} 
\usepackage{xspace}
\usepackage{url}
\usepackage{tikz}
\usepackage[hidelinks]{hyperref}

\long\def\fsl#1{{\color{black}#1}}

\newcommand{\eg}{\textit{e.g.}}
\newcommand{\ie}{\textit{i.e.}}
\newcommand{\etal}{\textit{et al.}}

\makeatletter
\let\BiFreqOriginalMakeCaption\@makecaption
\newif\ifBiFreqResultCaption
\BiFreqResultCaptionfalse

\long\def\@makecaption#1#2{%
  \ifBiFreqResultCaption
    \begingroup
      \let\BiFreqOriginalNormalFont\normalfont
      \def\normalfont{%
        \BiFreqOriginalNormalFont
        \fontfamily{ptm}\selectfont}%
      \BiFreqOriginalMakeCaption{#1}{#2}%
    \endgroup
  \else
    \BiFreqOriginalMakeCaption{#1}{#2}%
  \fi
}

\newcommand{\resultcaption}[1]{%
  \BiFreqResultCaptiontrue
  \caption{#1}%
  \BiFreqResultCaptionfalse
}
\makeatother
\begin{document}

\title{Bidirectional Cross-frequency Domain Interaction Network for UHD Low-Light Image Enhancement}

\author{Bo~Fu, Hai~Chi, Liyan~Wang, Shilin~Fu, Wei~Wang, Kin-Man Lam, Jinshan~Pan, and Cong~Wang %

\thanks{Bo Fu and Hai Chi are with the School of Computer and Artificial Intelligence, Liaoning Normal University, Dalian 116081, China (e-mail: fubo@lnnu.edu.cn; davidchi226@163.com).}%

\thanks{Liyan Wang is with the School of Mathematical Sciences, Dalian University of Technology, Dalian 116024, China (e-mail: wangliyanlee@163.com). Liyan Wang is the corresponding author.}%

\thanks{Shilin Fu is with the School of Computer Sciences, Dalian University of Technology, Dalian 116024, China (e-mail: fslecho@gmail.com).}%

\thanks{Kin-Man Lam and Cong Wang are with the Hong Kong Polytechnic University, Hong Kong (e-mail: kin.man.lam@polyu.edu.hk; supercong94@gmail.com).}%

\thanks{Wei Wang is with the School of Cyber Science and Technology, Shenzhen Campus of Sun Yat-sen University, Shenzhen, China (e-mail: wangwei29@mail.sysu.edu.cn).}%

\thanks{Jinshan Pan is with the School of Computer Science and Engineering, Nanjing University of Science and Technology, Nanjing 210094, China (e-mail: sdluran@gmail.com).}%

}
\maketitle

\begin{abstract}

Existing Ultra-High-Definition (UHD) low-light image enhancement (LLIE) methods typically rely on either spatial- or frequency-domain modeling, or independently process different frequency components for efficient restoration. However, they often fail to jointly exploit complementary spatial--frequency information and the inherent relationship between illumination and structural information across frequencies, leading to inconsistent illumination, residual noise, and detail artifacts. In this paper, we propose a novel Bidirectional Cross-frequency Domain Interaction Network (BiFreqNet) for UHD LLIE. Specifically, BiFreqNet incorporates a Dual-domain Feature Interaction (DFI) module into the multi-scale representations of a pretrained variational autoencoder (VAE) to perform reciprocal enhancement between low- and high-frequency features. DFI consists of a Spatial and Spectral-Aware Low-Frequency Enhancement (S$^2$LFE) block and a Cross-Frequency Enhancement (CFE) block. S$^2$LFE exploits high-frequency structural cues to guide low-frequency restoration and jointly models spatial and spectral information to recover global exposure and spatially varying illumination while suppressing sensor noise. CFE further leverages the enhanced low-frequency contextual information to guide adaptive high-frequency refinement, enabling faithful reconstruction of structures and fine details. Extensive experiments on four general and UHD LLIE benchmarks demonstrate that BiFreqNet achieves competitive or superior performance compared with state-of-the-art methods.

\end{abstract}

\section{Introduction}

\IEEEPARstart{U}ltra-high-definition (UHD) images are widely used in professional photography, digital media production, and high-quality video rendering due to their superior perceptual quality and rich fine-scale details. However, UHD images captured under low-light conditions often suffer from illumination degradation, sensor noise, and detail loss~\cite{wang2025deep}, which not only deteriorate visual quality but also hinder downstream vision tasks, such as object recognition~\cite{zhou2026multi,zhou2026incentive}, detection~\cite{MI_DETR}, and tracking~\cite{Liu2024NTVOT}. A straightforward solution is to directly apply existing general low-light image enhancement (LLIE) methods (\eg,~\cite{XuKeCVPR2020, li2021zero, liu2021retinex, SNRNet, jiang2023diffll, cai2023retinexformer, weng2024mamballie, wang2024perceplie,bai2024retinexmamba, he2025retidiff, URWKV,wang2025intra, HVI_CIDNet, zhou2025gppllie, BiBridge,wang2026neural,wang2026ultra,fan2026mambalie}) to restore UHD low-light images. However, these methods are mainly designed for conventional-resolution images and struggle to process UHD inputs efficiently, incurring high computational costs and offering limited capability to restore illumination while preserving fine-grained details. Hence, it is essential to develop an effective UHD LLIE method that achieves both accurate illumination restoration and faithful detail preservation.

Recently, a number of UHD LLIE methods have been proposed. Some of them rely mainly on either spatial- or frequency-domain modeling for efficient UHD LLIE, \eg, Fourier-domain enhancement~\cite{li2023uhdfour}, axis-based Transformer modeling~\cite{LLFormer}, correlation-guided high-/low-resolution interaction~\cite{wang2024uhdformer}, normal- and gradient-prior-guided restoration~\cite{Wang2026uhddip}, and sparse-curve-based look-up table (LUT) enhancement~\cite{wu2026sce}.
\begin{figure*}[!t]
\centering
\setlength{\tabcolsep}{1pt}

\begingroup
\fontsize{10pt}{7pt}\selectfont

\begin{tabular}{@{}cccc@{}}

\includegraphics[width=0.498\columnwidth]{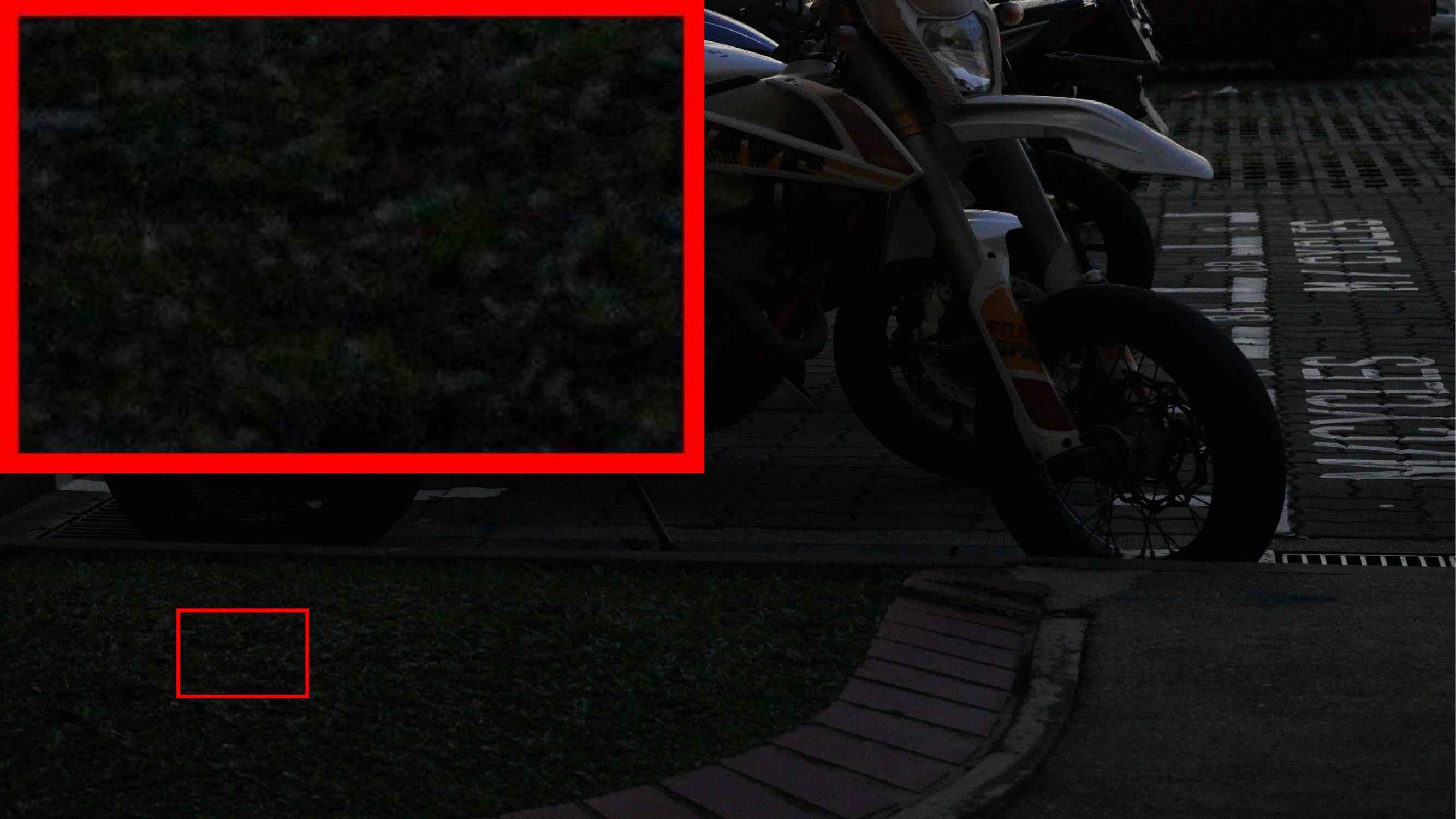} &
\includegraphics[width=0.498\columnwidth]{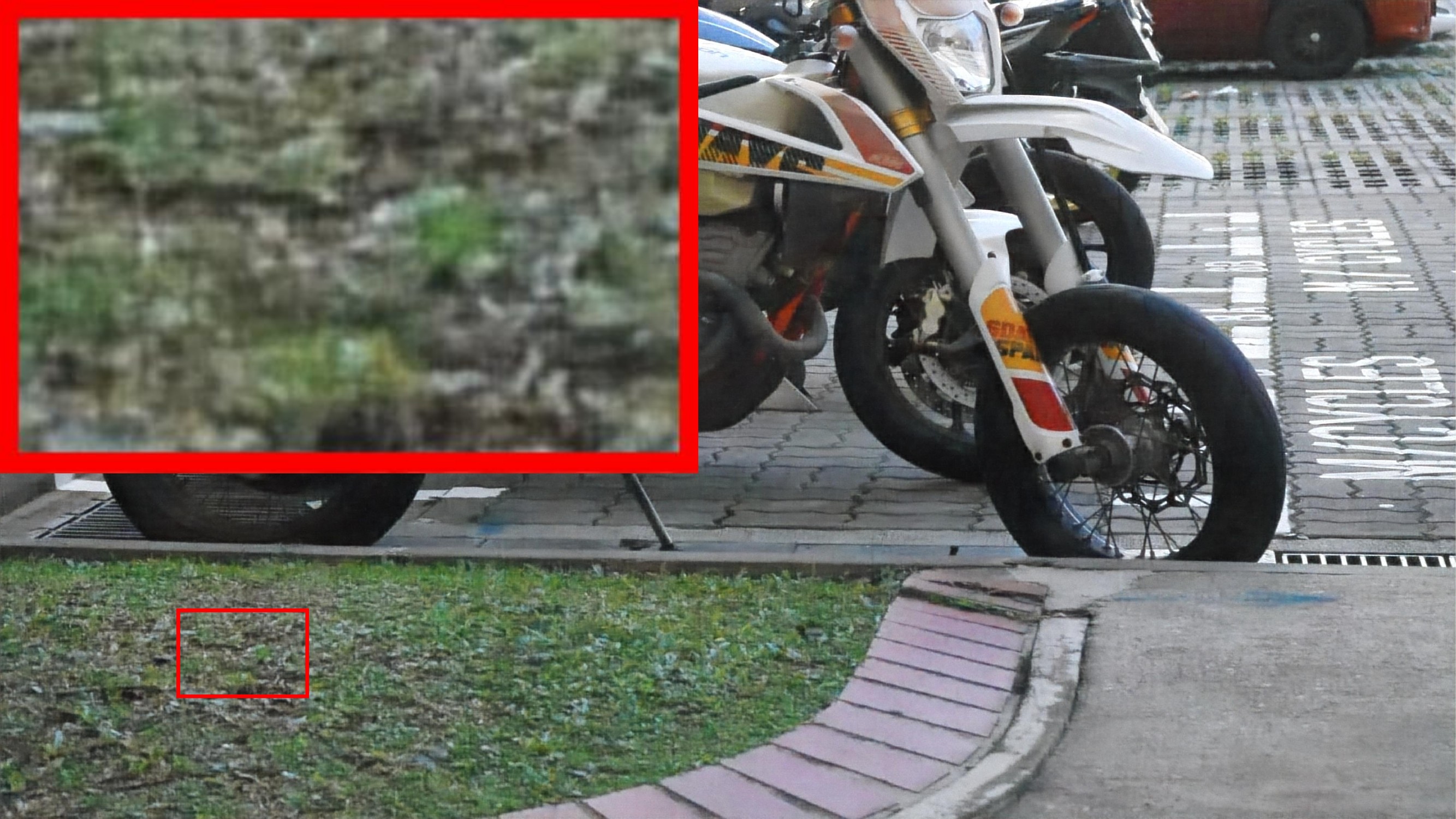} &
\includegraphics[width=0.498\columnwidth]{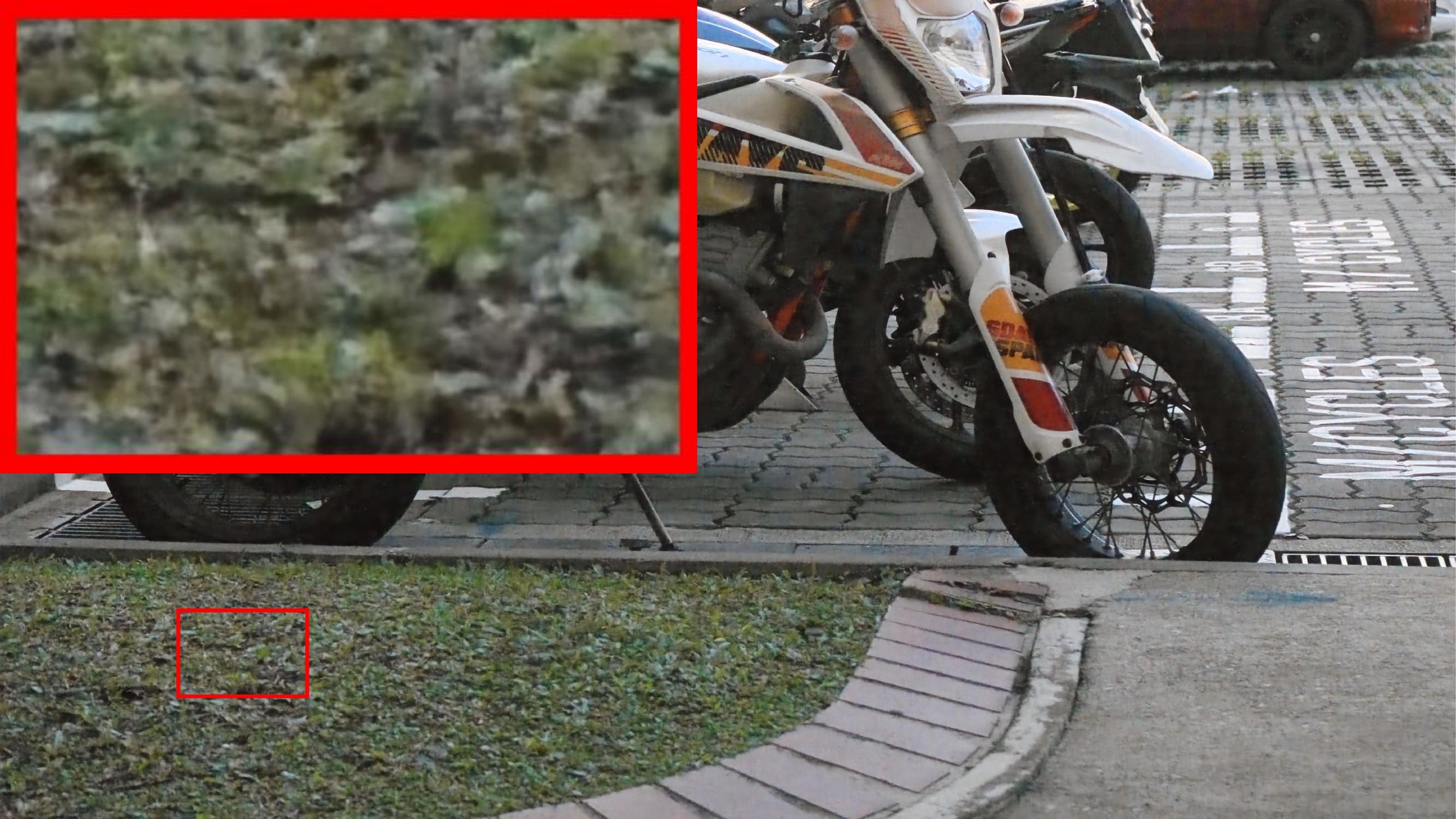} &
\includegraphics[width=0.498\columnwidth]{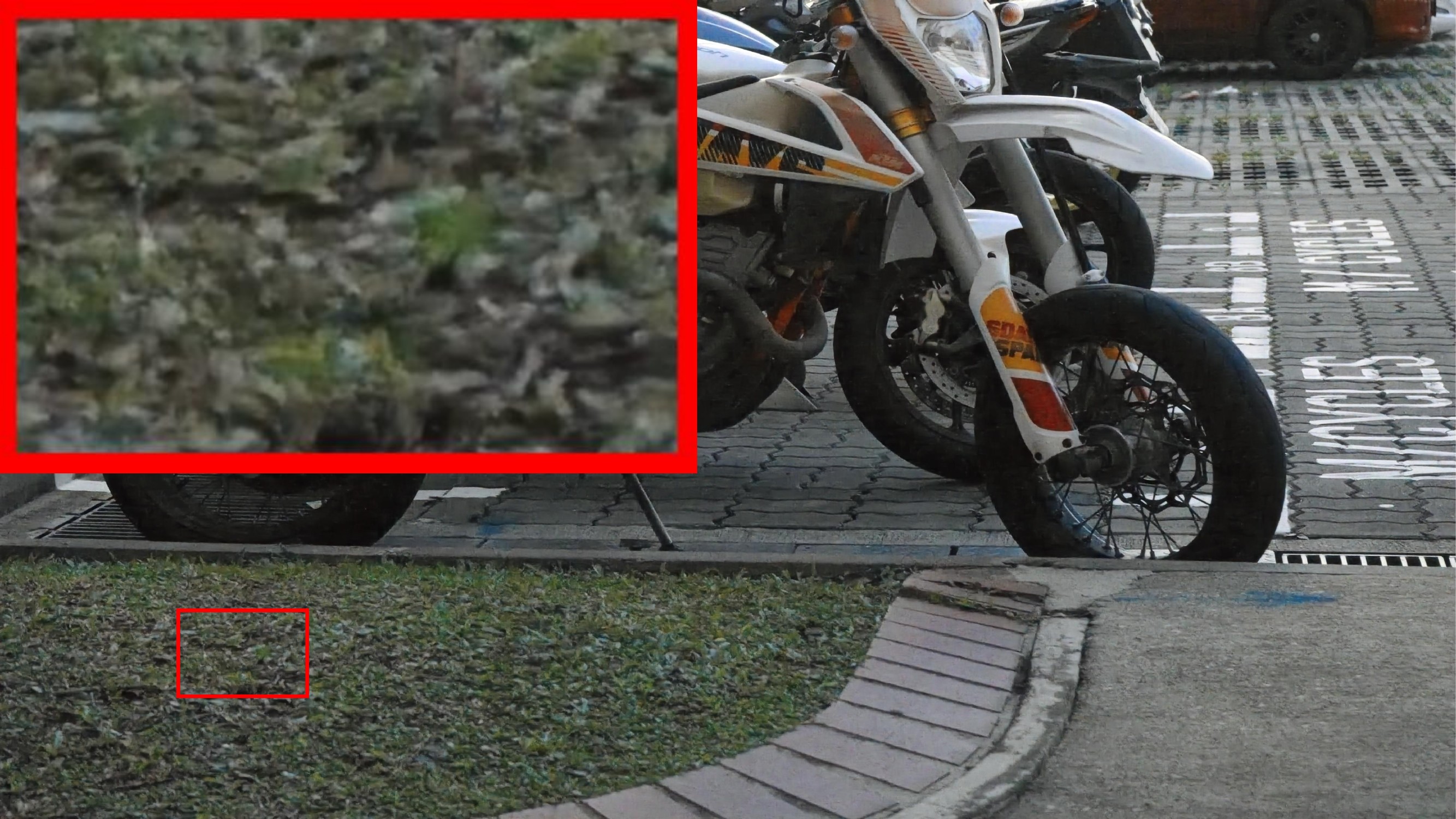}
\\[1.5pt]

\small(a) Input &
\small(b) UHDFour~\cite{li2023uhdfour} &
\small(c) UHDFormer~\cite{wang2024uhdformer} &
\small(d) UHDDIP~\cite{Wang2026uhddip}
\\[1.5pt]

\includegraphics[width=0.498\columnwidth]{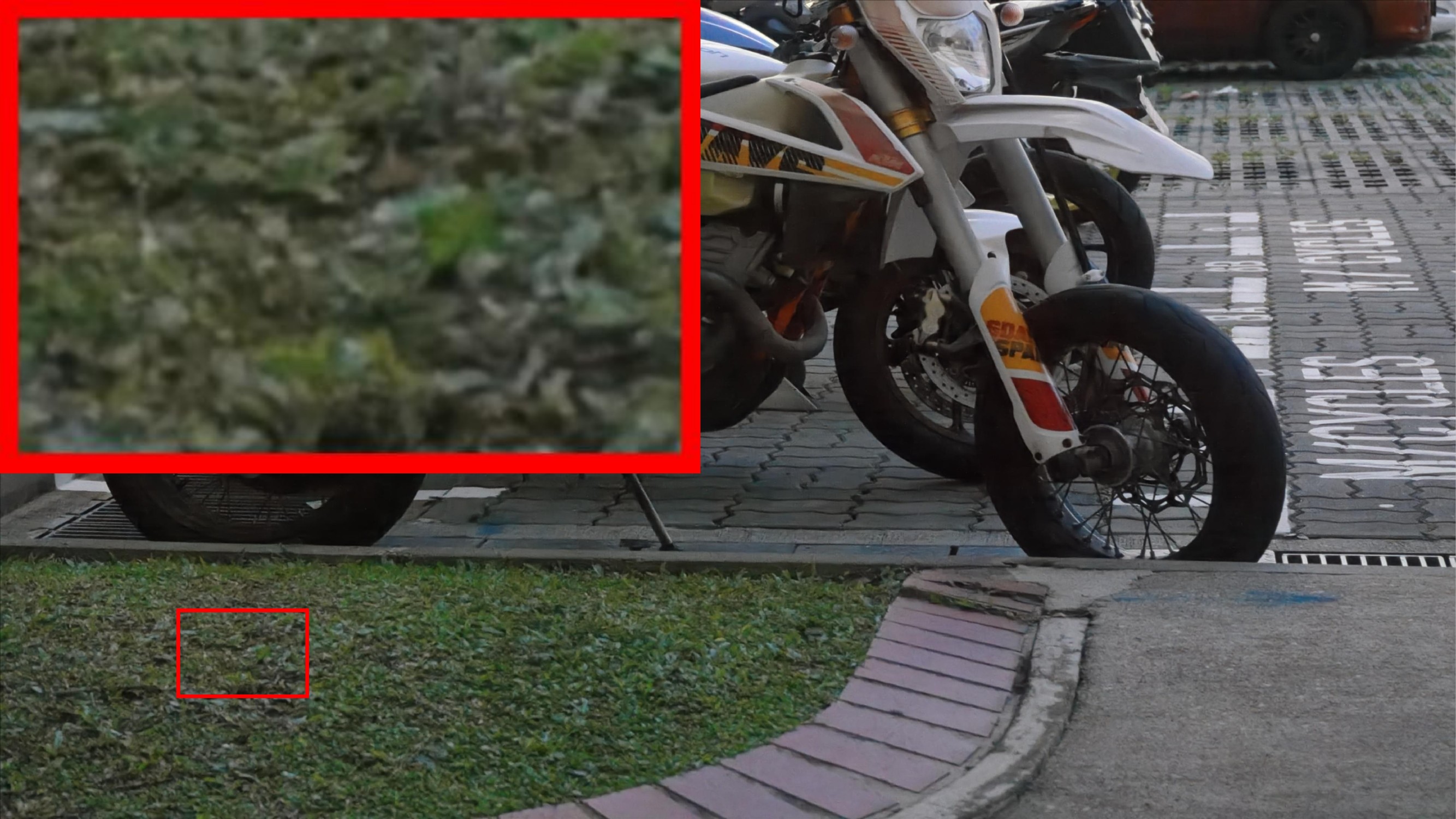} &
\includegraphics[width=0.498\columnwidth]{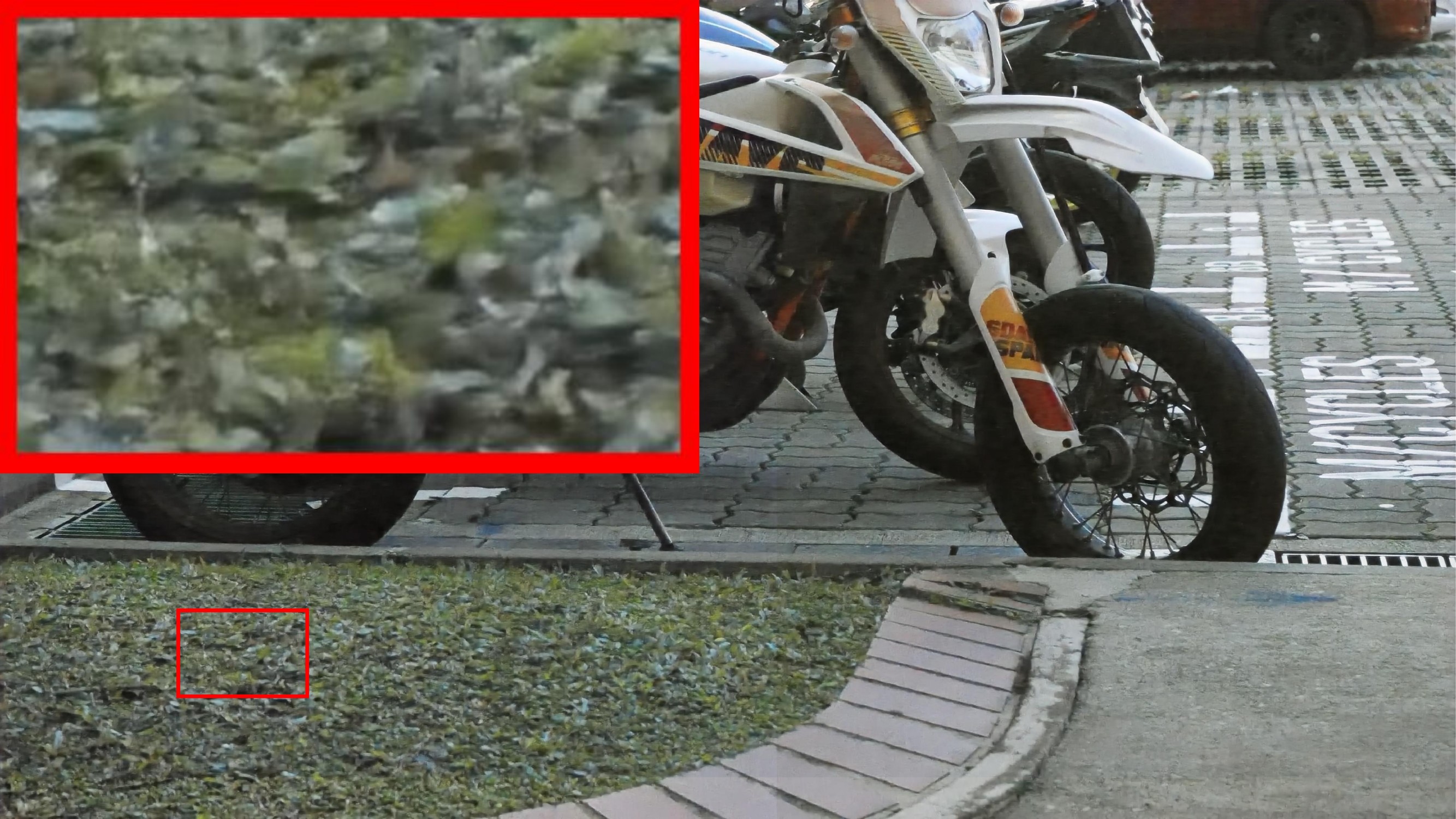} &
\includegraphics[width=0.498\columnwidth]{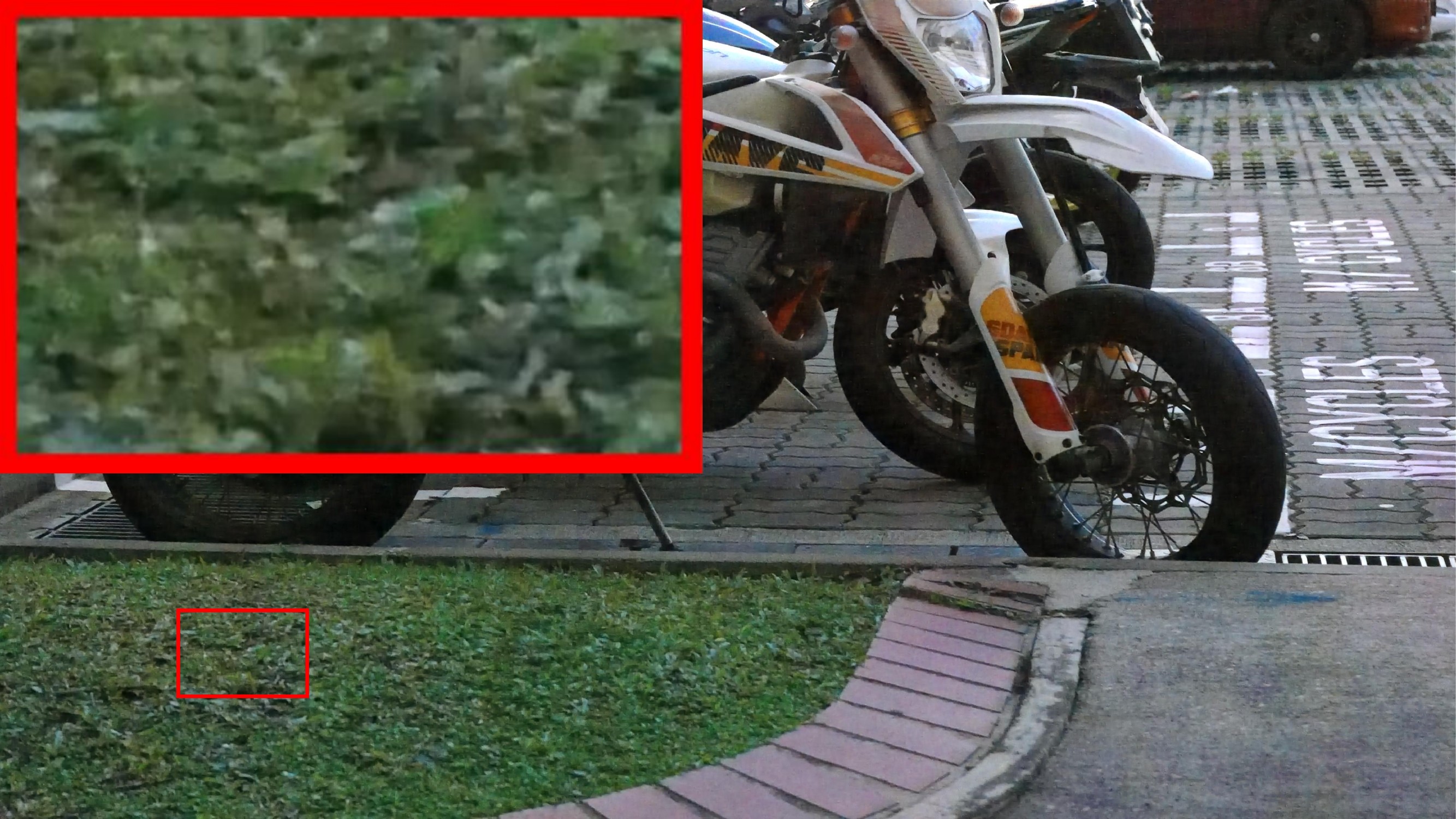} &
\includegraphics[width=0.498\columnwidth]{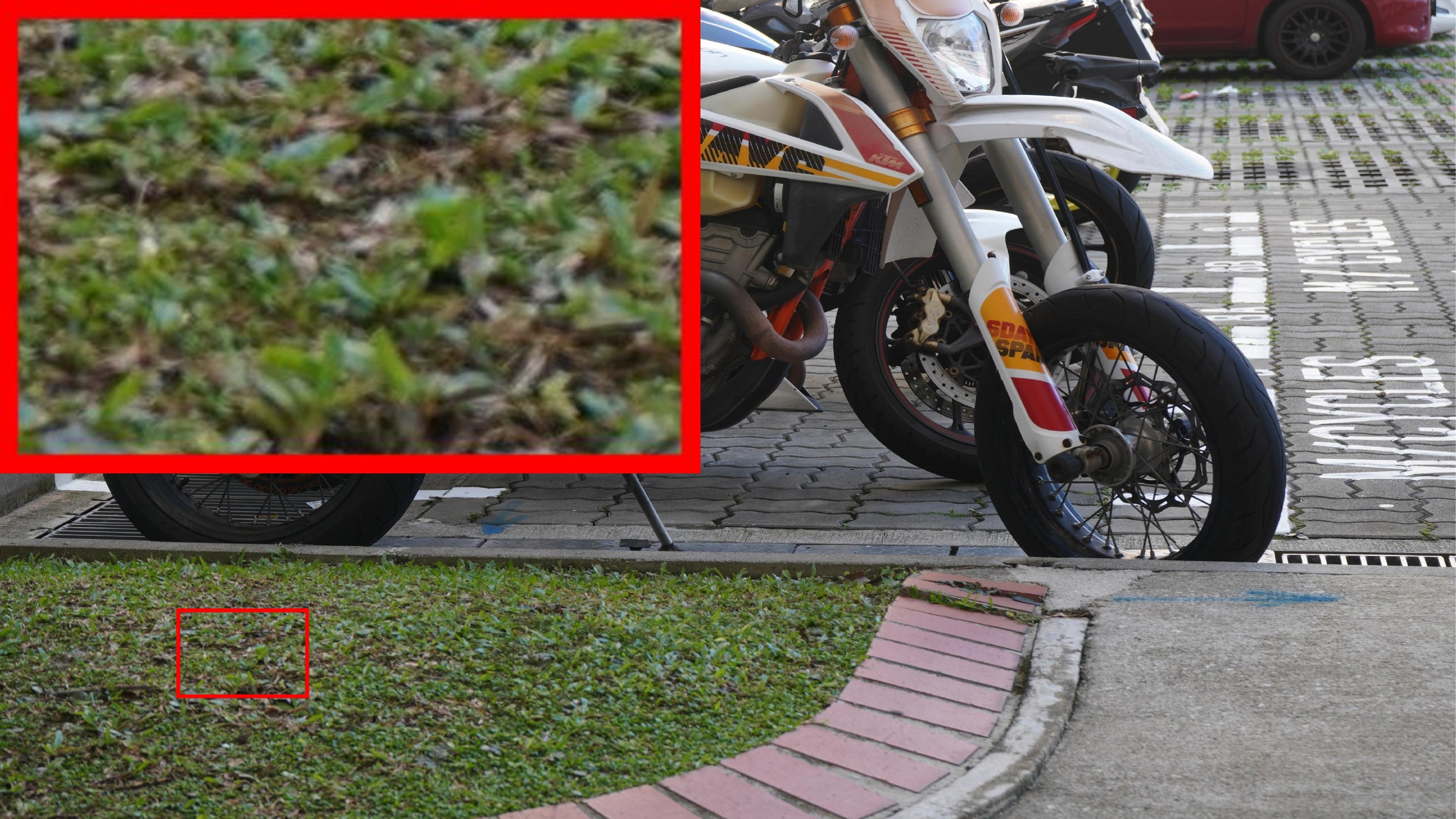}
\\[1.5pt]

\small(e) Wave-Mamba~\cite{zou2024wavemamba} &
\small(f) UHD-Processor~\cite{liu2025uhdprocessor} &
\small\textbf{(g) BiFreqNet (Ours)} &
\small(h) GT
\\

\end{tabular}

\endgroup

\resultcaption{
Visual comparison of UHD low-light image enhancement.
Existing spatial-/frequency-domain methods (b)--(d) and
frequency-decomposition-based methods (e)--(f) suffer from suboptimal
restoration results due to insufficient cross-frequency interaction.
In contrast, our BiFreqNet (g) produces visually pleasing results through
bidirectional interaction between low- and high-frequency representations.
}
\label{fig:teaser}
\vspace{-12pt}
\end{figure*}
However, due to their domain-specific modeling, they fail to fully exploit complementary spatial and frequency information, which makes it difficult to jointly restore illumination and fine details and leads to suboptimal results, as shown in Fig.~\ref{fig:teaser}(b)--(d). To address this problem, other works explicitly decompose image representations into different frequency components for illumination and detail restoration in UHD low-light images. Zou \etal~\cite{zou2024wavemamba} combine wavelet decomposition with state-space models to separately restore low- and high-frequency components. Liu \etal~\cite{liu2025uhdprocessor} progressively learn restoration mappings from low to high frequencies in a compact VAE latent space. Despite their success, these methods overlook the inherent relationship between illumination and structural information across frequency components, leading to unsatisfactory enhancement results, as shown in Fig.~\ref{fig:teaser}(e)--(f).

We observe that high-frequency structures provide boundary and local-contrast cues that distinguish different scene regions, which helps preserve structural boundaries when modeling spatially varying illumination during low-frequency restoration. In turn, the restored low-frequency representation provides richer illumination context for recovering missing structures and fine details in the high-frequency domain. This reveals that low- and high-frequency restoration are inherently interdependent rather than separable. Our key insight is that high-frequency structural cues can facilitate low-frequency illumination restoration, while the restored low-frequency context can in turn guide high-frequency detail reconstruction. Motivated by this insight, we explore a reciprocal enhancement process between low- and high-frequency representations with spatial--spectral collaboration in the low-frequency domain for UHD LLIE.

In this paper, we propose a novel Bidirectional Cross-frequency Domain Interaction Network, named BiFreqNet, for UHD LLIE. Specifically, BiFreqNet integrates a key component, \ie, a Dual-domain Feature Interaction (DFI) module, into the multi-scale features of a pretrained VAE to perform reciprocal enhancement between low- and high-frequency representations while collaboratively modeling spatial and spectral information in the low-frequency domain. DFI comprises two core components: the Spatial and Spectral-Aware Low-Frequency Enhancement (S$^2$LFE) block and the Cross-Frequency Enhancement (CFE) block. The S$^2$LFE block exploits high-frequency detail information to guide low-frequency restoration and jointly models spatial and spectral information to correct global exposure, recover spatially varying illumination, and suppress noise. Then, the CFE block leverages the recovered contextual information from the enhanced low-frequency representation to guide high-frequency refinement, facilitating faithful reconstruction of structures and fine details. As shown in Fig.~\ref{fig:teaser}(g), BiFreqNet achieves more natural illumination restoration while preserving clearer structures and finer details.
Our main contributions are summarized as follows:
\vspace{-2mm}
\begin{itemize}
    \item 
    We propose a novel Bidirectional Cross-frequency Domain Interaction Network (BiFreqNet) that enables reciprocal enhancement between low- and high-frequency representations for UHD low-light image enhancement.
    \item 
    We design a Dual-domain Feature Interaction (DFI) module with an S$^2$LFE block for high-frequency-guided low-frequency illumination enhancement and noise suppression, and a CFE block for low-frequency-guided high-frequency detail refinement, enabling effective bidirectional cross-frequency interaction.
    \item 
    Extensive experiments on four public UHD and general LLIE datasets demonstrate that BiFreqNet performs favorably against state-of-the-art general and UHD LLIE methods.
\end{itemize}

\section{Related Work}

\subsection{UHD Low-Light Image Enhancement}

UHD LLIE aims to recover illumination and image details while suppressing sensor noise in poorly illuminated 4K/8K images. However, general LLIE methods~\cite{XuKeCVPR2020, li2021zero, liu2021retinex, SNRNet, jiang2023diffll, cai2023retinexformer, weng2024mamballie, bai2024retinexmamba, he2025retidiff, URWKV, HVI_CIDNet, zhou2025gppllie, BiBridge} are mainly designed for conventional-resolution images and often suffer from high computational costs and limited capability to restore illumination and fine details at UHD resolutions. To address this issue, recent works have focused on designing efficient UHD LLIE methods. Some methods rely mainly on spatial- or frequency-domain modeling. Wang \etal~\cite{LLFormer} propose an efficient Transformer architecture to capture long-range dependencies with reduced computational cost. Li \etal~\cite{li2023uhdfour} perform Fourier-domain modeling to efficiently enhance illumination and suppress noise. Wu \etal~\cite{wu2026sce} convert sparse curves into look-up tables (LUTs) for efficient illumination and contrast adjustment. In addition, several UHD restoration methods also include LLIE as one of their tasks. Wang \etal~\cite{wang2024uhdformer} transfer useful high-resolution information to the low-resolution space through correlation matching to guide high-resolution restoration, while Wang \etal~\cite{Wang2026uhddip} extract spatial and detail information from normal and gradient priors in the low-resolution space to guide high-resolution restoration. Other methods explicitly exploit different frequency components for UHD restoration. Zou \etal~\cite{zou2024wavemamba} combine the wavelet transform with state space models to capture low-frequency illumination information and recover high-frequency details with linear complexity. Liu \etal~\cite{liu2025uhdprocessor} progressively learn restoration mappings from low to high frequencies in a compact VAE latent space with degradation-aware prompts.

Nevertheless, these methods mainly focus on either spatial- or frequency-domain modeling, or model frequency components without fully exploiting complementary spatial--spectral information and bidirectional interaction between low- and high-frequency features, leading to color distortion and detail loss. In this work, we introduce a spatial--frequency-aware bidirectional interaction method for UHD LLIE, which jointly models low-frequency spatial--spectral information for illumination restoration and noise suppression, and enables cross-frequency interaction for detail recovery.

\begin{figure*}[!t]
    \centering
    \begin{overpic}[width=0.98\textwidth]{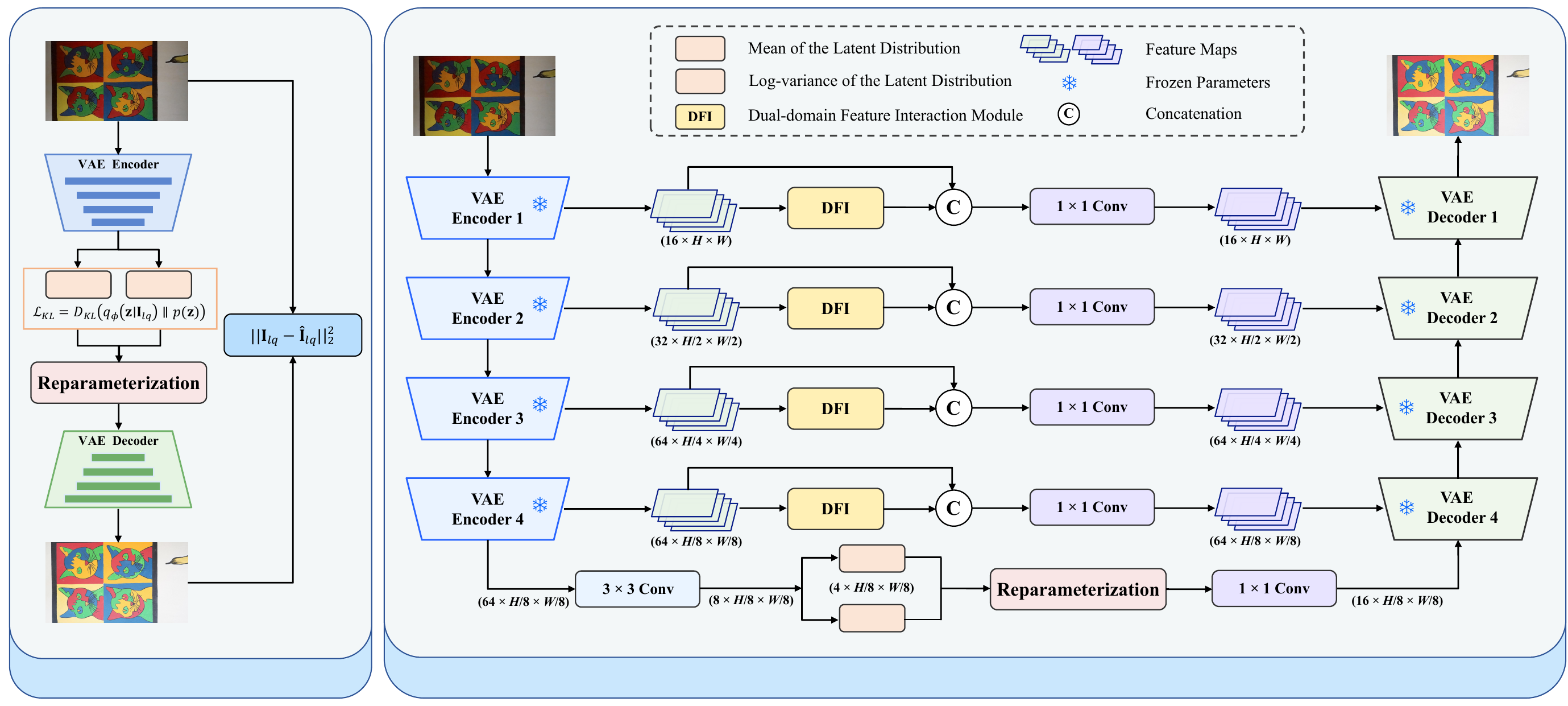} 
        \put(24.8, 41.9){{\scriptsize\fontfamily{ptm}\selectfont Low-light Image} {\footnotesize $\mathbf{I}_{lq}$}}
        \put(86.8, 41.9){{\scriptsize\fontfamily{ptm}\selectfont Enhanced Image} {\footnotesize $\mathbf{I}_{out}$}}

        \put(1.4, 42.7){{\scriptsize\fontfamily{ptm}\selectfont Low-light Image} {\footnotesize $\mathbf{I}_{lq}$}}
        \put(1.4, 3.8){{\scriptsize\fontfamily{ptm}\selectfont Reconstructed Image} {\footnotesize $\hat{\mathbf{I}}_{lq}$}}
        
        \put(37.9, 32.2){$\scriptstyle \mathbf{F}_1$}
        \put(37.9, 25.8){$\scriptstyle \mathbf{F}_2$}
        \put(37.9, 19.4){$\scriptstyle \mathbf{F}_3$}
        \put(37.9, 13){$\scriptstyle \mathbf{F}_4$}

        \put(74, 32.5){$\scriptstyle \mathbf{F}^{fus}_1$}
        \put(74, 26.1){$\scriptstyle \mathbf{F}^{fus}_2$} 
        \put(74, 19.7){$\scriptstyle \mathbf{F}^{fus}_3$} 
        \put(74, 13.4){$\scriptstyle \mathbf{F}^{fus}_4$} 

        \put(34.9, 7.9){$\scriptstyle \mathbf{h}$}
        \put(86.9, 7.9){$\scriptstyle \mathbf{Z}$} 

        \put(3.3, 26.4){$\scriptscriptstyle \mathbf{\log \sigma^2}$}
        \put(9.8, 26.6){$\scriptscriptstyle \mathbf{\mu}$} 

        \put(3.6, 1.2){{\scriptsize\fontfamily{ptm}\selectfont (a) Stage I: VAE Pre-training}}
        \put(31.2, 1.2){{\scriptsize\fontfamily{ptm}\selectfont (b) Stage II: The Full Training of Bidirectional Cross-frequency Domain Interaction Network (BiFreqNet)}}
             
        \put(54, 9){$\scriptscriptstyle \mathbf{\log \sigma^2}$}
        \put(55.2, 5.4){$\scriptscriptstyle \mathbf{\mu}$} 
    
        \put(43.3, 41.5){\scalebox{0.8}{$\scriptscriptstyle \mathbf{\log \sigma^2}$}}
        \put(44.3, 39.6){\scalebox{0.8}{$\scriptscriptstyle \mathbf{\mu}$}}

        \put(57.3, 32.2){$\scriptstyle \mathbf{\hat{F}}_1$} 
        \put(57.3, 25.8){$\scriptstyle \mathbf{\hat{F}}_2$} 
        \put(57.3, 19.4){$\scriptstyle \mathbf{\hat{F}}_3$} 
        \put(57.3, 13.2){$\scriptstyle \mathbf{\hat{F}}_4$} 
    \end{overpic}

    \caption{
    Overview of our proposed bidirectional cross-frequency domain interaction network (BiFreqNet), with two training stages.
    In Stage I (a), we pre-train a VAE to establish a compact latent space and extract reliable multi-scale image features for subsequent low-light enhancement.
    In Stage II (b), given a low-light image $\mathbf{I}_{lq}$, we first extract multi-scale features $\{\mathbf{F}_{i}\}_{i=1}^{4}$ using the frozen VAE encoder and enhance them with the proposed DFI modules, yielding $\{\widetilde{\mathbf{F}}_{i}\}_{i=1}^{4}$.
    We then fuse each $\widetilde{\mathbf{F}}_{i}$ with its corresponding $\mathbf{F}_{i}$ via concatenation and a $1\times1$ convolution to obtain $\{\mathbf{F}^{fus}_{i}\}_{i=1}^{4}$, while deriving the latent representation $\mathbf{Z}$ from $\boldsymbol{\mu}$ and $\log\boldsymbol{\sigma}^{2}$ via reparameterization.
    Finally, $\mathbf{Z}$ and $\{\mathbf{F}^{fus}_{i}\}_{i=1}^{4}$ are progressively decoded by the frozen VAE decoder to generate the enhanced image $\mathbf{I}_{out}$.
    }
    \label{fig:overall}
    \vspace{-4mm}
\end{figure*}

\subsection{Variational Auto-Encoders}

Variational Autoencoders (VAEs)~\cite{Kingma2014VAE} learn compact latent representations and have been widely used in image generation. With the development of latent diffusion models~\cite{Rombach2022LDM}, VAEs have enabled efficient image generation in low-dimensional latent spaces and have recently been extended to image restoration. Wang \etal~\cite{Wang2024StableSR} and Wu \etal~\cite{Wu2024SeeSR} exploit pretrained latent diffusion priors for real-world image super-resolution. Cascarano \etal~\cite{Cascarano2025RELD} extend latent diffusion priors to denoising, deblurring, and super-resolution. Biquard \etal~\cite{Biquard2025VBLE} perform variational inference directly in the latent space of compressive autoencoders for efficient restoration and uncertainty estimation. For UHD image restoration, VAEs have further been adopted to reduce the computational cost of processing high-resolution images. UHD-Processor~\cite{liu2025uhdprocessor} progressively learns restoration mappings from low to high frequencies in the VAE latent space with degradation-aware prompts. DecoupleUHD~\cite{DecoupleUHD25} decouples degradation and background information in the VAE latent space and uses reversible feature fusion to preserve useful content. DreamUHD~\cite{liu2025dreamuhd} introduces a frequency-enhanced VAE for UHD restoration, using Fourier-based lightweight modeling and a wavelet-based adapter to compensate for high-frequency information loss.

Nevertheless, existing VAE-based UHD restoration methods mainly focus on improving latent representations and compensating for information loss caused by VAE compression, while the interaction between low-frequency illumination information and high-frequency structural details remains underexplored. In this work, we perform bidirectional cross-frequency interaction on multi-scale VAE encoder features to jointly improve illumination restoration and detail recovery for UHD low-light image enhancement.

\subsection{State Space Models}

State space models (SSMs) provide an efficient framework for modeling long-range dependencies with linear computational complexity. Gu \etal~\cite{Gu2023MambaSSM} introduce Mamba with selective state updates for efficient long-sequence modeling. Recent works further extend Mamba from sequential modeling to visual representation and image restoration. Zhu \etal~\cite{Zhu2024VisionMambaICML} introduce state-space modeling into visual representation learning, while Guo \etal~\cite{Guo2024MambaIR} apply Mamba to image restoration. To better adapt selective scanning to two-dimensional image features, Huang \etal~\cite{Huang2024LocalMamba} employ windowed selective scanning to strengthen local interactions, while Li \etal~\cite{Li2025MaIR} preserve spatial locality and continuity during state propagation. Such methods enable efficient long-range spatial modeling while preserving local image structures. More recently, Mamba has been introduced into low-light image enhancement. Weng \etal~\cite{weng2024mamballie} combine global and local state-space modeling for low-light enhancement. Bai \etal~\cite{bai2024retinexmamba} incorporate Retinex-based illumination modeling into Mamba, while Zou \etal~\cite{zou2024wavemamba} combine wavelet decomposition with state-space modeling for low- and high-frequency restoration.

Nevertheless, these methods do not jointly exploit long-range spatial dependencies for modeling spatially varying illumination, spectral information, and high-frequency structural guidance. In this work, we employ Mamba to model long-range dependencies across spatially varying illumination regions, while incorporating spectral information and high-frequency structural cues for reliable illumination restoration and detail preservation.

\section{Methodology}
\subsection{Overview}

The overall pipeline of BiFreqNet is shown in Fig.~\ref{fig:overall}, which involves two training stages: VAE pre-training (a) and full training of BiFreqNet (b). The former learns a compact latent representation and reliable multi-scale image features for subsequent enhancement, while the latter employs BiFreqNet to enhance these multi-scale features through bidirectional cross-frequency interaction and reconstruct the enhanced image.

In the VAE pre-training stage, we first train the VAE on the UHD-LL dataset~\cite{li2023uhdfour} to establish a compact latent space and provide reliable multi-scale image features $\{\mathbf{F}_i \in \mathbb{R}^{ C_i \times H/2^{i-1}\times W/2^{i-1}}\}_{i=1}^{4}$ with rich contextual information for subsequent low-light enhancement. Following the VAE design in~\cite{Rombach2022LDM}, the encoder maps the low-light image $\mathbf{I}_{lq} \in \mathbb{R}^{3 \times H \times W}$ into the latent space and estimates the mean $\boldsymbol{\mu} \in \mathbb{R}^{ 4 \times H / 8 \times W / 8}$ and log-variance $\log \boldsymbol{\sigma}^{2} \in \mathbb{R}^{ 4 \times H / 8 \times W / 8}$ of the posterior distribution $q_{\phi}(\mathbf{z}\mid\mathbf{I}_{lq})$. Finally, we apply the reparameterization trick to sample the latent representation $\mathbf{z} \in \mathbb{R}^{4 \times H / 8 \times W / 8}$ from the learned posterior distribution, \ie, $\mathbf{z}=\boldsymbol{\mu}+\boldsymbol{\sigma}\odot\boldsymbol{\epsilon}$ with $\boldsymbol{\epsilon}\sim\mathcal{N}(0,\mathbf{I})$, and feed $\mathbf{z}$ into the VAE decoder to reconstruct $\mathbf{\hat{I}}_{lq}$. The training objective of this stage is given in Eq.~\ref{eq:vae_loss} of Sec.~\ref{sec:loss}.

After pre-training, the VAE encoder and decoder are frozen in the subsequent LLIE stage. Specifically, the frozen encoder extracts multi-scale features $\{\mathbf{F}_i\}_{i=1}^{4}$ from $\mathbf{I}_{lq}$. The dual-domain feature interaction module (Sec.~\ref{sec:dfi}) then performs cross-frequency interaction on each $\mathbf{F}_i \in \mathbb{R}^{ C_i \times H/2^{i-1}\times W/2^{i-1}}$ to obtain the enhanced feature $\hat{\mathbf{F}}_i  \in \mathbb{R}^{ C_i \times H/2^{i-1}\times W/2^{i-1}}$, which is concatenated with $\mathbf{F}_i$ and projected by a $1\times1$ convolution to produce $\mathbf{F}^{{fus}}_i  \in \mathbb{R}^{ C_i \times H/2^{i-1}\times W/2^{i-1}}$. Finally, the fused multi-scale features are fed into the frozen VAE decoder to reconstruct the enhanced image $\mathbf{I}_{out}  \in \mathbb{R}^{3 \times H \times W}$. The training objective of the LLIE stage is given in Eq.~\ref{eq:rec_loss} of Sec.~\ref{sec:loss}.

\subsection{Dual-Domain Feature Interaction (DFI) Module}
\label{sec:dfi}

\begin{figure*}[t]
    \centering
    \vspace{2mm}
    \begin{overpic}[width=0.92\textwidth]{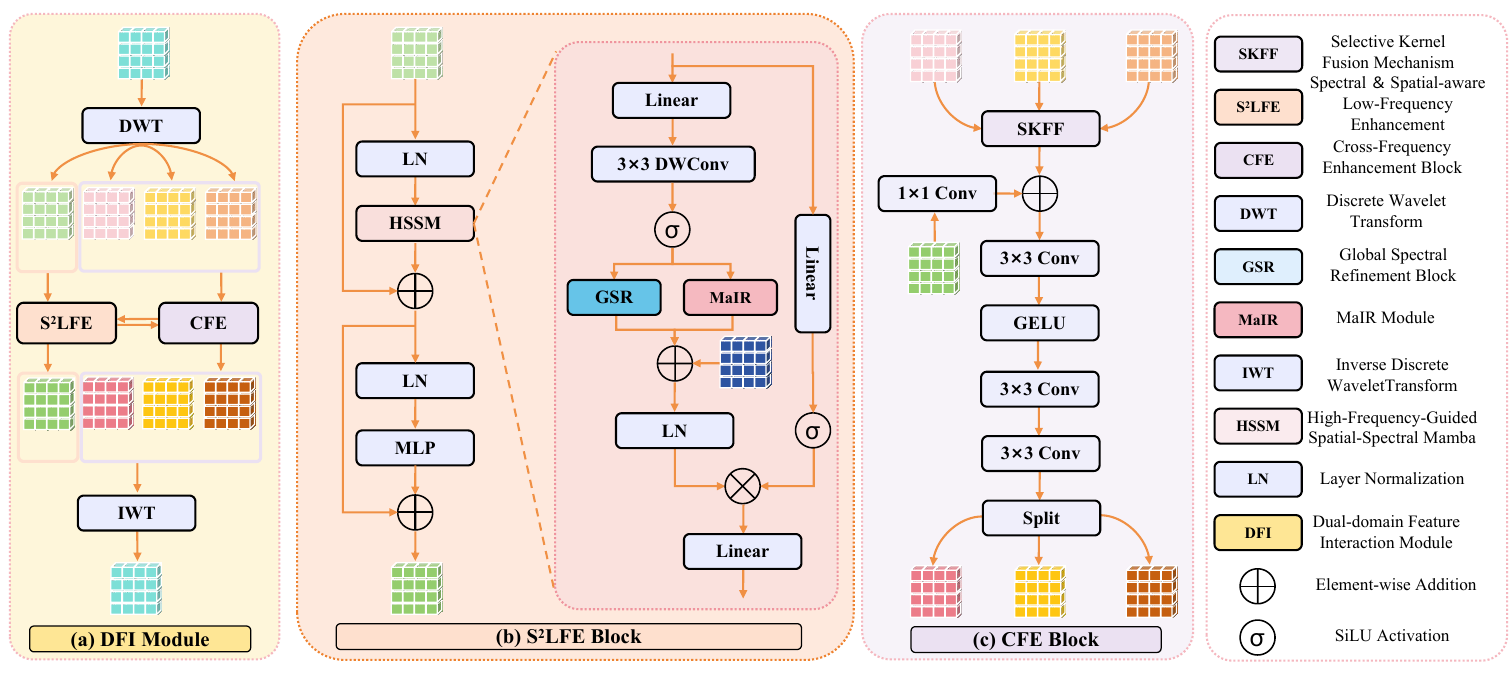}
        \put(96.2, 40.5){\scalebox{0.5}{\cite{ZamirAKHKYS20} }}
        \put(94.6, 23.7){\scalebox{0.5}{~\cite{Li2025MaIR} }}
        
        \put(1.8, 27.8){$\scriptscriptstyle \mathbf{F}^{LL}_{i}$}
        \put(5.8, 27.8){$\scriptscriptstyle \mathbf{F}^{HL}_{i}$}
        \put(9.9, 27.8){$\scriptscriptstyle \mathbf{F}^{LH}_{i}$}
        \put(13.9, 27.8){$\scriptscriptstyle \mathbf{F}^{HH}_{i}$}

        \put(22.7, 41.1){$\scriptscriptstyle \mathbf{F}^{LL}_{i}$}
        \put(28.1, 27.45){$\scriptscriptstyle \mathbf{\bar{F}}^{LL}_{i}$}
        \put(57.4, 41.1){\scalebox{0.9}{$\scriptscriptstyle \mathbf{F}^{HL}_{i}$}}
        \put(64.2, 41.1){\scalebox{0.9}{$\scriptscriptstyle \mathbf{F}^{LH}_{i}$}}
        \put(71.4, 41.1){\scalebox{0.9}{$\scriptscriptstyle \mathbf{F}^{HH}_{i}$}}

        \put(28.1, 22.6){$\scriptscriptstyle \mathbf{\tilde{F}}_{i}^{LL}$}
        \put(28.9, 10.6){$\scriptscriptstyle \mathbf{F}^{LL}_{i+1}$}
        
        \put(23, 5.6){$\scriptscriptstyle \hat{\mathbf{F}}^{LL}_{i}$}
        \put(57.3, 5.6){\scalebox{0.9}{$\scriptscriptstyle \hat{\mathbf{F}}^{HL}_{i}$}}
        \put(64.3, 5.6){\scalebox{0.9}{$\scriptscriptstyle \hat{\mathbf{F}}^{LH}_{i}$}}
        \put(71.4, 5.6){\scalebox{0.9}{$\scriptscriptstyle \hat{\mathbf{F}}^{HH}_{i}$}}

        \put(2, 15.2){$\scriptscriptstyle \hat{\mathbf{F}}^{LL}_{i}$}
        \put(5.9, 15.2){$\scriptscriptstyle \hat{\mathbf{F}}^{HL}_{i}$}
        \put(9.9, 15.2){$\scriptscriptstyle \hat{\mathbf{F}}^{LH}_{i}$}
        \put(13.9, 15.2){$\scriptscriptstyle \hat{\mathbf{F}}^{HH}_{i}$}
        
        \put(5.6, 5.5){$\scriptscriptstyle \hat{\mathbf{F}}^{}_{i}$}
        \put(6.1, 41){$\scriptscriptstyle \mathbf{F}_{i}$}
        \put(60.4, 24){$\scriptscriptstyle \hat{\mathbf{F}}^{LL}_{i}$}
        \put(46, 5.8){$\scriptscriptstyle \mathbf{\bar{F}}_{i}^{LL}$}

        \put(8.4, 24.6){$\scriptscriptstyle \mathbf{F}^{H}_{i}$}
        \put(7.6, 21.6){$\scriptscriptstyle \hat{\mathbf{F}}^{LL}_{i}$}
        
        \put(42.3, 35.8){$\scriptscriptstyle \mathbf{U}_i$}
        \put(40.6, 28.2){$\scriptscriptstyle \mathbf{U}^s_i$}
        \put(46.8, 28.2){$\scriptscriptstyle \mathbf{U}^s_i$}
        \put(50.5, 22.9){$\scriptscriptstyle \mathbf{Y}^g_i$}
        \put(37.3, 22.5){$\scriptscriptstyle \mathbf{Y}^f_i$}
        \put(69.5, 33.8){$\scriptscriptstyle \mathbf{F}^{H}_{i}$}
        \put(39.4, 20.6){$\scriptscriptstyle \mathbf{Y}_{i}^{mix}$}

        \put(48.4, 17.8){$\scriptscriptstyle \mathbf{F}^{H}_{i}$}
        \put(51.4, 13.8){$\scriptscriptstyle \mathbf{G}_{i}$}
        \put(69.5, 29.7){$\scriptscriptstyle \mathbf{H}^{g}_{i}$}
        \put(41.3, 40.1){$\scriptscriptstyle \mathbf{F}_{i}^{LL}$}
    \end{overpic}

    \caption{Overview of our proposed dual-domain feature interaction (DFI) module. It consists of a spatial- and spectral-aware low-frequency enhancement (S$^2$LFE) block and a cross-frequency enhancement (CFE) block for reciprocally enhancing the low- and high-frequency features.}
    \label{fig:dfi}
    \vspace{-2mm}
\end{figure*}

Previous UHD LLIE methods typically learn effective spatial- or frequency-domain representations and perform reduced-resolution feature processing to achieve a favorable trade-off between computational efficiency and enhancement quality. Specifically, frequency-domain methods such as UHDFour~\cite{li2023uhdfour} perform Fourier-domain restoration for illumination adjustment and detail recovery, while spatial-domain methods such as UHDFormer~\cite{wang2024uhdformer} and UHDDIP~\cite{Wang2026uhddip} reduce computational cost through high-ratio downsampling and recover high-resolution features using cross-resolution interaction or structural priors. More recently, cross-frequency methods such as Wave-Mamba~\cite{zou2024wavemamba} employ wavelet decomposition to model long-range dependencies in low-frequency features and use the enhanced low-frequency information to guide high-frequency refinement. Despite these advances, two limitations remain:
(1) spatial and spectral information are not jointly exploited for modeling global exposure, spatially varying illumination, and sensor noise;
and (2) cross-frequency interaction is mainly dominated by low-to-high guidance, while high-frequency structural cues are rarely used to constrain low-frequency restoration.

To this end, we propose a Dual-Domain Feature Interaction (DFI) module, which consists of $N$ stacked Spatial and Spectral-Aware Low-Frequency Enhancement (S$^2$LFE) blocks followed by a Cross-Frequency Enhancement (CFE) block, as shown in Fig.~\ref{fig:dfi}(a), to jointly model spatial and spectral information in the low-frequency domain and enable bidirectional interaction between low- and high-frequency features. S$^2$LFE employs Mamba-based selective scanning to propagate long-range spatial context across regions with different illumination degradations and to suppress sensor noise under the guidance of the fused multi-directional high-frequency representation, while FFT-based amplitude--phase modeling adjusts global exposure. CFE subsequently leverages the recovered context from the enhanced low-frequency feature to guide high-frequency feature refinement for faithful detail restoration.

Specifically, given an input feature
$\mathbf{F}_{i} \in \mathbb{R}^{C_i\times H / 2^{i-1}\times W / 2^{i-1}}$ from the $i$-th VAE encoder stage ($i\in\{1,\ldots,4\}$), we first use the Haar discrete wavelet transform to decompose it into one low-frequency feature $\mathbf{F}^{LL}_{i} \in \mathbb{R}^{C_i \times H/2^{i} \times W/2^{i}}$
and three directional high-frequency features $\{\mathbf{F}^{HL}_{i}, \mathbf{F}^{LH}_{i}, \mathbf{F}^{HH}_{i}\} \in \mathbb{R}^{C_i \times H/2^{i} \times W/2^{i}}$. Then, $\mathbf{F}^{LL}_{i}$ is fed into the $N$ stacked S$^2$LFE blocks to recover illumination and color information while suppressing sensor noise under the guidance of the SKFF-fused high-frequency feature $\mathbf{F}_{i}^{H} \in \mathbb{R}^{C_i \times H/2^{i} \times W/2^{i}}$, yielding the context-enhanced low-frequency feature $\hat{\mathbf{F}}^{LL}_{i} \in \mathbb{R}^{C_i \times H/2^{i} \times W/2^{i}}$. Meanwhile, $\{\mathbf{F}^{HL}_{i}, \mathbf{F}^{LH}_{i}, \mathbf{F}^{HH}_{i}\}$ are passed to the CFE block to produce the context-aware high-frequency enhanced features $\{\hat{\mathbf{F}}^{HL}_{i}, \hat{\mathbf{F}}^{LH}_{i}, \hat{\mathbf{F}}^{HH}_{i}\}  \in \mathbb{R}^{C_i \times H/2^{i} \times W/2^{i}}$ guided by the recovered contextual information from $\hat{\mathbf{F}}^{LL}_{i}$. Subsequently, we employ the inverse Haar wavelet transform to reconstruct the enhanced feature
$\hat{\mathbf{F}}_{i} \in \mathbb{R}^{C_i \times H/2^{i-1} \times W/2^{i-1}}$
from $\hat{\mathbf{F}}^{LL}_{i}$ and
$\{\hat{\mathbf{F}}^{HL}_{i}, \hat{\mathbf{F}}^{LH}_{i}, \hat{\mathbf{F}}^{HH}_{i}\}$.

The resulting enhanced feature $\hat{\mathbf{F}}_{i}$ is finally concatenated with the corresponding encoder feature $\mathbf{F}_{i}$ and projected by a $1\times1$ convolution to obtain $\mathbf{F}^{fus}_{i}$, which is fed into the corresponding stage of the frozen VAE decoder for progressive image reconstruction.

\subsubsection{Spatial and Spectral-Aware Low-Frequency Enhancement (S$^2$LFE) Block}

\begin{figure*}[t]
    \centering
    \vspace{2mm}
    \begin{overpic}[width=0.92\textwidth]{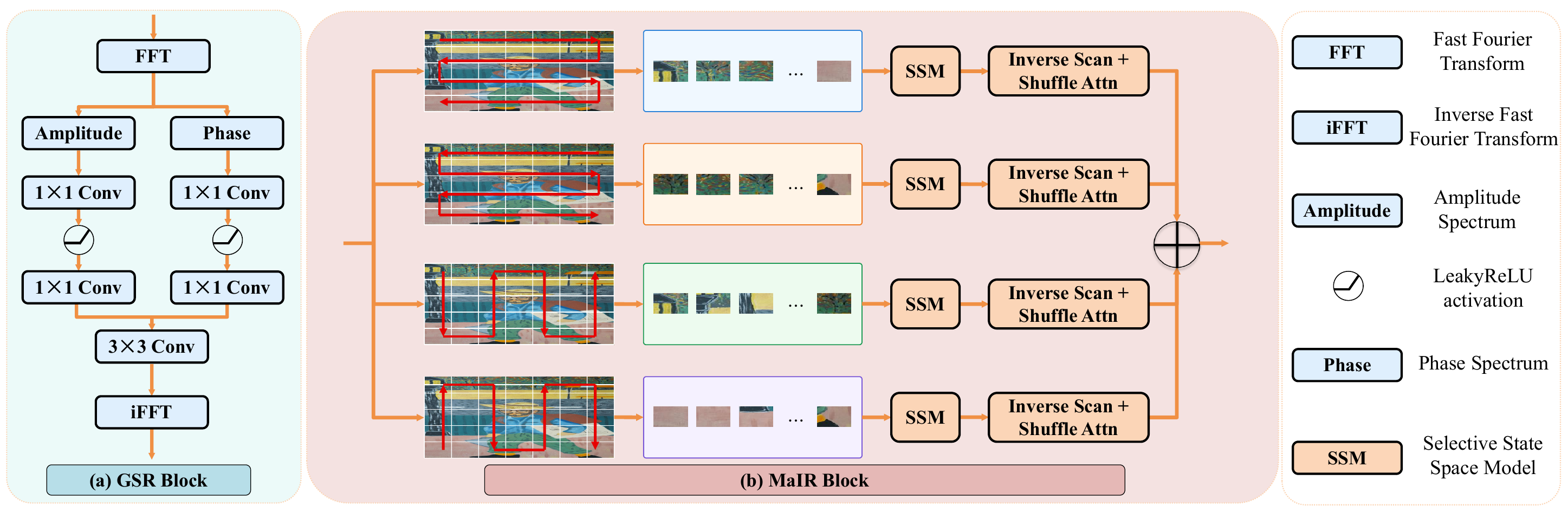}
        \put(10.2, 30.8){$\scriptscriptstyle \mathbf{U}_{i}^{s}$}
        \put(20, 16.8){$\scriptscriptstyle \mathbf{U}_{i}^{s}$}
        \put(3.1, 22.1){$\scriptscriptstyle \mathbf{A}_i$}
        \put(14.9, 22.1){$\scriptscriptstyle \mathbf{P}_i$}
        \put(3.2, 11.8){$\scriptscriptstyle \widehat{\mathbf{A}}_i$}
        \put(14.7, 11.8){$\scriptscriptstyle \widehat{\mathbf{P}}_i$}
        \put(10.2, 3.8){$\scriptscriptstyle \mathbf{Y}^{f}_i$}
        \put(10.2, 26.6){$\scriptscriptstyle \mathbf{Z}_{i}^{f}$}
        \put(10.2, 8){$\scriptscriptstyle \hat{\mathbf{Z}}_{i}^{f}$}
        \put(39.2, 29.1){\scalebox{0.7}{$\scriptscriptstyle \mathbf{S}_{1}^{(l)}$}}
        \put(39.2, 21.9){\scalebox{0.7}{$\scriptscriptstyle \mathbf{S}_{2}^{(l)}$}}
        \put(39.2, 14.2){\scalebox{0.7}{$\scriptscriptstyle \mathbf{S}_{3}^{(l)}$}}
        \put(39.2, 7){\scalebox{0.7}{$\scriptscriptstyle \mathbf{S}_{4}^{(l)}$}}
        \put(61.3, 26.6){\scalebox{0.9}{$\scriptscriptstyle {\mathbf{Y}}_{1}$}}
        \put(61.3, 21.4){\scalebox{0.9}{$\scriptscriptstyle {\mathbf{Y}}_{2}$}}
        \put(61.3, 13.8){\scalebox{0.9}{$\scriptscriptstyle {\mathbf{Y}}_{3}$}}
        \put(61.3, 6.6){\scalebox{0.9}{$\scriptscriptstyle {\mathbf{Y}}_{4}$}}
        \put(73.3, 26.6){\scalebox{0.9}{$\scriptscriptstyle \hat{\mathbf{Y}}_{1}$}}
        \put(73.3, 21.4){\scalebox{0.9}{$\scriptscriptstyle \hat{\mathbf{Y}}_{2}$}}
        \put(73.3, 13.8){\scalebox{0.9}{$\scriptscriptstyle \hat{\mathbf{Y}}_{3}$}}
        \put(73.3, 6.6){\scalebox{0.9}{$\scriptscriptstyle \hat{\mathbf{Y}}_{4}$}}
        \put(78.2, 16.8){$\scriptscriptstyle \mathbf{Y}_{i}^{s}$}
    \end{overpic}

    \caption{Overview of the global spectral refinement (GSR) block (a) and the Mamba for image restoration (MaIR) block (b). The GSR block aims to correct global exposure by refining the amplitude and phase spectra in the Fourier domain, while the MaIR block aims to model spatially varying illumination and suppress sensor noise through a nested S-shaped scanning strategy.
    }
    \label{fig:gsr_mair}
    \vspace{-2mm}
\end{figure*}

To effectively correct global exposure and restore spatially varying illumination while suppressing sensor noise in low-light images, we jointly model low-frequency representations in the spectral and spatial domains. However, low-frequency features contain limited structural information, which may lead to inaccurate context propagation across object boundaries and loss of useful details during noise suppression. Therefore, we propose a Spatial and Spectral-Aware Low-Frequency Enhancement (S$^2$LFE) block that incorporates our High-Frequency-Guided Spatial-Spectral Mamba (HSSM) block to jointly perform global exposure correction and spatially varying illumination restoration under high-frequency structural guidance while suppressing sensor noise, enabling reliable low-frequency restoration with preserved structural details, as shown in Fig.~\ref{fig:dfi}(b). The S$^2$LFE block can be formulated as:

\begin{equation}
\begin{aligned}
\mathbf{\tilde{F}}^{LL}_i &= \operatorname{HSSM}(\operatorname{LN}(\mathbf{F}^{LL}_i)) + \mathbf{F}^{LL}_i, \\
\mathbf{\hat{F}}^{LL}_i &= \operatorname{MLP}(\operatorname{LN}(\mathbf{\tilde{F}}^{LL}_i)) + \mathbf{\tilde{F}}^{LL}_i,
\end{aligned}  
\end{equation}
where $\operatorname{LN}(\cdot)$ denotes layer normalization, $\operatorname{HSSM}(\cdot)$ represents the HSSM block, and $\operatorname{MLP}(\cdot)$ is a multilayer perceptron.

\noindent\textbf{High-Frequency-Guided Spatial-Spectral Mamba (HSSM).} HSSM employs two parallel branches to model the spectral and spatial information of low-frequency features, while incorporating high-frequency structural cues to guide reliable low-frequency restoration. Following~\cite{Zhu2024VisionMambaICML}, the input low-frequency feature $\mathbf{{F}}^{LL}_i \in \mathbb{R}^{C_i\times H / 2^i \times W / 2^i}$ is first processed by a linear layer that expands its channel dimension to $2C$. The expanded feature is then fed into a $3 \times 3$ depth-wise convolution followed by SiLU activation to obtain the spatially enhanced content feature $\mathbf{U}^{s}_i \in \mathbb{R}^{2C \times H \times W}$, defined as:
\begin{equation}
    \mathbf{U}^{s}_i = \sigma(\mathcal{C}^{DW}_{3\times3}(\operatorname{Linear}(\mathbf{{F}}^{LL}_i))),
\end{equation}
where $\sigma(\cdot)$ denotes the SiLU activation function. $\mathbf{U}^{s}_i$ is then fed into the global spectral refinement (GSR) block and the Mamba for image restoration (MaIR)~\cite{Li2025MaIR} block for global exposure correction and spatially varying illumination restoration while suppressing sensor noise, respectively.

\noindent\textbf{Global Spectral Refinement (GSR) Block.} For global exposure correction, as shown in Fig.~\ref{fig:gsr_mair}(a), GSR first transforms $\mathbf{U}^{s}_i$ into the Fourier domain and decomposes the resulting spectrum into amplitude and phase components, defined as:
\begin{equation}
\begin{aligned}
\mathbf{Z}^{f}_i
&=\mathcal{F}(\mathbf{U}^{s}_i)
=\operatorname{Re}(\mathbf{Z}^{f}_i)
+j\operatorname{Im}(\mathbf{Z}^{f}_i),\\
\mathbf{A}_i
&=\sqrt{
\operatorname{Re}(\mathbf{Z}^{f}_i)^2+
\operatorname{Im}(\mathbf{Z}^{f}_i)^2
},\\
\mathbf{P}_i
&=\operatorname{atan2}(
\operatorname{Im}(\mathbf{Z}^{f}_i),
\operatorname{Re}(\mathbf{Z}^{f}_i)
),
\end{aligned}
\end{equation}
where $\mathcal{F}(\cdot)$ denotes the Fast Fourier Transform (FFT), $\operatorname{Re}(\cdot)$ and $\operatorname{Im}(\cdot)$ denote the real and imaginary components, and $\mathbf{A}_i$ and $\mathbf{P}_i$ represent the amplitude and phase spectra, respectively.
The amplitude spectrum mainly describes the global intensity distribution related to image exposure, while the phase spectrum retains important spatial structure information. We independently refine the amplitude and phase components using two $1\times1$ convolutions with LeakyReLU activation, defined as:
\begin{equation}
\begin{aligned}
\hat{\mathbf{A}}_i
&=\mathcal{C}_{1\times1}
\left(
\delta\left(
\mathcal{C}_{1\times1}(\mathbf{A}_i)
\right)
\right),\\
\hat{\mathbf{P}}_i
&=\mathcal{C}_{1\times1}
\left(
\delta\left(
\mathcal{C}_{1\times1}(\mathbf{P}_i)
\right)
\right),
\end{aligned}
\end{equation}
where $\delta(\cdot)$ denotes the LeakyReLU activation function.
The refined amplitude and phase components are then used to reconstruct the spectral representation:
\begin{equation}
\hat{\mathbf{Z}}^{f}_i
=\mathcal{R}(\hat{\mathbf{A}}_i,\hat{\mathbf{P}}_i)
=\hat{\mathbf{A}}_i\odot\exp\left(j\hat{\mathbf{P}}_i\right),
\end{equation}
where $\mathcal{R}(\cdot)$ denotes spectral reconstruction and $j$ is the imaginary unit.
The reconstructed spectrum is further refined by a $3\times3$ convolution and transformed back to the spatial domain, defined as:
\begin{equation}
\mathbf{Y}^{f}_i
=\mathcal{F}^{-1}
\left(
\mathcal{C}_{3\times3}
\left(
\hat{\mathbf{Z}}^{f}_i
\right)
\right),
\end{equation}
where $\mathcal{F}^{-1}(\cdot)$ denotes the inverse Fast Fourier Transform.
Through joint amplitude--phase refinement, GSR adjusts the global exposure distribution while preserving the spatial structure of the low-frequency representation.

\noindent\textbf{Mamba for Image Restoration (MaIR) Block.} To model spatially varying illumination and suppress sensor noise, as shown in Fig.~\ref{fig:gsr_mair}(b), we adopt the MaIR block~\cite{Li2025MaIR} to capture long-range dependencies across different spatial regions. Specifically, it employs a nested S-shaped scanning strategy~\cite{Li2025MaIR} with a shift-stripe mechanism between neighboring blocks to propagate context across illumination regions while preserving local continuity, facilitating spatially varying illumination modeling and suppressing scattered noise. After scanning, each sequence is processed by a State-Space Model (SSM) and restored to its original spatial ordering, followed by shuffle attention for adaptive channel modulation, yielding the spatial response $\mathbf{Y}^{s}_i$. 

The spectral response $\mathbf{Y}^{f}_i$ and spatial response $\mathbf{Y}^{s}_i$ are further integrated with high-frequency structural cues to provide complementary boundary and detail information for low-frequency restoration, helping suppress sensor noise and reduce inaccurate context propagation across object boundaries, defined as:
\begin{equation}
\mathbf{Y}^{mix}_i
=\mathbf{Y}^{f}_i+\mathbf{Y}^{s}_i+\mathbf{F}^{H}_i,
\end{equation}
where $\mathbf{F}^{H}_i$ is derived from the fused multi-directional high-frequency representation provided by the SKFF module~\cite{ZamirAKHKYS20} in CFE. 
\fsl{
The high-frequency representation $\mathbf{F}^{H}_i$ provides complementary boundary and local-contrast cues to distinguish different scene regions, helping preserve structural boundaries during spatially varying illumination modeling.
}

Meanwhile, the low-frequency feature $\mathbf{F}^{LL}_{i}$ is projected through a parallel linear layer followed by SiLU activation to generate a content-adaptive gating response $\mathbf{G}_{i}$, which selectively modulates the normalized fused representation $\mathbf{Y}_i^{mix}$ via element-wise multiplication, thereby adaptively regulating the information aggregated from the spatial and spectral branches under high-frequency structural guidance. Finally, a linear projection is applied to produce the output of HSSM, denoted as $\mathbf{\bar{F}}_{i}^{LL}$, defined as:
\begin{equation}
\begin{aligned}
\mathbf{G}_i = &\sigma(\operatorname{Linear}(\mathbf{F}_i^{LL})), \\
\mathbf{\bar{F}}_{i}^{LL}
=&\operatorname{Linear}(
\operatorname{LN}(\mathbf{Y}^{mix}_i)
\odot \mathbf{G}_i),
\end{aligned}
\end{equation}
where $\odot$ denotes element-wise multiplication. $\mathbf{\bar{F}}_{i}^{LL}$ is reshaped back to the spatial domain to obtain the context-enhanced low-frequency feature $\hat{\mathbf{F}}^{LL}_{i}$, which is subsequently delivered to the CFE block for high-frequency detail refinement.

\subsubsection{Cross-Frequency Enhancement (CFE) Block}

Although the enhanced low-frequency features provide useful contextual information, restoring high-frequency details remains challenging because these details are highly susceptible to noise and strongly depend on low-frequency illumination cues. Existing methods such as Wave-Mamba~\cite{zou2024wavemamba} exploit low-frequency information for high-frequency refinement, but their cross-frequency interaction mainly relies on hierarchical feature propagation, without explicitly modeling the relationships among directional high-frequency sub-bands.

To address this limitation, we propose a Cross-Frequency Enhancement (CFE) block that jointly refines the three directional high-frequency features to promote cross-directional information interaction and structural consistency, while leveraging the recovered contextual information from the enhanced low-frequency feature to guide high-frequency refinement for faithful detail restoration.

Given the context-enhanced low-frequency feature
$\hat{\mathbf{F}}^{LL}_{i} \in \mathbb{R}^{C_i \times H / 2^i \times W/ 2^i}$, the three directional high-frequency features
$\{\mathbf{F}^{HL}_{i},\mathbf{F}^{LH}_{i},\mathbf{F}^{HH}_{i}\}$ are first aggregated by the selective kernel feature fusion
(SKFF) block~\cite{ZamirAKHKYS20}, yielding the high-frequency fusion representation $\mathbf{F}^{H}_{i} \in \mathbb{R}^{C_i \times H / 2^i \times W/ 2^i}$, defined as:
\begin{equation}
\mathbf{F}^{H}_{i} =\operatorname{SKFF}
\left(
\mathbf{F}^{HL}_{i},
\mathbf{F}^{LH}_{i},
\mathbf{F}^{HH}_{i}
\right).
\label{eq:cfe_skff}
\end{equation}

Then, we leverage $\mathbf{\hat{F}}_i^{LL}$ to guide the refinement of $\mathbf{F}^{H}_{i}$, recovering missing detail information and obtaining the context-aware high-frequency fused feature $\mathbf{H}_{i}^{g}$, defined as:
\begin{equation}
\mathbf{H}_{i}^{g} = \mathbf{F}^{H}_{i} + \mathcal{C}_{1\times 1}(\hat{\mathbf{F}}^{LL}_{i}).
\label{eq:cfe_guidance}
\end{equation}
Through this cross-frequency interaction in Eq.~(\ref{eq:cfe_guidance}), the contextual information recovered in the low-frequency feature directly guides high-frequency reconstruction, facilitating detail recovery under spatially varying illumination.

Finally, the guided representation is further refined by two $3\times3$ convolutions with a GELU activation function, and then projected to $3C_i$ channels and split into three directional high-frequency components, generating the context-aware high-frequency enhanced
features $\{\hat{\mathbf{F}}^{HL}_{i}, \hat{\mathbf{F}}^{LH}_{i},  \hat{\mathbf{F}}^{HH}_{i}\}$, defined as:
\begin{equation}
\hat{\mathbf{F}}^{HL}_{i},
\hat{\mathbf{F}}^{LH}_{i},
\hat{\mathbf{F}}^{HH}_{i}
 =
\operatorname{Split}
(
\mathcal{C}_{3\times 3}(
\operatorname{GELU}
(
\mathcal{C}_{3 \times 3}(
\mathbf{H}^{g}_i)
))
),
\label{eq:cfe_refinement}
\end{equation}
where $\mathcal{C}_{3 \times 3}(\cdot)$, $\operatorname{Split}(\cdot)$, and $\operatorname{GELU}(\cdot)$ denote the $3\times3$ convolution, channel-wise splitting, and GELU activation operations, respectively.

\subsection{Loss Function} 
\label{sec:loss}
During the VAE pre-training stage, following~\cite{Rombach2022LDM}, we adopt the MSE loss and the Kullback--Leibler (KL) divergence~\cite{Kingma2014VAE} to learn the multi-scale image representations while preserving the compact latent representation of the low-light image with learned generative priors, defined as:
\begin{equation} 
\mathcal{L}_{VAE} = \lVert \mathbf{\hat{I}}_{lq} - \mathbf{I}_{lq} \rVert_{2}^{2} +  \lambda_{KL} \mathbb{D}_{KL}(q_{\phi}(\mathbf{z}\mid\mathbf{I}_{lq}) || p(\mathbf{z})),
\label{eq:vae_loss} 
\end{equation} 
where $\lambda_{{KL}}$ is the weight of the KL divergence term and is set to $10^{-6}$ following~\cite{Rombach2022LDM}.

During the full training phase, to faithfully reconstruct the enhanced image $\mathbf{I}_{out}$ from the degraded observation $\mathbf{I}_{lq}$, we first employ the $\ell_1$ loss to align $\mathbf{I}_{out}$ with its corresponding ground-truth image $\mathbf{I}_{gt}$, formulated as:
\begin{equation} 
\mathcal{L}_1  = 
\lVert \mathbf{I}_{out}  - \mathbf{I}_{gt} \rVert_1. 
\label{eq:l1_loss} 
\end{equation}

Then, we use the multi-scale Charbonnier loss~\cite{CharbonnierBAB94} to enforce illumination and color consistency at coarse scales while refining local structures and details at finer scales, defined as:
\begin{equation}
\begin{aligned}
\mathcal{L}_{char} &= \frac{1}{|\mathcal{S}|}\sum_{s=1}^{4}\frac{1}{P_{s}}\sum_{p=1}^{P_s} \\
&\quad \sqrt{\bigl(\mathcal{I}_{p}(\mathcal{D}_{s}(\mathbf{I}_{out}))
 - \mathcal{I}_{p}(\mathcal{D}_{s}(\mathbf{I}_{gt}))\bigr)^2 + \epsilon^2},
\end{aligned}
\label{eq:charbonnier_loss} 
\end{equation}
where $|\mathcal{S}|$ denotes the number of scales and is set to 4, $P_s$ represents the number of pixels at the $s$-th scale,
$\mathcal{D}_{s}(\cdot)$  denotes the downsampling operator with a scale factor of $2^{s-1}$, and $\mathcal{I}_{p}(\cdot)$ represents the indexing operation at the $p$-th pixel.
$\epsilon$ is a small constant introduced for numerical stability and is empirically set to $1\times10^{-6}$.

In summary, the overall loss for the full training stage is formulated as:
\begin{equation} 
\label{eq:rec_loss}
\mathcal{L}_{{rec}} 
= 
\mathcal{L}_1 
+ 
\lambda_{{char}}\mathcal{L}_{{char}}, 
\end{equation} 
where $\lambda_{{char}}$ is the weight of the Charbonnier loss and is set to $0.2$ by default.

\flushbottom

\begin{table*}[t]
\centering
\renewcommand{\arraystretch}{1.0}
\setlength{\tabcolsep}{8.7pt}

\caption{Quantitative comparison with state-of-the-art methods on the UHD-LL~\cite{li2023uhdfour} and UHD-LOL4K~\cite{LLFormer} datasets. 
Higher PSNR and SSIM values indicate better performance, whereas lower Params and FLOPs values are preferred. FLOPs are computed on a $512\times512$ input.
}
\label{tab:sota_uhd}

\footnotesize
\begin{tabular}{c|l|l|cccc|cc}
\Xhline{1pt}

\multirow{2}{*}{\textbf{Type}}
&
\multirow{2}{*}{\textbf{Method}}
&
\multirow{2}{*}{\textbf{Venue}}
&
\multicolumn{2}{c}{\textbf{UHD-LL}~\cite{li2023uhdfour}}
&
\multicolumn{2}{c|}{\textbf{UHD-LOL4K}~\cite{LLFormer}}
&
\multirow{2}{*}{\textbf{Params (M) $\downarrow$}}
&
\multirow{2}{*}{\textbf{FLOPs (G) $\downarrow$}}
\\[-1pt]

\cline{4-5}
\cline{6-7}

& & &
\textbf{PSNR}~$\uparrow$
&
\textbf{SSIM}~$\uparrow$
&
\textbf{PSNR}~$\uparrow$
&
\textbf{SSIM}~$\uparrow$
&
&
\\

\Xhline{0.5pt}
\multirow{6}{*}{non-UHD}

& Zero-DCE++~\cite{li2021zero}
& TPAMI'22
& 16.41 & 0.630
& 15.58 & 0.835
& 0.001
& 1.32 \\

& RUAS~\cite{liu2021retinex}
& CVPR'21
& 13.56 & 0.749
& 14.68 & 0.758
& 0.0003
& 0.83 \\

& Uformer~\cite{wang2022uformer}
& CVPR'22
& 19.28 & 0.849
& 29.99 & 0.980
& 20.63 & 43.86 \\

& Restormer~\cite{zamir2022restormer}
& CVPR'22
& 22.25 & 0.871
& 36.91 & 0.988
& 26.13 & 141.75 \\

& DiffLL~\cite{jiang2023diffll}
& SIGGRAPH'23
& 21.36 & 0.872
& 36.95 & 0.989
& 17.29 & 29.74 \\

& Reti-Diff~\cite{he2025retidiff}
& ICLR'25
& 26.92 & 0.905
& 25.15 & 0.949
& 26.11 & 87.63 \\

\Xhline{0.5pt}

\multirow{8}{*}{UHD}
& LLFormer~\cite{LLFormer}
& AAAI'23
& 22.79 & 0.853
& \underline{37.33} & 0.989
& 24.52 & 22.52 \\

& UHDFour~\cite{li2023uhdfour}
& ICLR'23
& 26.23 & 0.900
& 36.12 & \underline{0.990}
& 17.54 & 75.63 \\

& UHDFormer~\cite{wang2024uhdformer}
& AAAI'24
& 27.11 & 0.927
& 36.28 & 0.989
& 0.34
& 48.37 \\

& Wave-Mamba~\cite{zou2024wavemamba}
& MM'24
& \underline{27.35} & 0.913
& \textbf{37.43} & \underline{0.990}
& 1.26 & 0.88 \\

& UHD-Processor~\cite{liu2025uhdprocessor}
& CVPR'25
& 27.22 & \textbf{0.929}
& 36.40 & 0.968
& 1.60 & 4.17 \\

& UHDDIP~\cite{Wang2026uhddip}
& TCSVT'25
& 26.75 & \underline{0.928}
& 35.46 & 0.978
& 0.81 & 34.73 \\

& SCE~\cite{wu2026sce}
& TCSVT'26
& 20.24 & 0.715
& 21.30 & 0.934
& 0.03
& -- \\
& 
\cellcolor{gray!30}\textbf{BiFreqNet (Ours)}
& \cellcolor{gray!30} --
& \cellcolor{gray!30}\textbf{27.39} 
& \cellcolor{gray!30}\textbf{0.929}
& \cellcolor{gray!30}37.29
& \cellcolor{gray!30}\textbf{0.992}
& \cellcolor{gray!30}9.26 
& \cellcolor{gray!30}55.33 \\

\Xhline{1pt}
\end{tabular}
\end{table*}
\begin{figure*}[!t]

    \centering

    \begingroup
    \setlength{\tabcolsep}{2.5pt}
    \renewcommand{\arraystretch}{1.0}

    \begin{tabular}{@{}ccccc@{}}

        \includegraphics[width=0.192\textwidth]
        {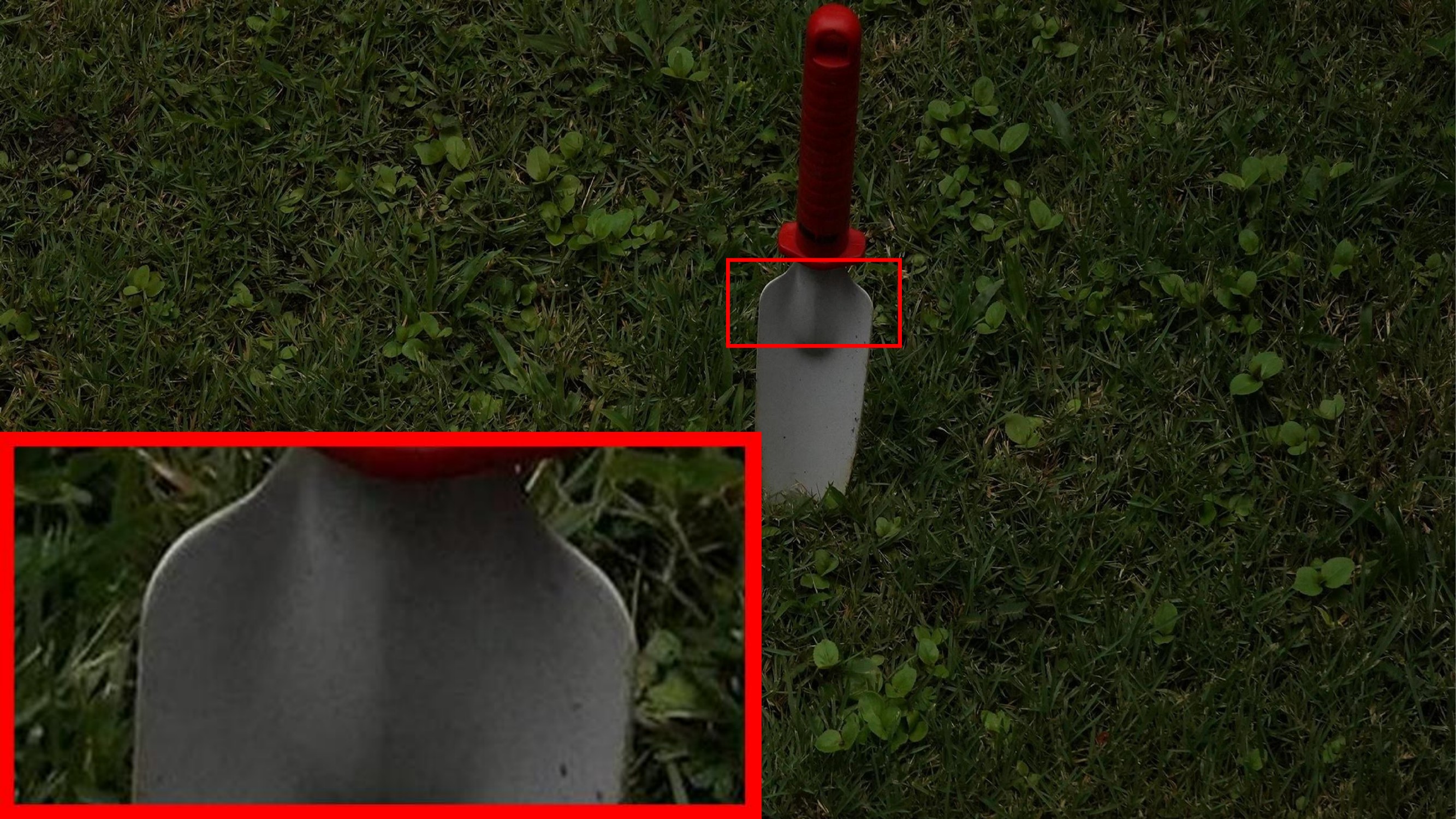}
        &
        \includegraphics[width=0.192\textwidth]
        {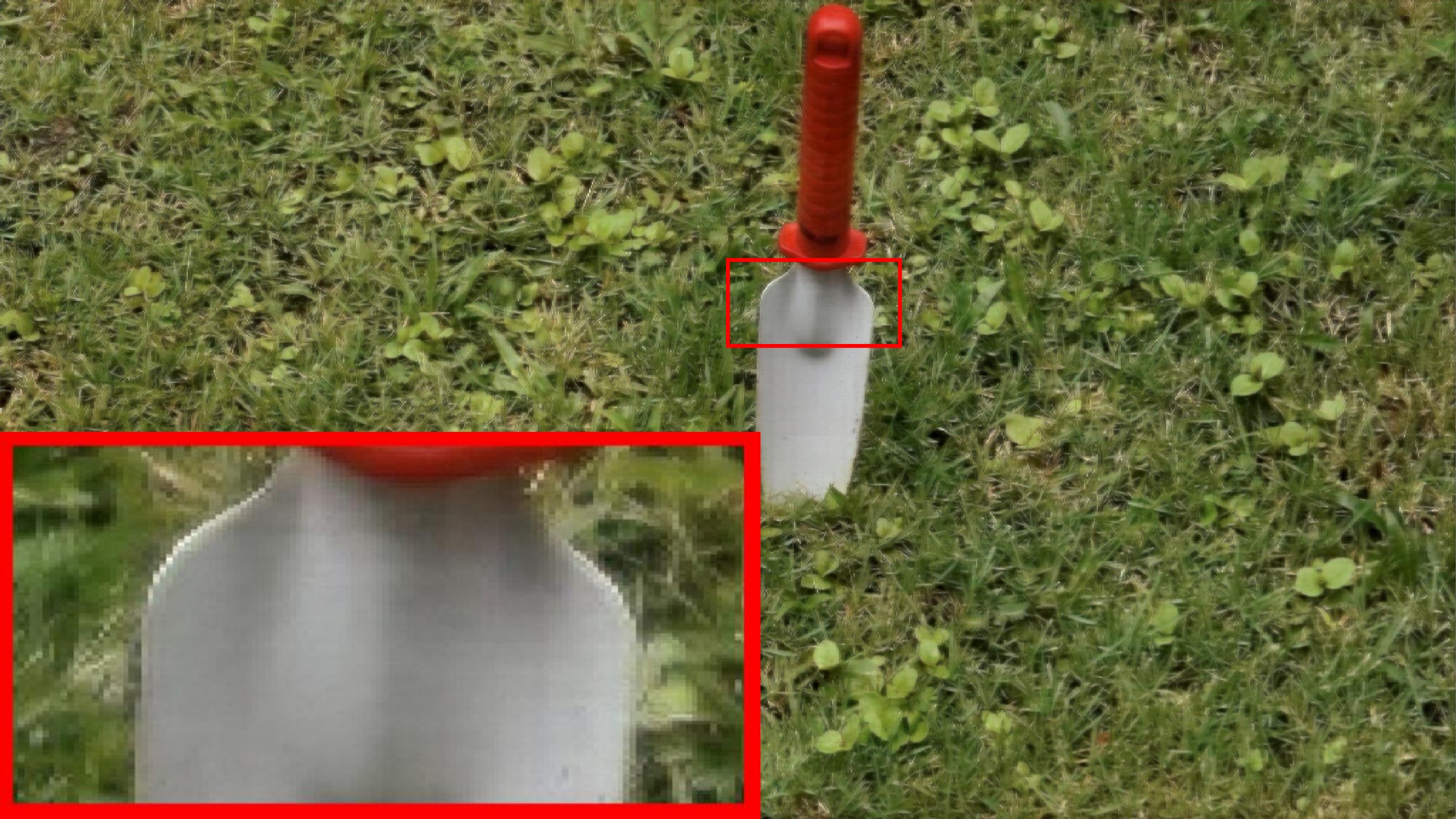}
        &
        \includegraphics[width=0.192\textwidth]
        {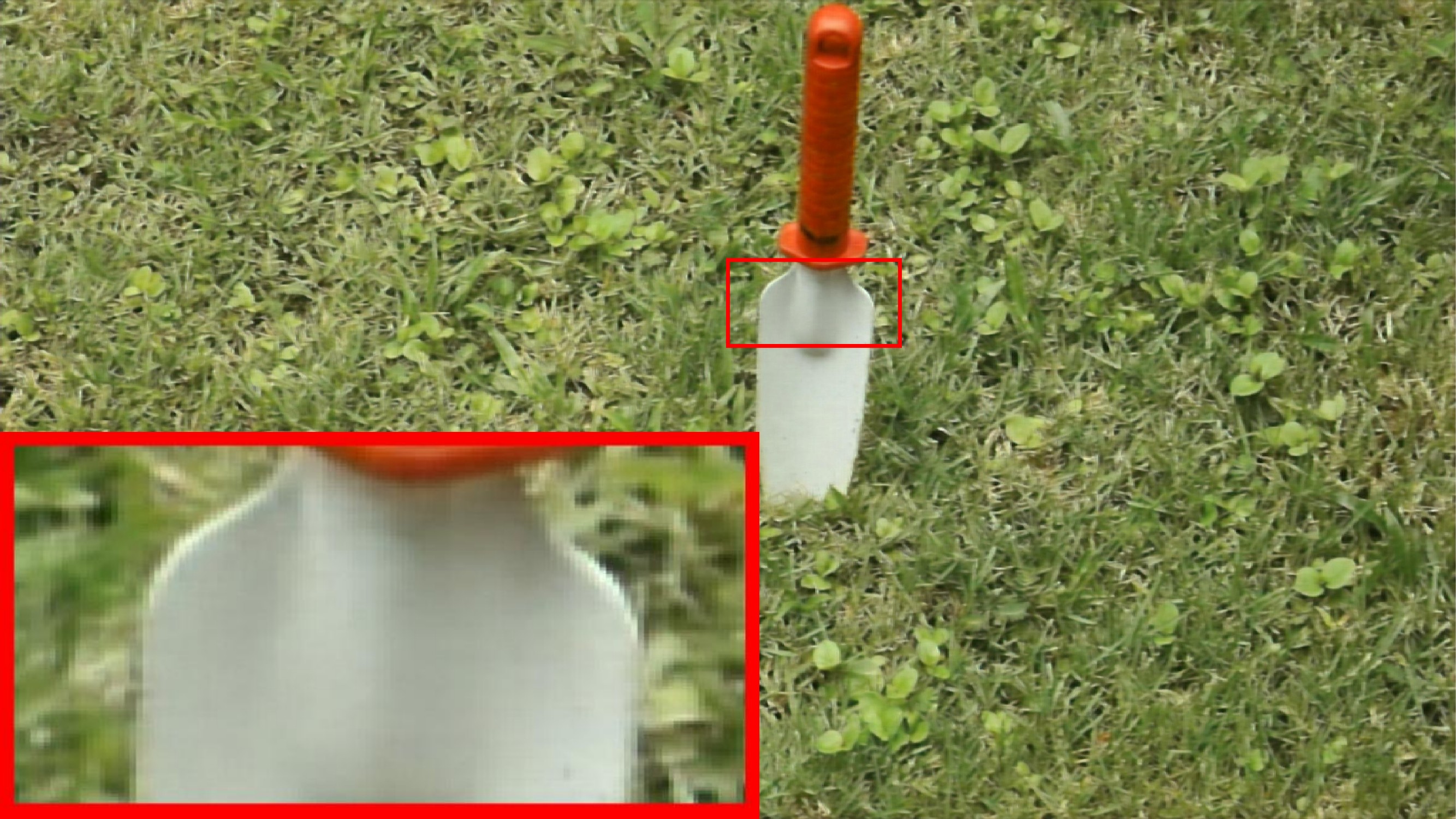}
        &
        \includegraphics[width=0.192\textwidth]
        {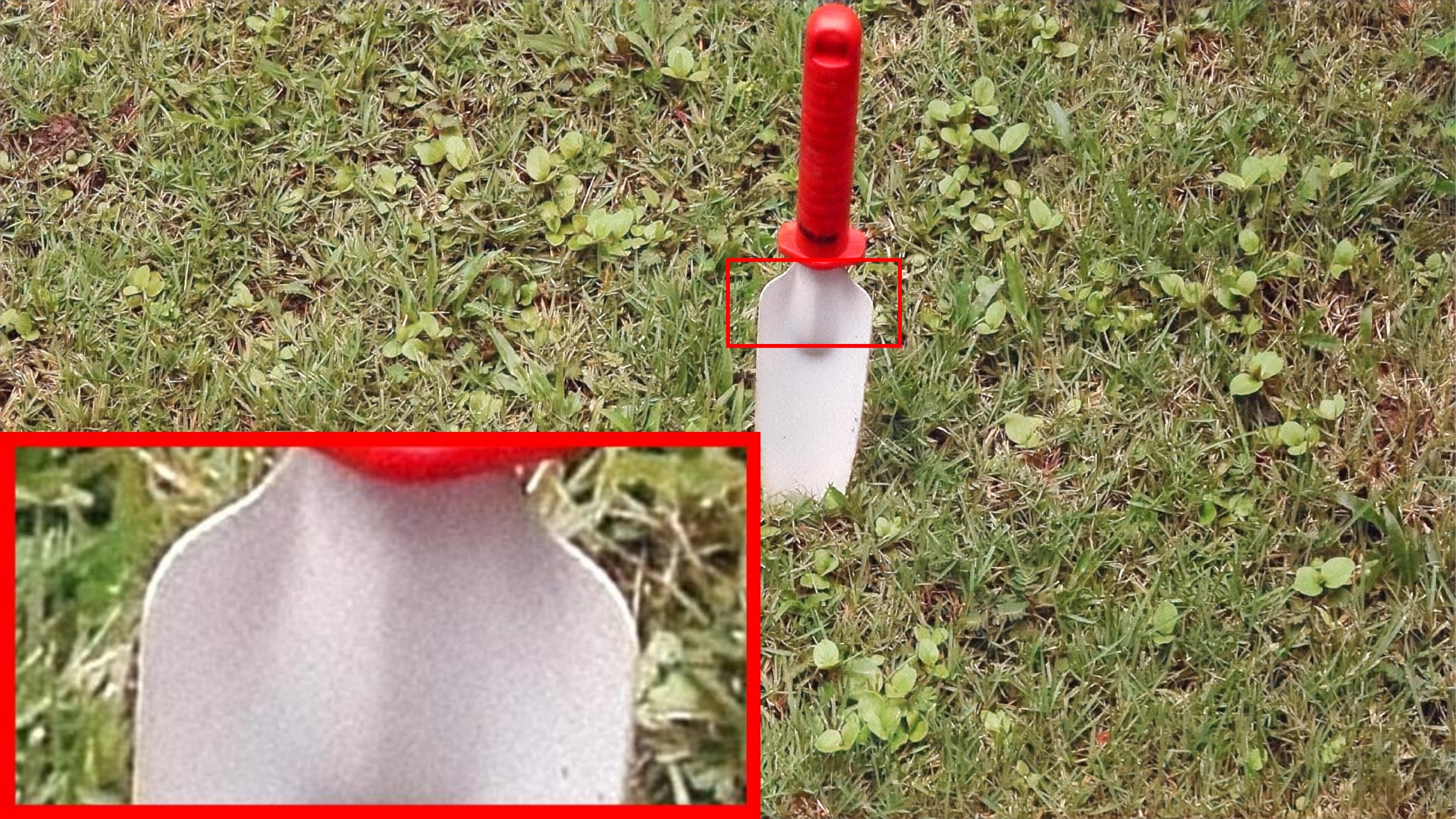}
        &
        \includegraphics[width=0.192\textwidth]
        {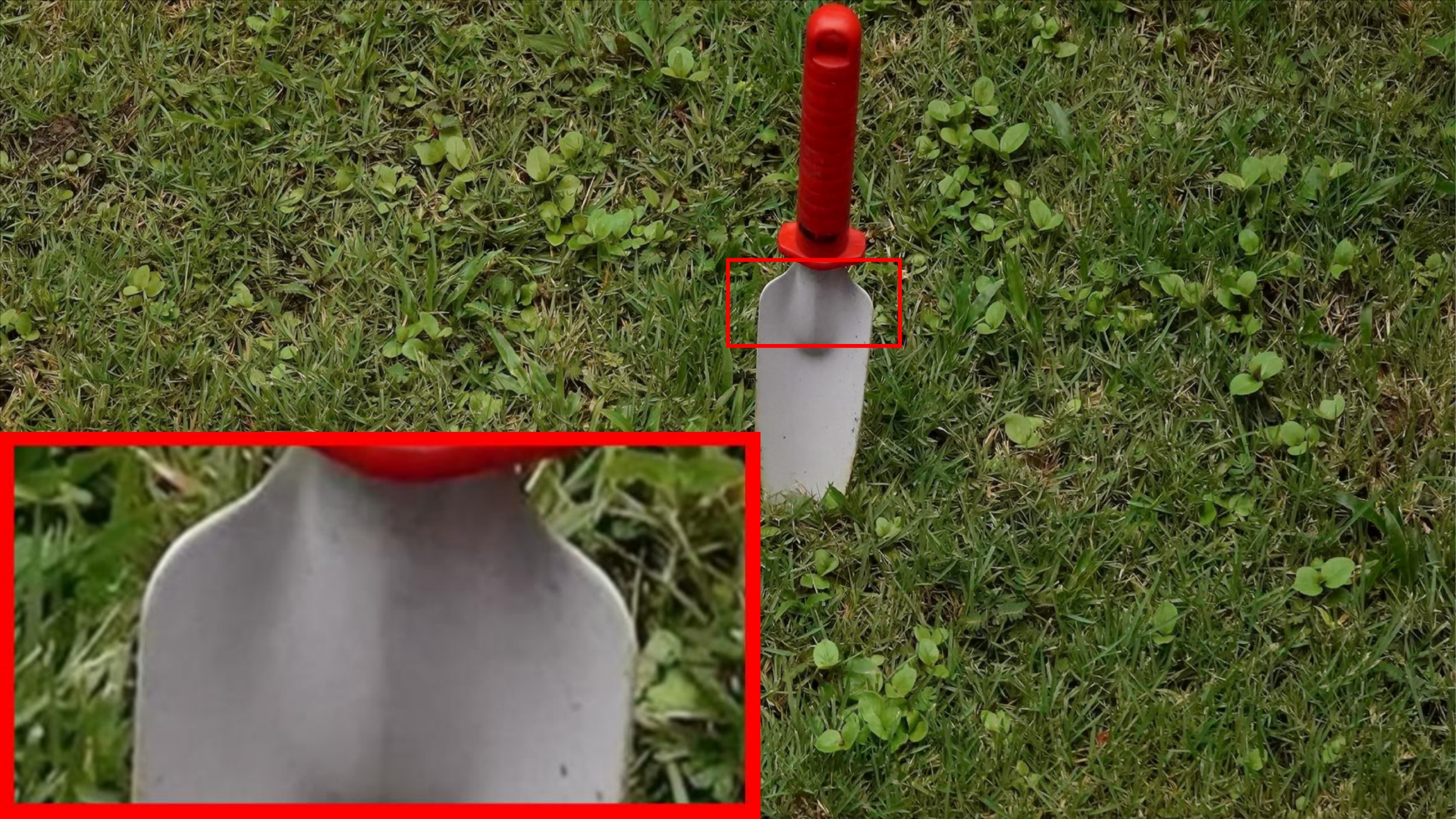}
        \\
        \includegraphics[width=0.192\textwidth]
        {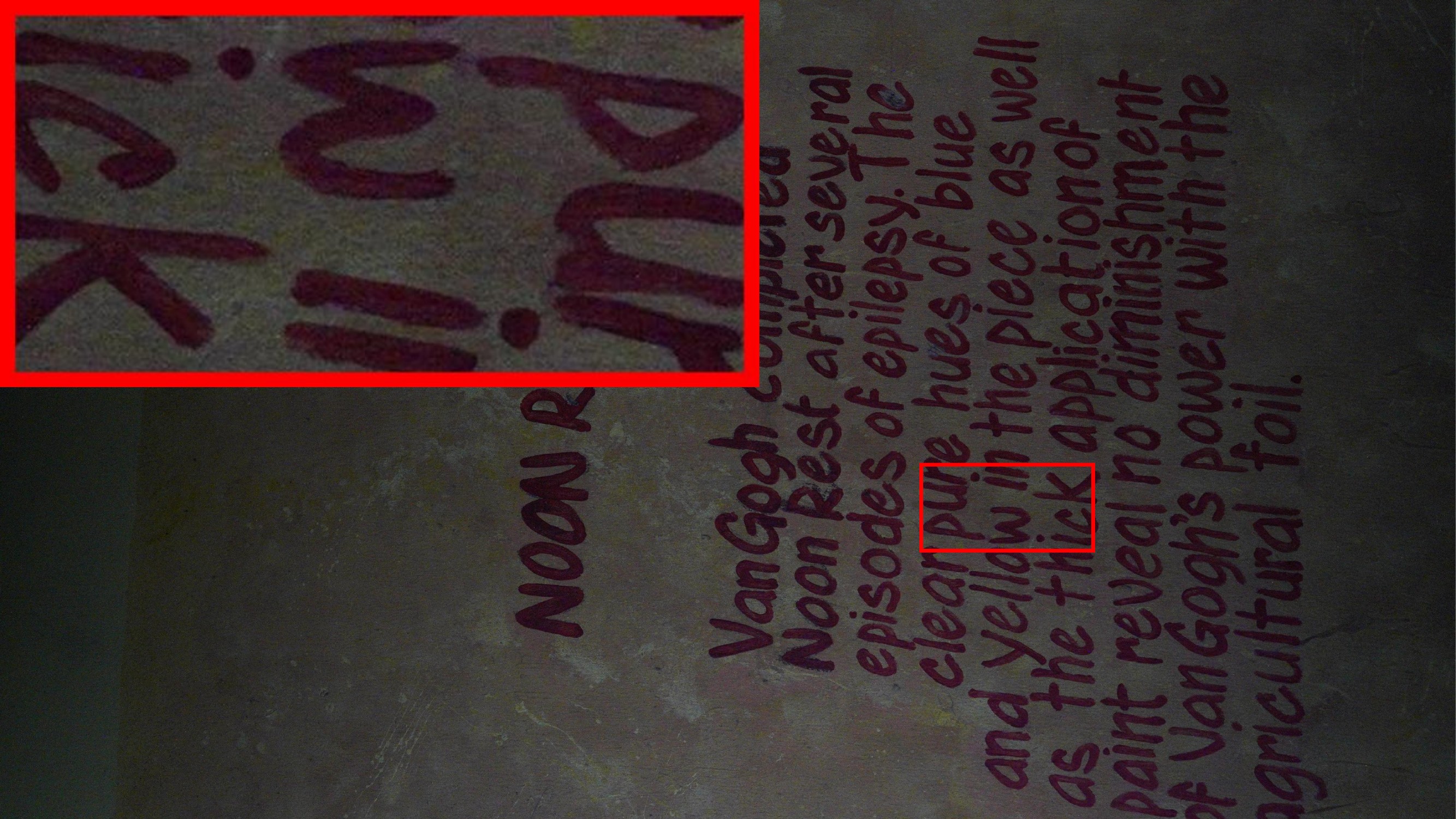}
        &
        \includegraphics[width=0.192\textwidth]
        {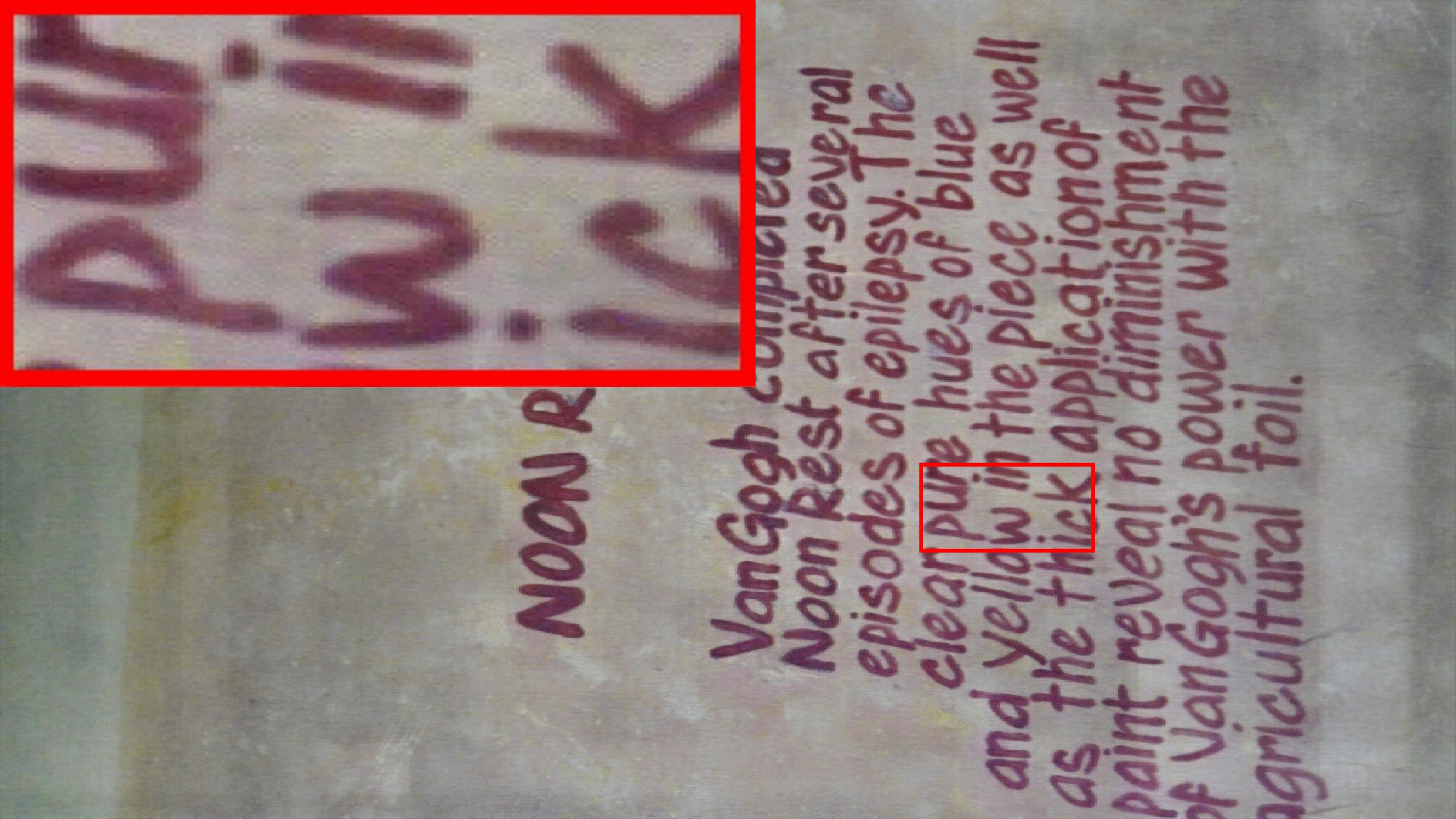}
        &
        \includegraphics[width=0.192\textwidth]
        {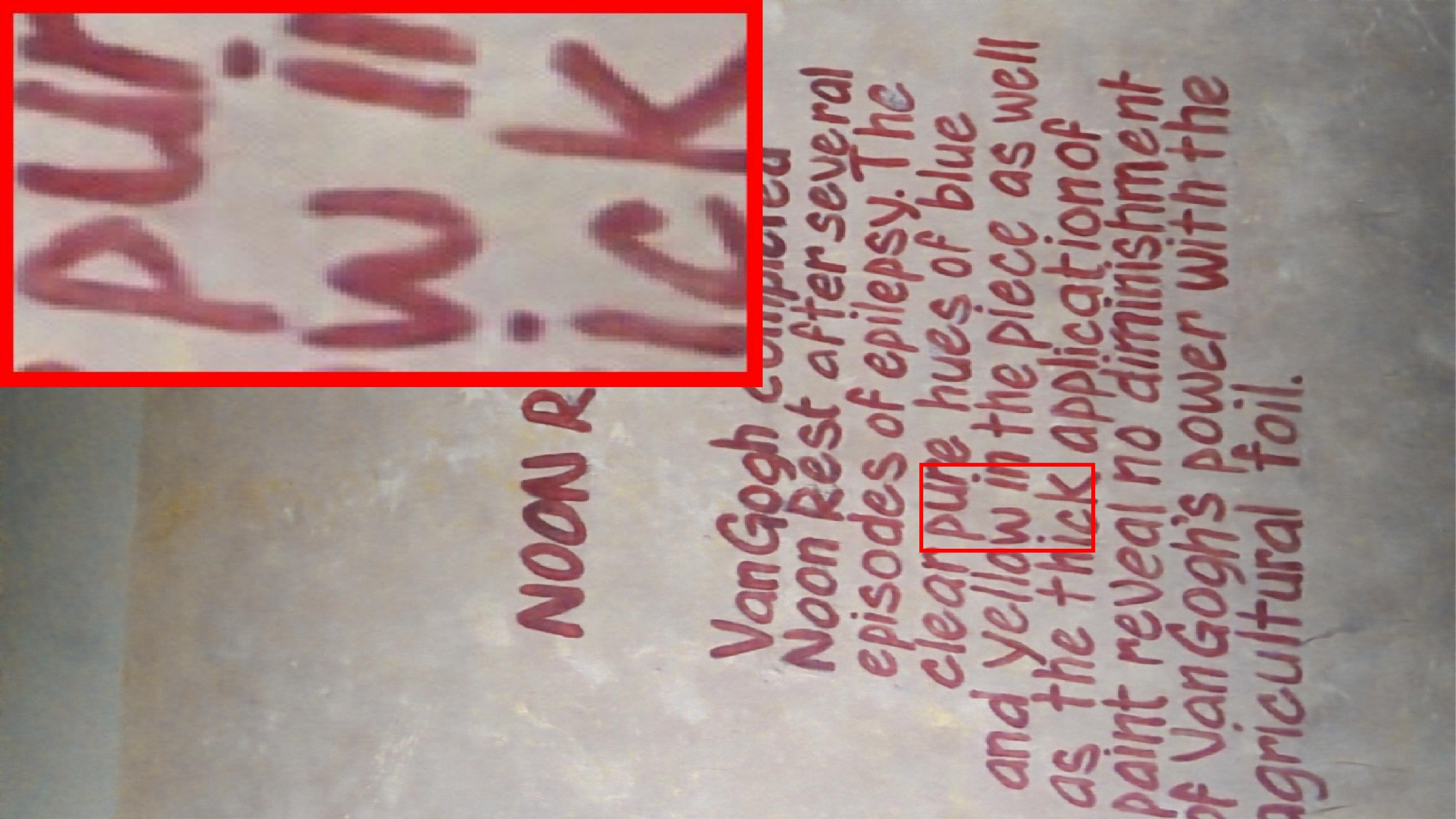}
        &
        \includegraphics[width=0.192\textwidth]
        {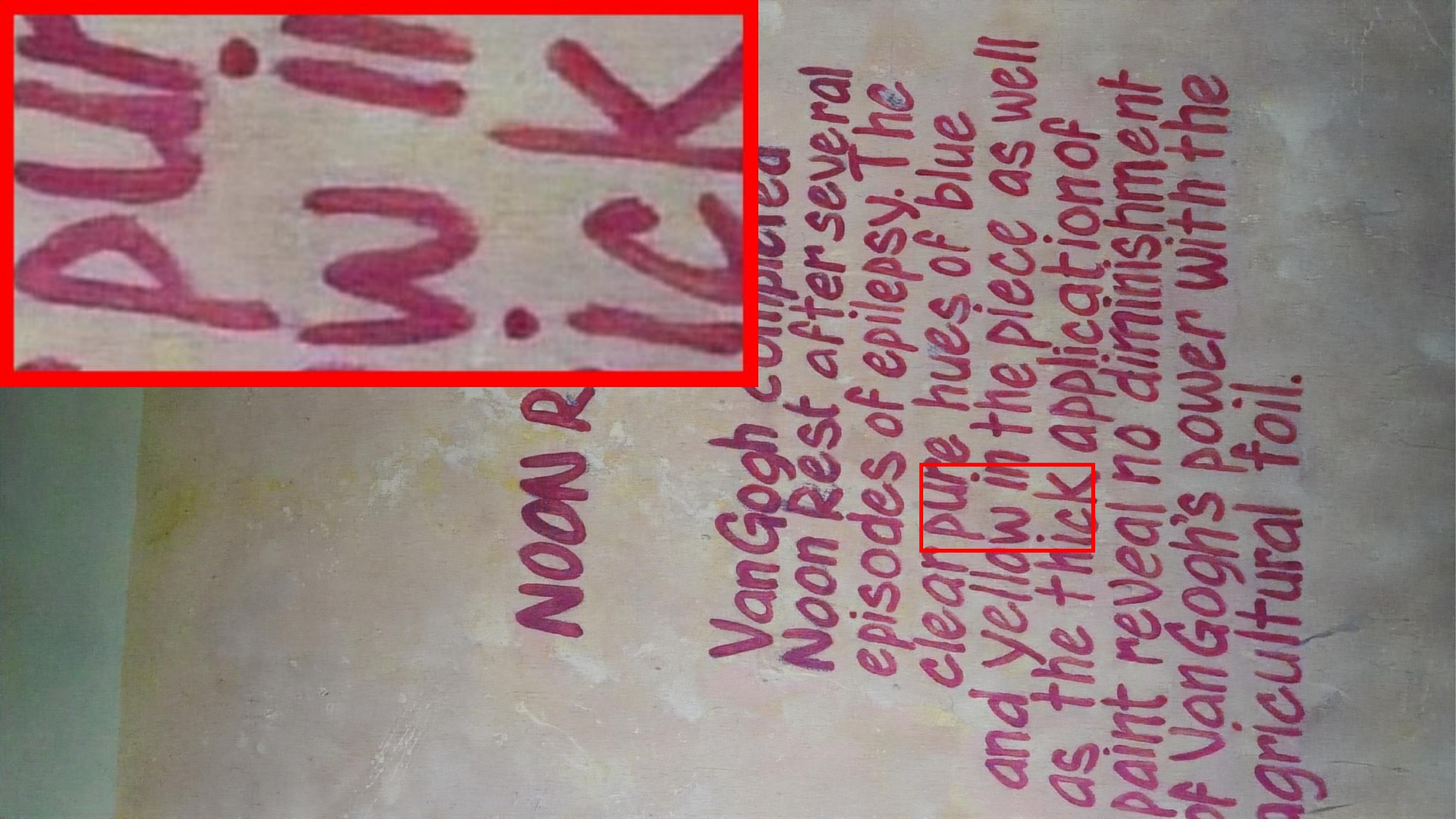}
        &
        \includegraphics[width=0.192\textwidth]
        {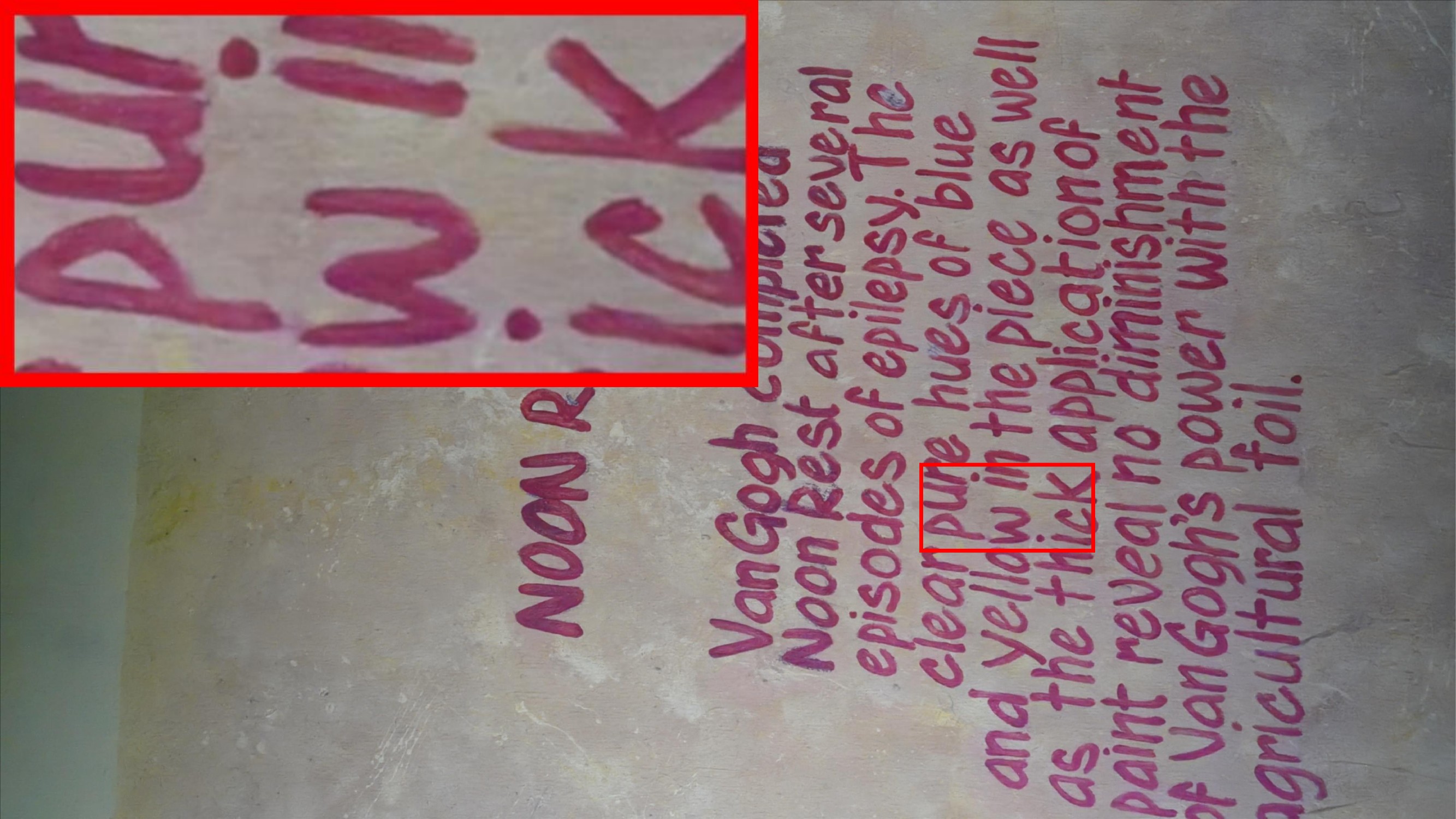}
        \\[-1mm]
        {\footnotesize (a) Input}
        &
        {\footnotesize (b) Uformer~\cite{wang2022uformer}}
        &
        {\footnotesize (c) Reti-Diff~\cite{he2025retidiff}}
        &
        {\footnotesize (d) UHDFour~\cite{li2023uhdfour}}
        &
        {\footnotesize (e) Wave-Mamba~\cite{zou2024wavemamba}}
        \\[1mm]

        \includegraphics[width=0.192\textwidth]
        {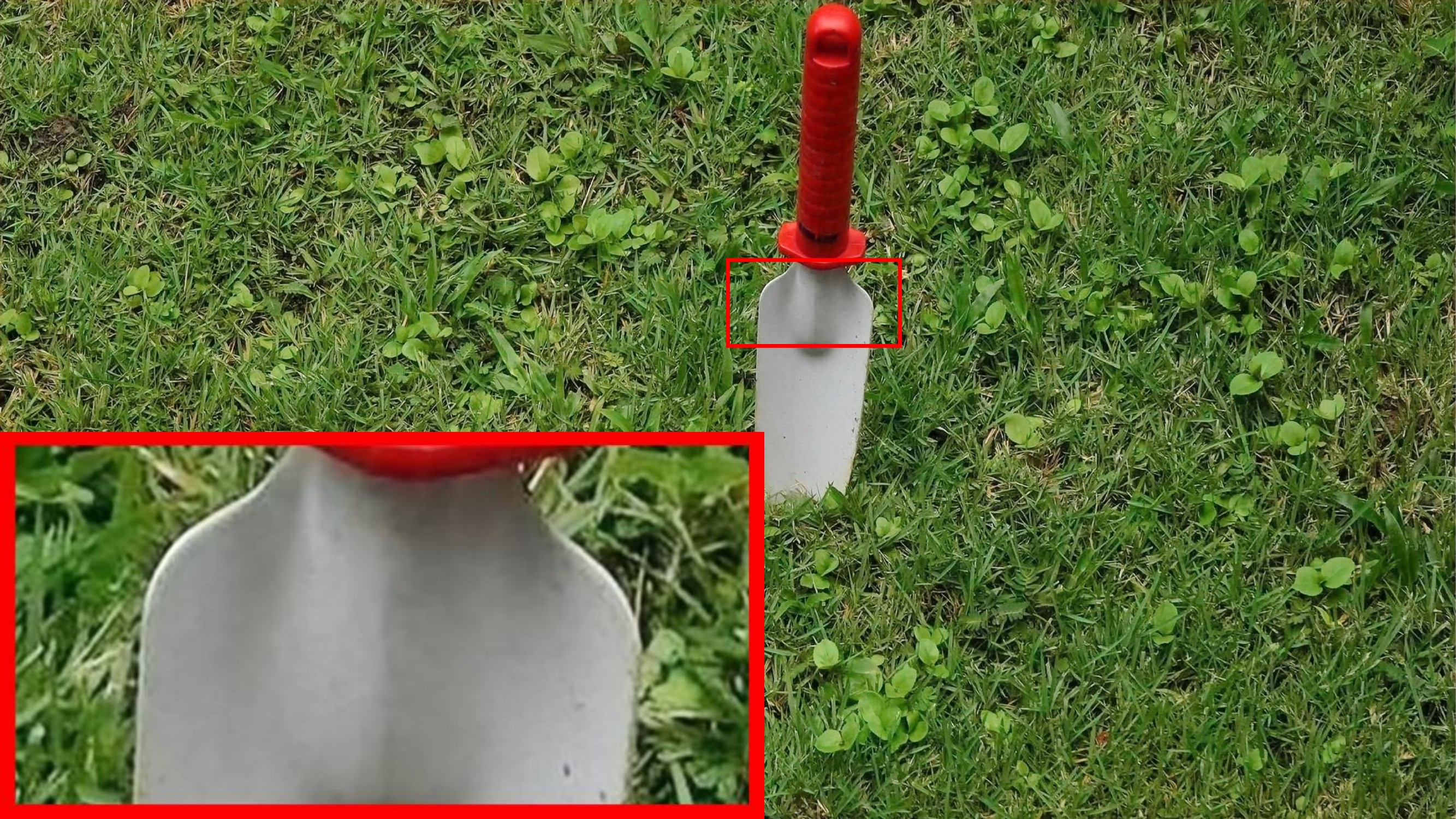}
        &
        \includegraphics[width=0.192\textwidth]
        {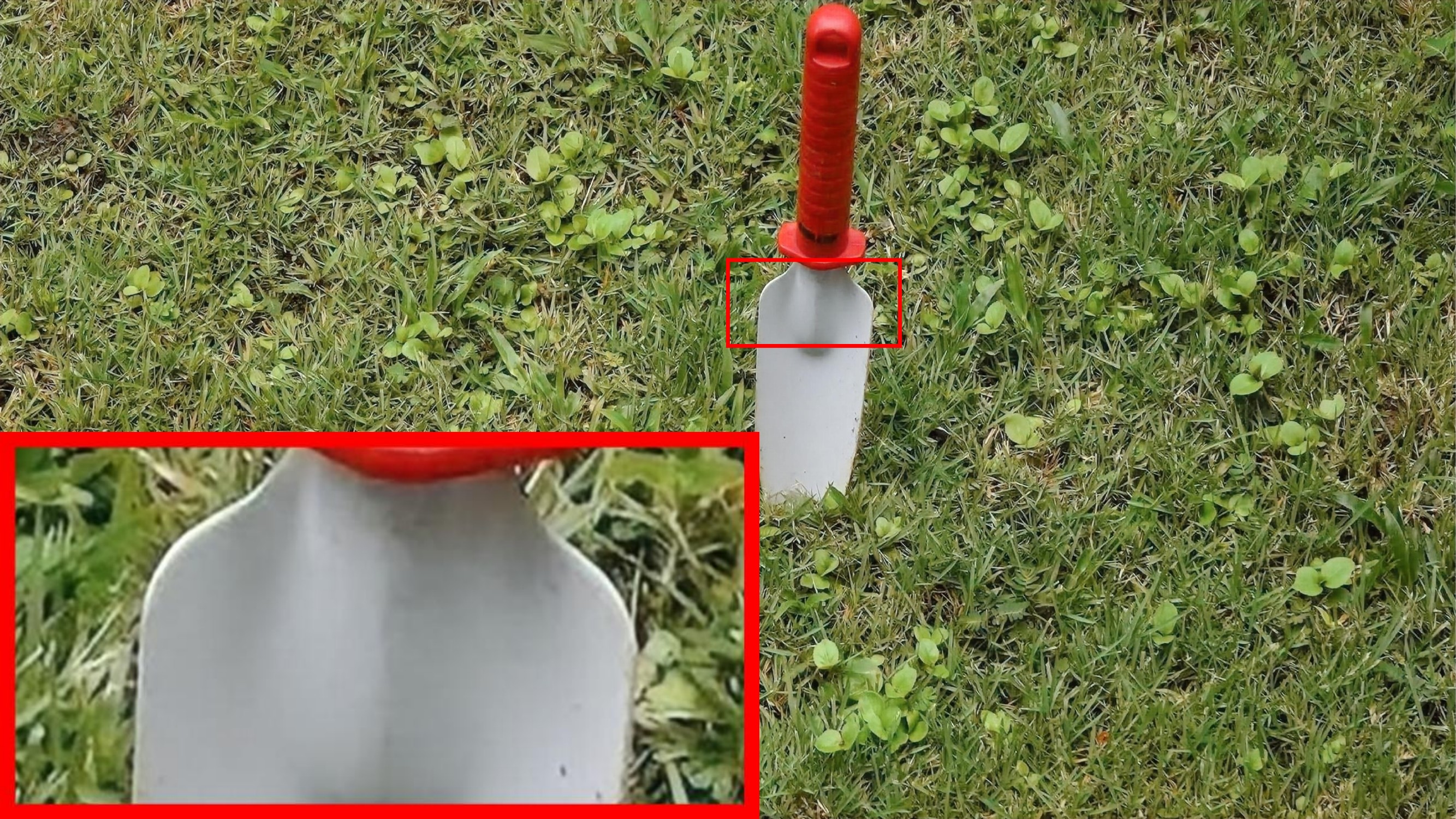}
        &
        \includegraphics[width=0.192\textwidth]
        {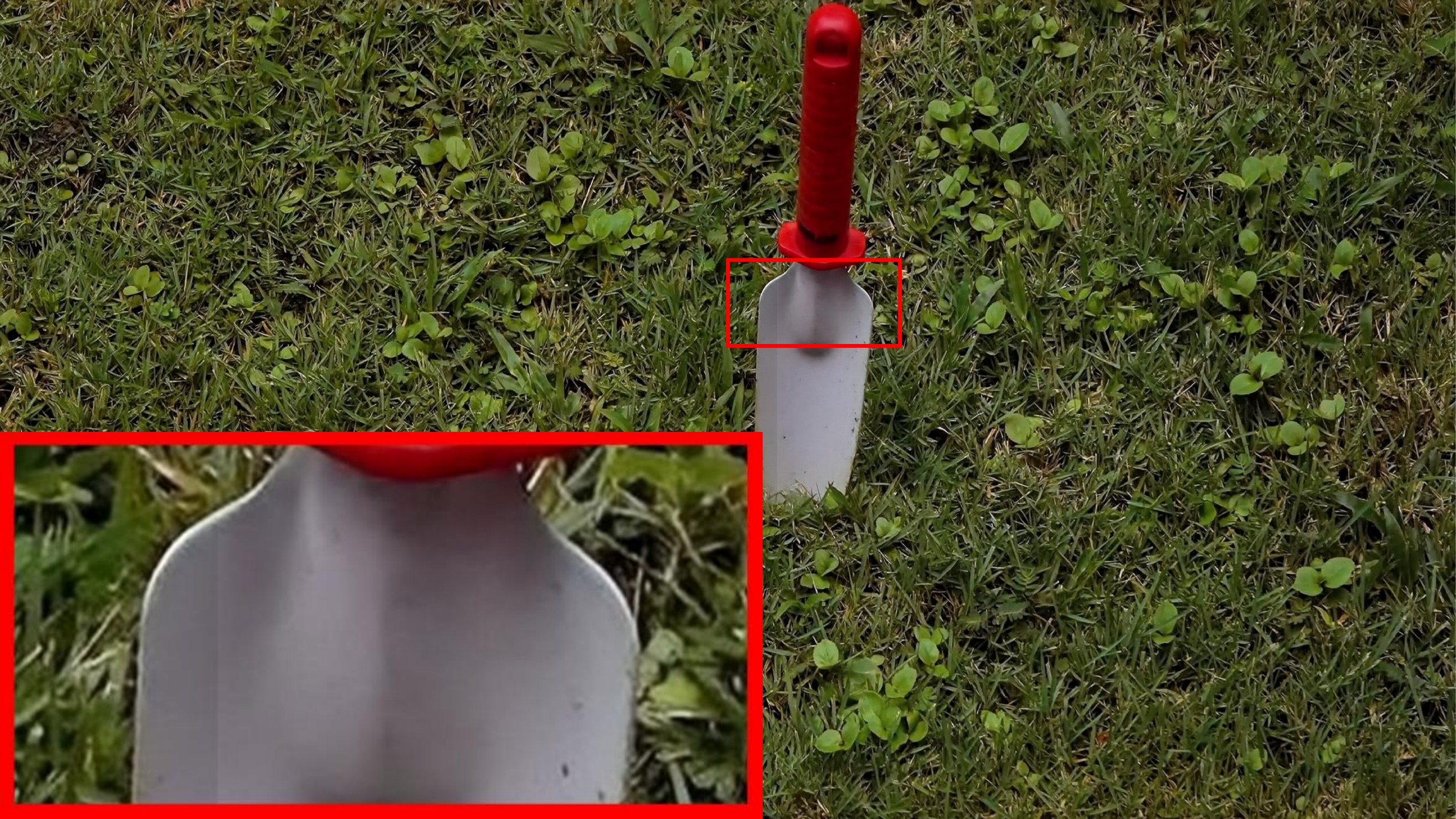}
        &
        \includegraphics[width=0.192\textwidth]
        {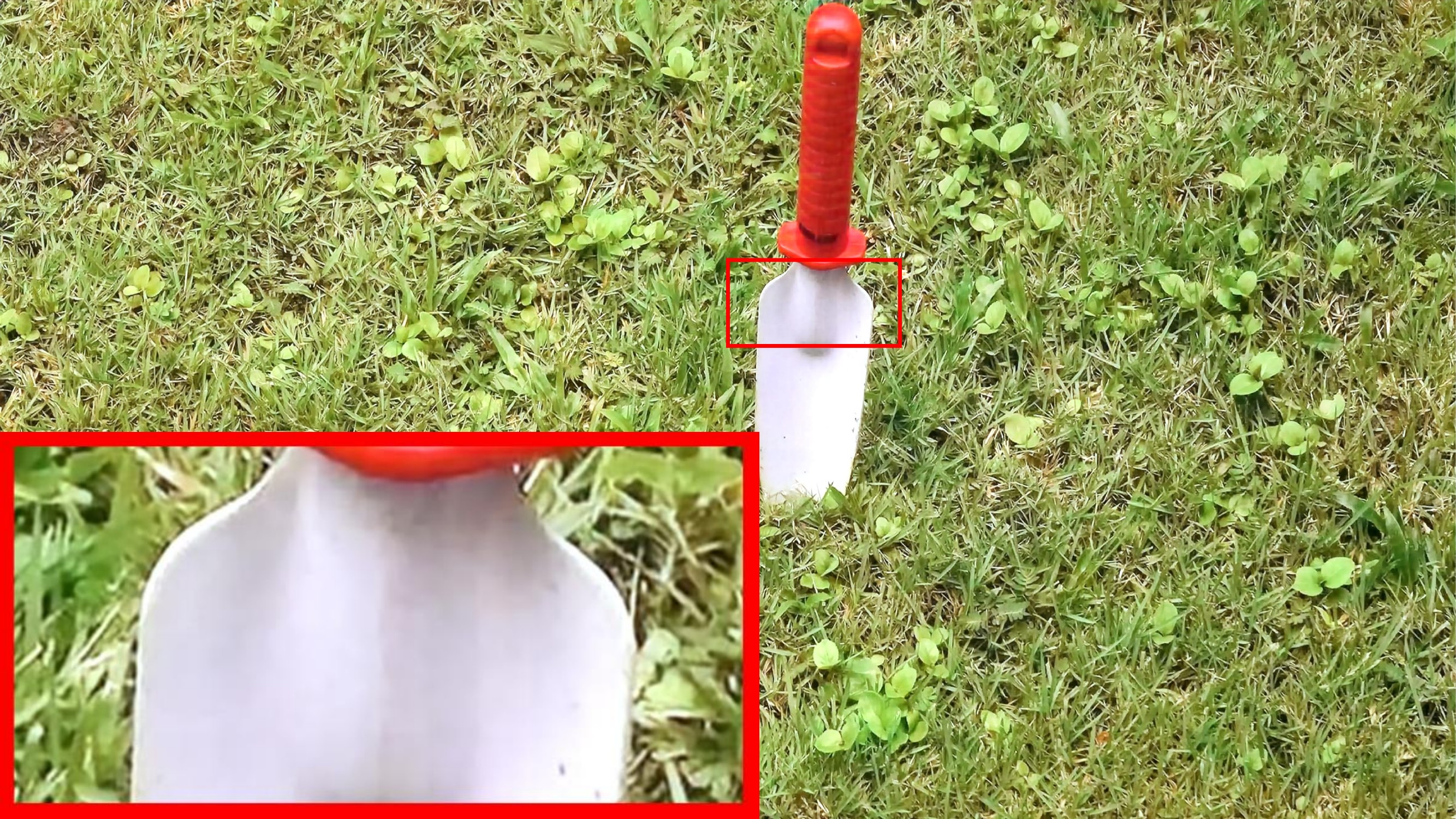}
        &
        \includegraphics[width=0.192\textwidth]
        {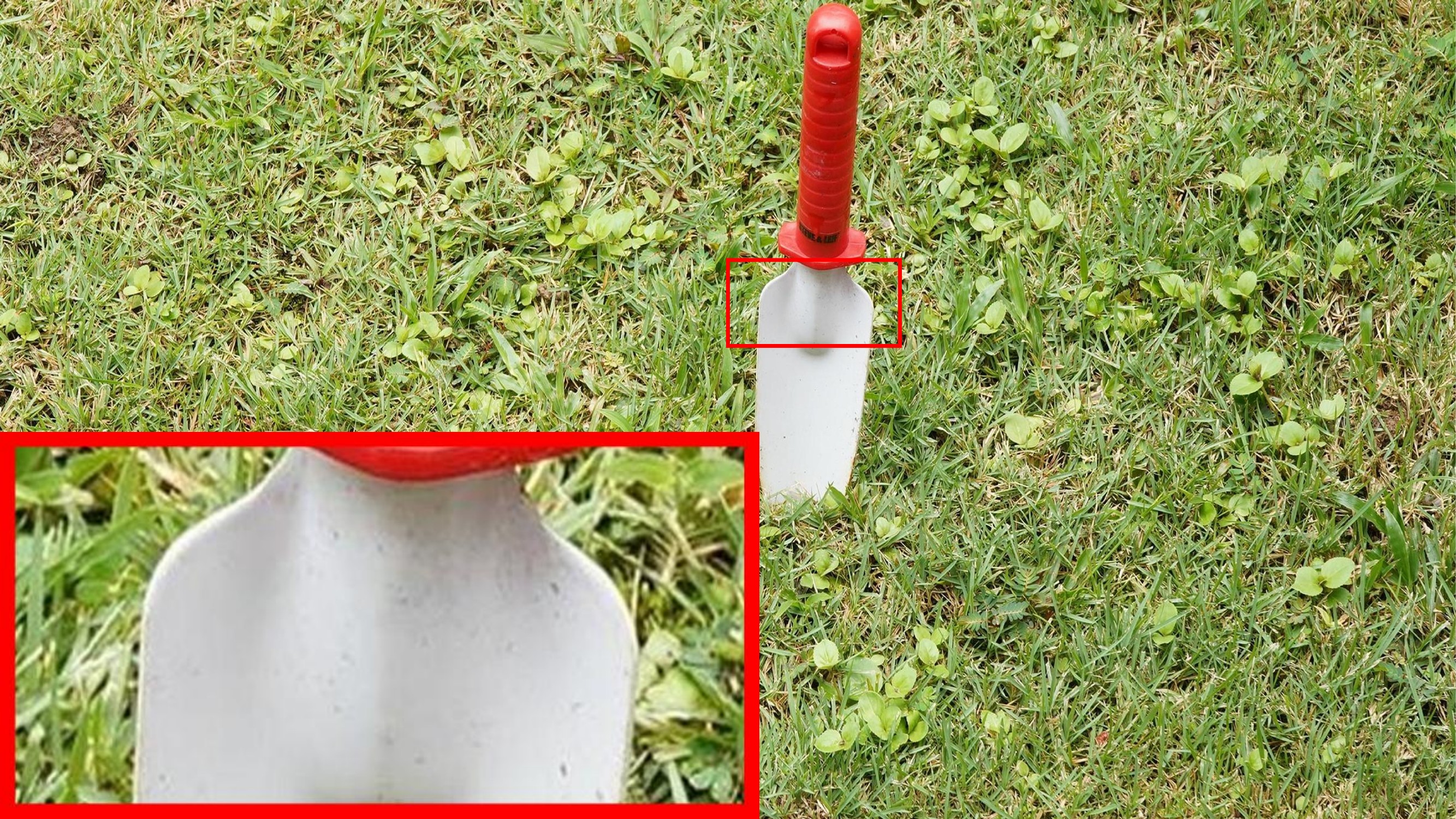}
        \\
        \includegraphics[width=0.192\textwidth]
        {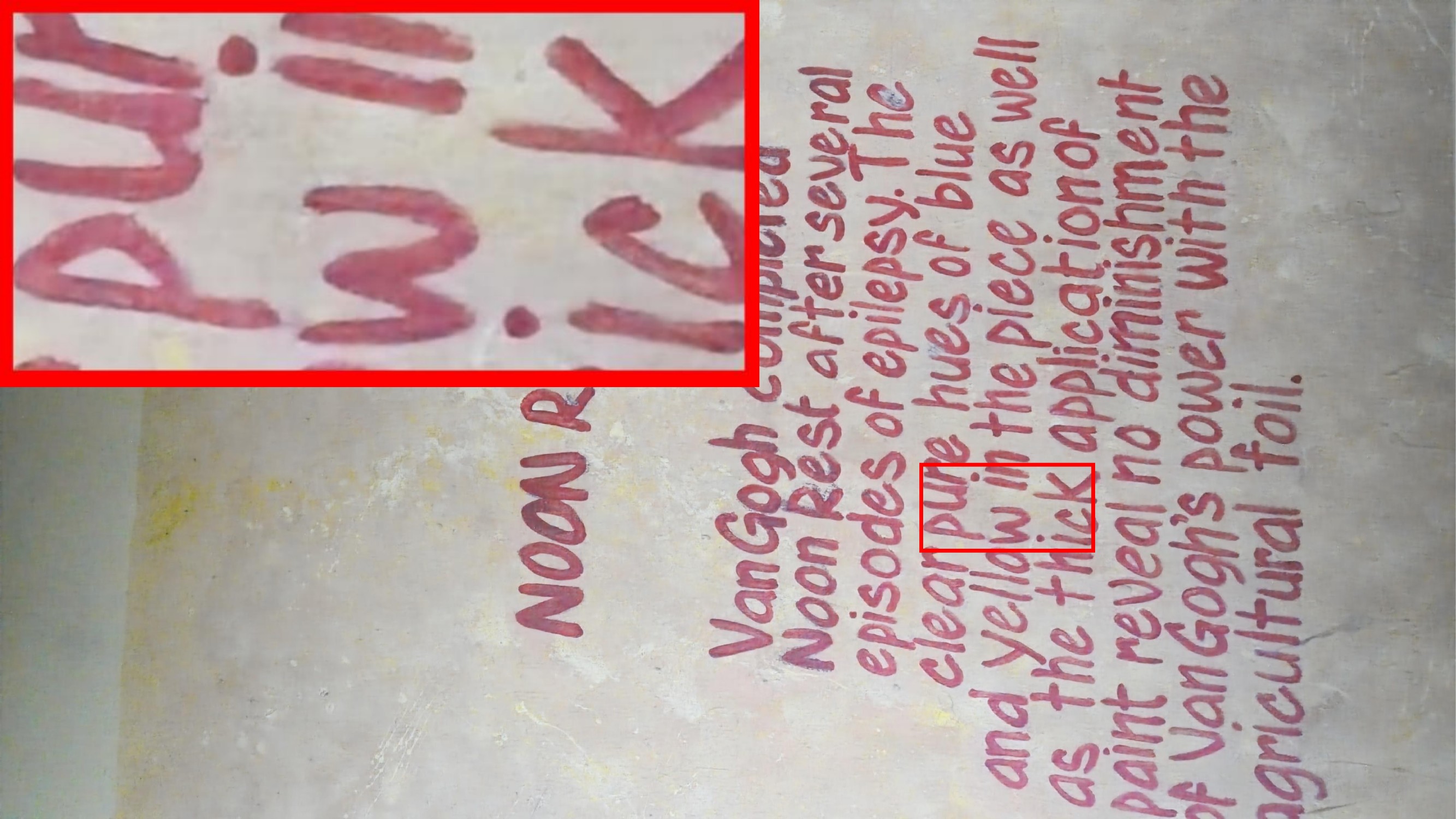}
        &
        \includegraphics[width=0.192\textwidth]
        {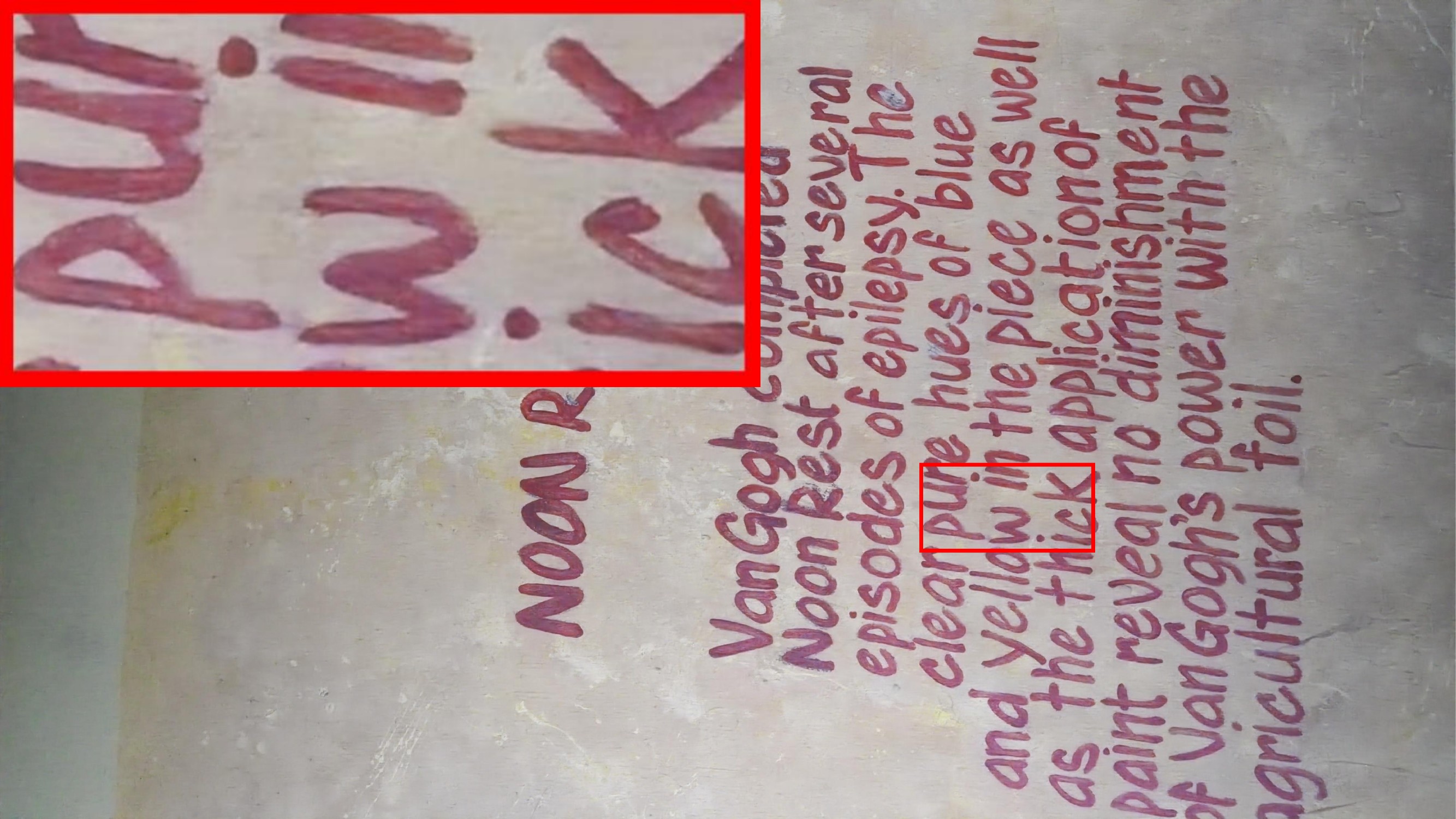}
        &
        \includegraphics[width=0.192\textwidth]
        {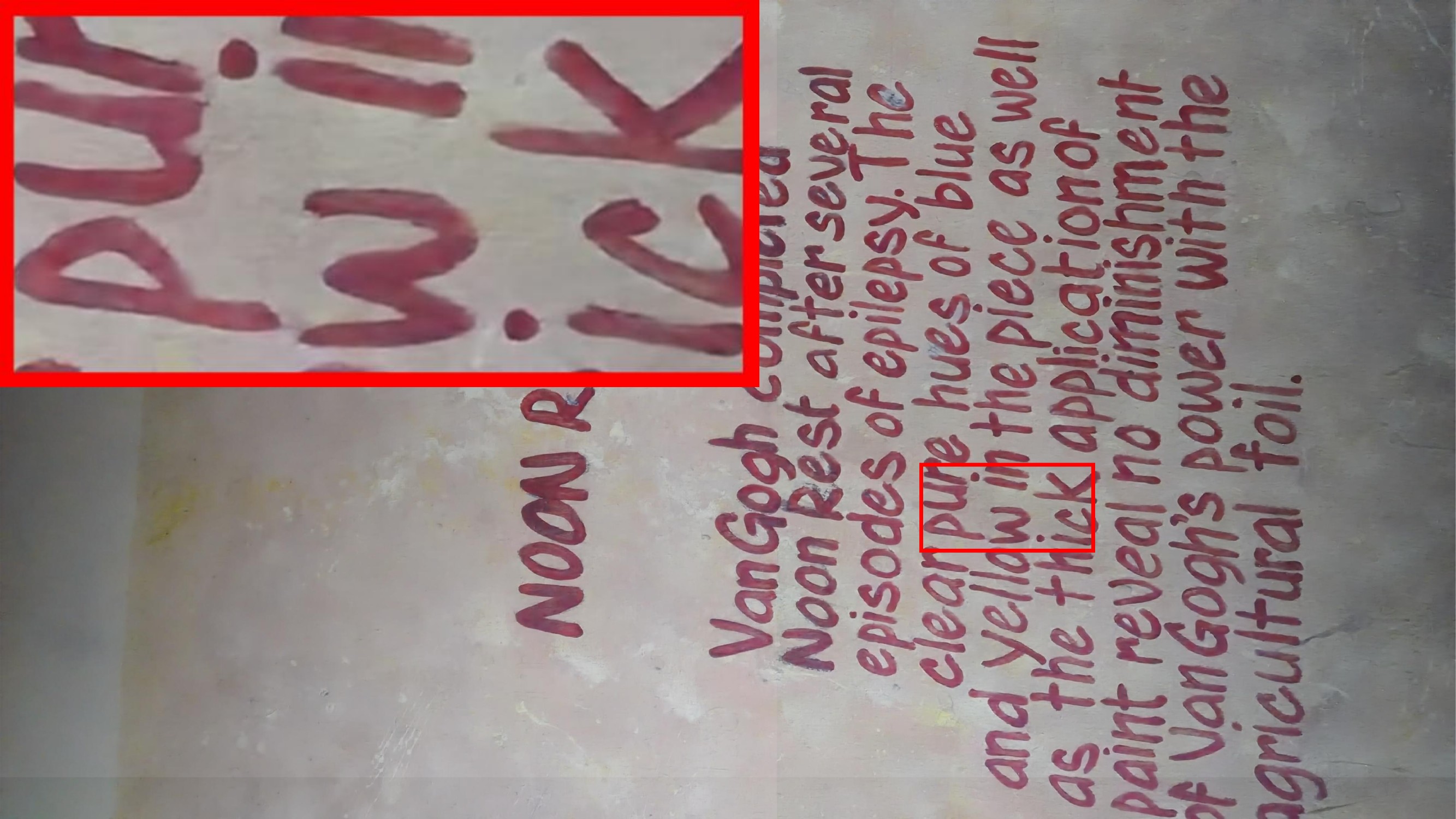}
        &
        \includegraphics[width=0.192\textwidth]
        {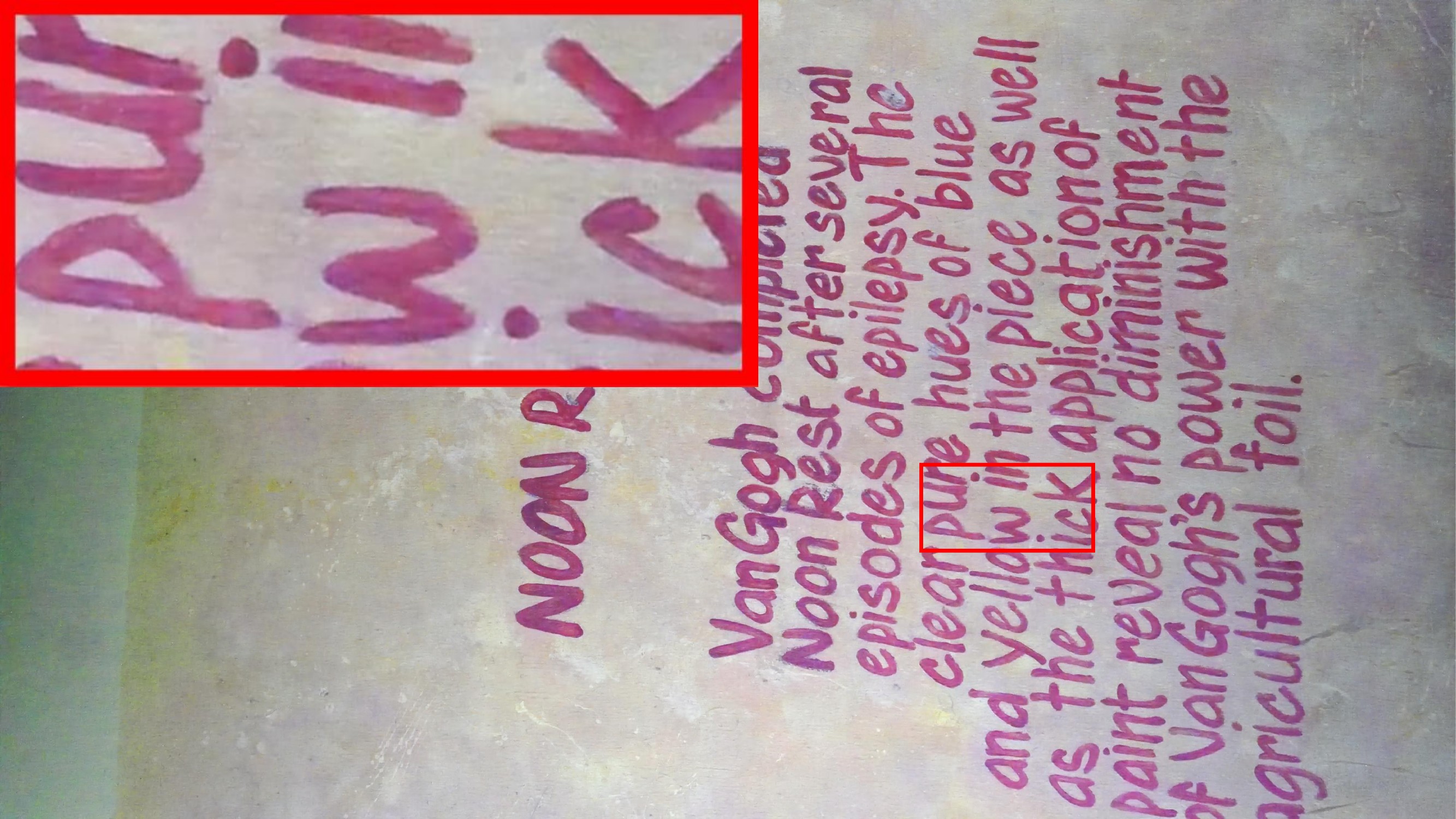}
        &
        \includegraphics[width=0.192\textwidth]
        {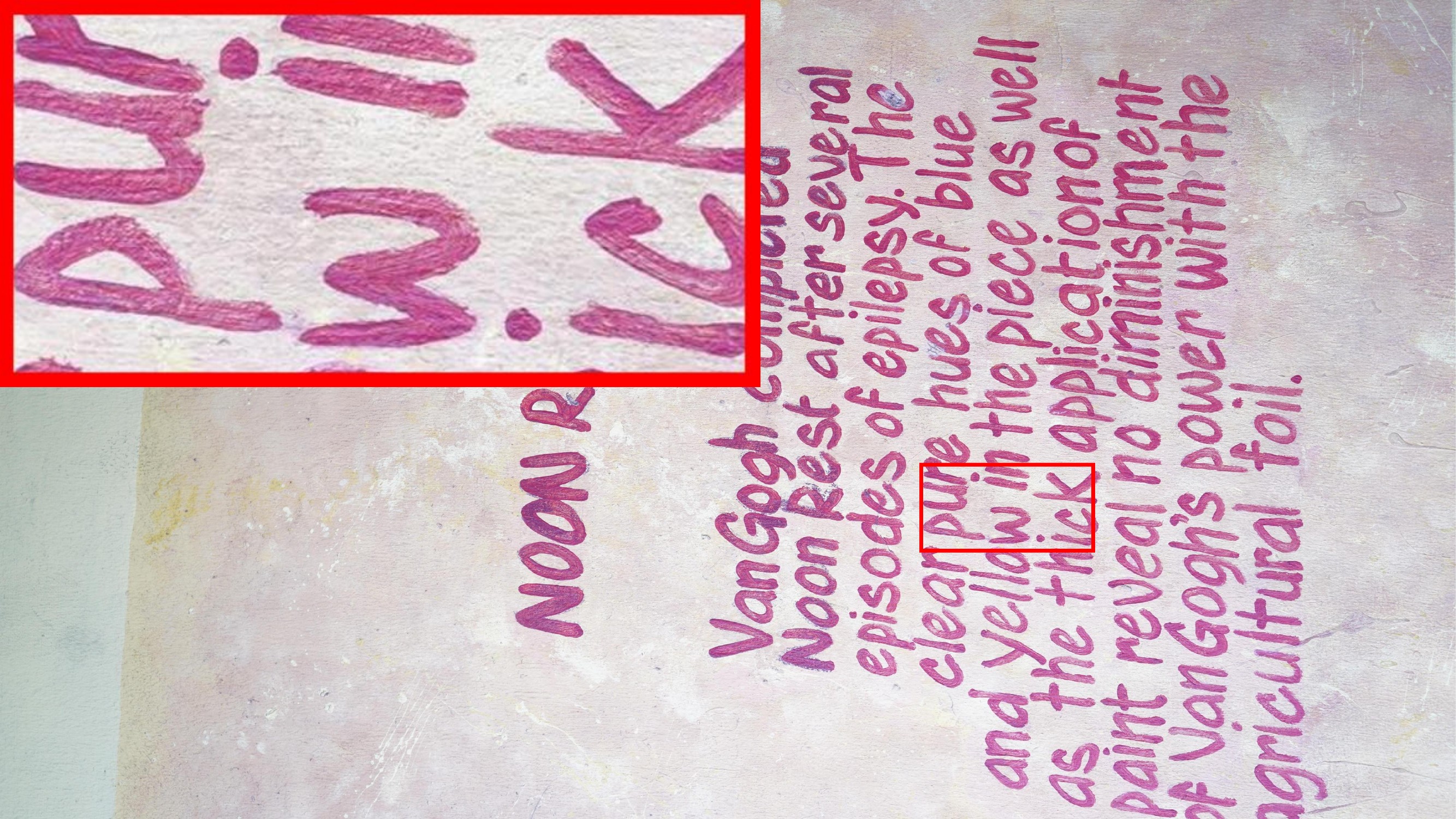}
        \\[-1mm]

        {\footnotesize (f) UHDFormer~\cite{wang2024uhdformer}}
        &
        {\footnotesize (g) UHDDIP~\cite{Wang2026uhddip}}
        &
        {\footnotesize (h) UHD-Processor~\cite{liu2025uhdprocessor}}
        &
        {\footnotesize \textbf{(i) BiFreqNet}}
        &
        {\footnotesize (j) GT}
        \\

    \end{tabular}
    \endgroup
    \resultcaption{
        Visual comparison of representative UHD low-light image enhancement
        methods on the UHD-LL dataset~\cite{li2023uhdfour}. BiFreqNet produces more visually pleasing results with more faithful illumination and 
        clearer details.
    }
    \label{fig:vis_uhdll}
\end{figure*}
\begin{figure*}[!t]
    \centering
    \begingroup
    \setlength{\tabcolsep}{2.5pt}
    \renewcommand{\arraystretch}{1.0}

    \begin{tabular}{@{}ccccc@{}}
        \includegraphics[width=0.192\textwidth]
        {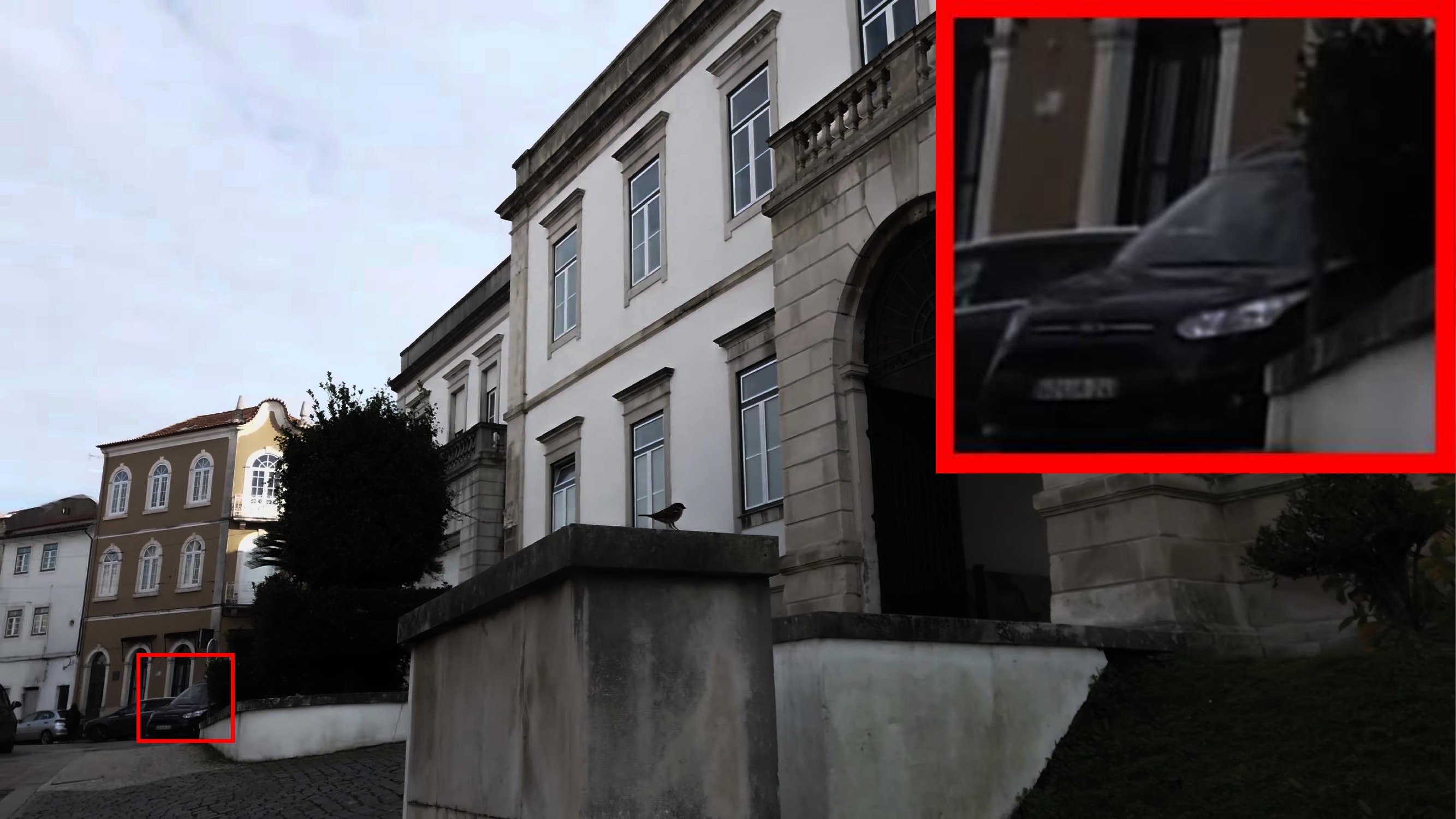}
        &
        \includegraphics[width=0.192\textwidth]
        {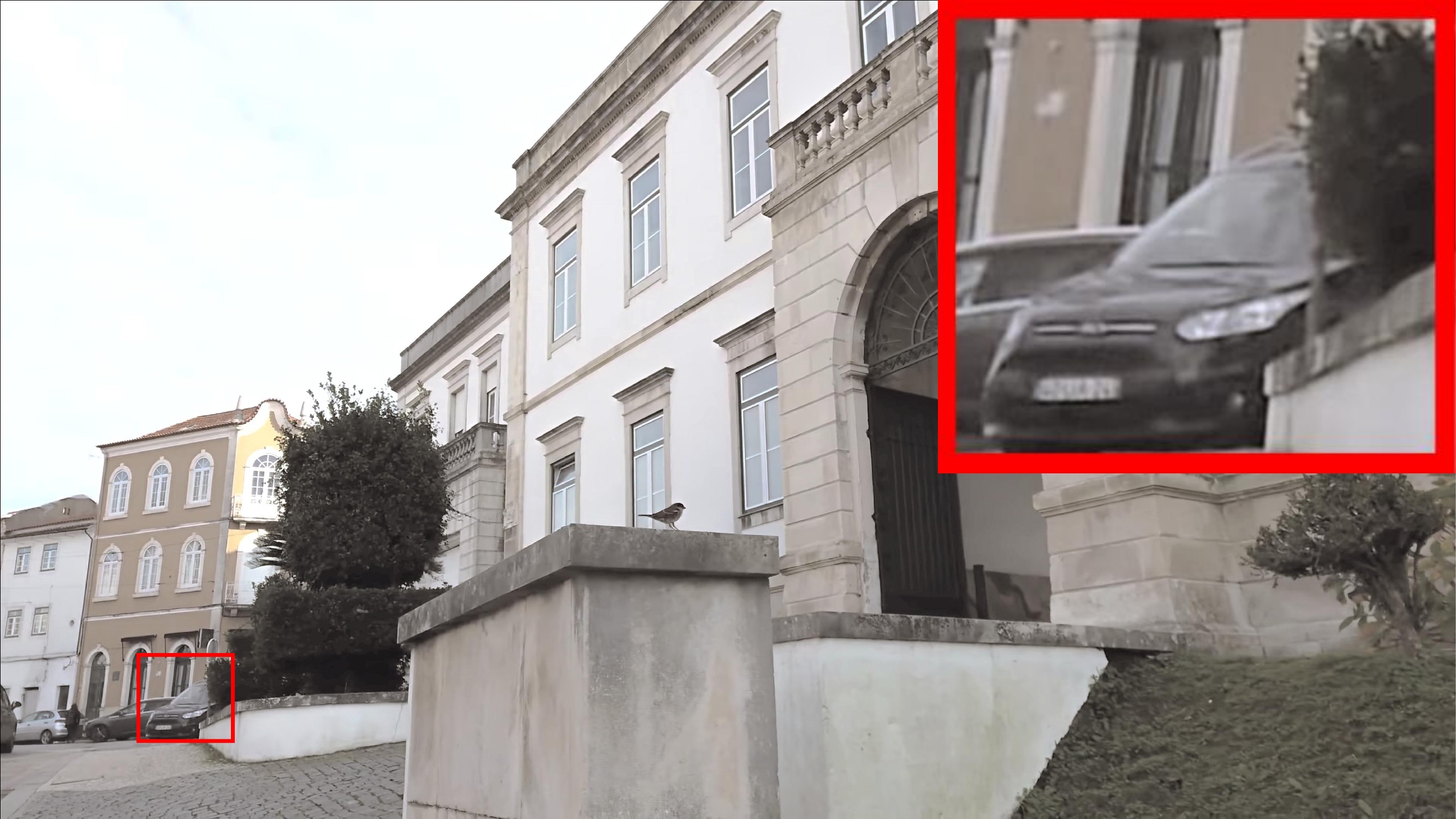}
        &
        \includegraphics[width=0.192\textwidth]
        {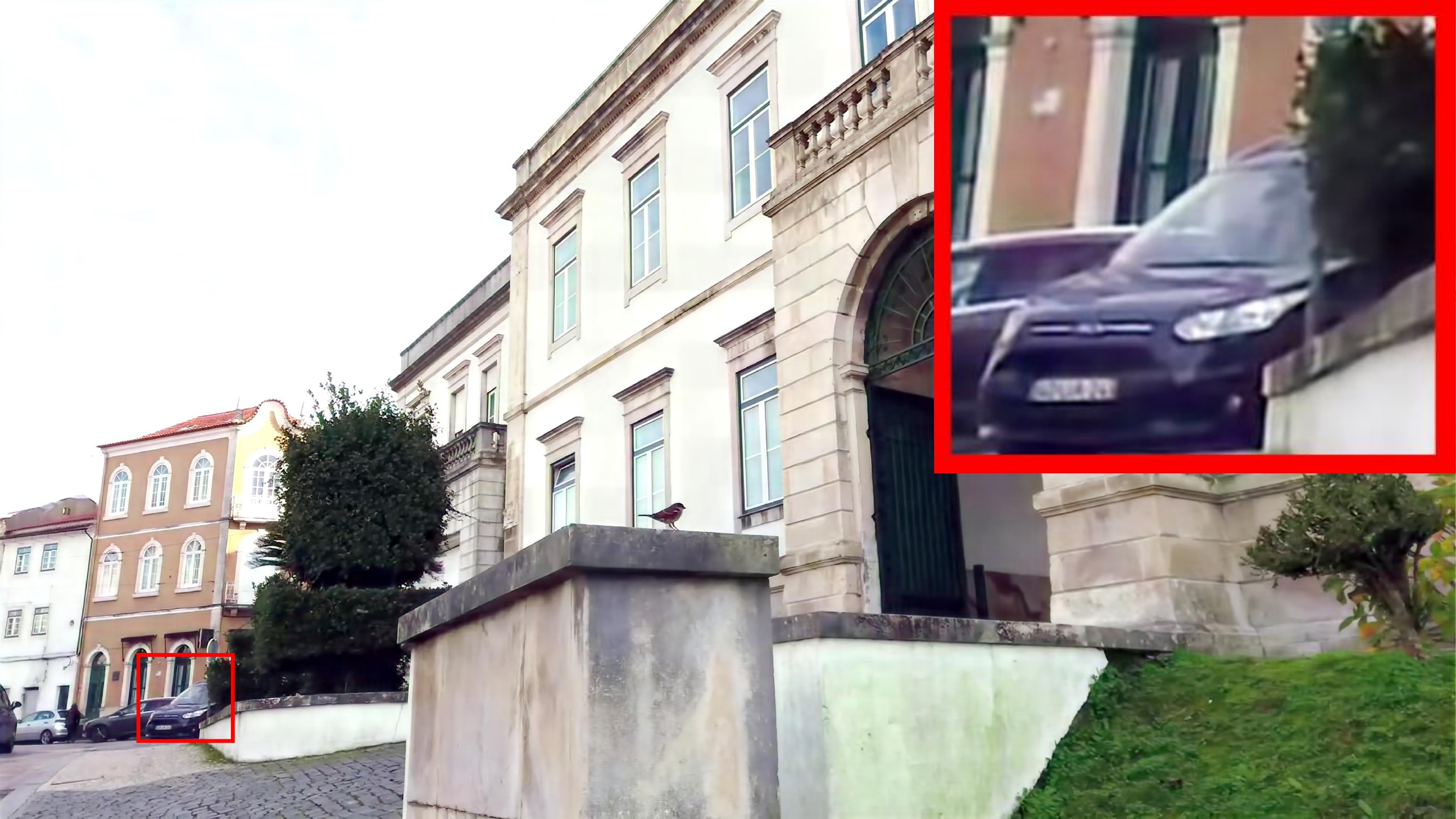}
        &
        \includegraphics[width=0.192\textwidth]
        {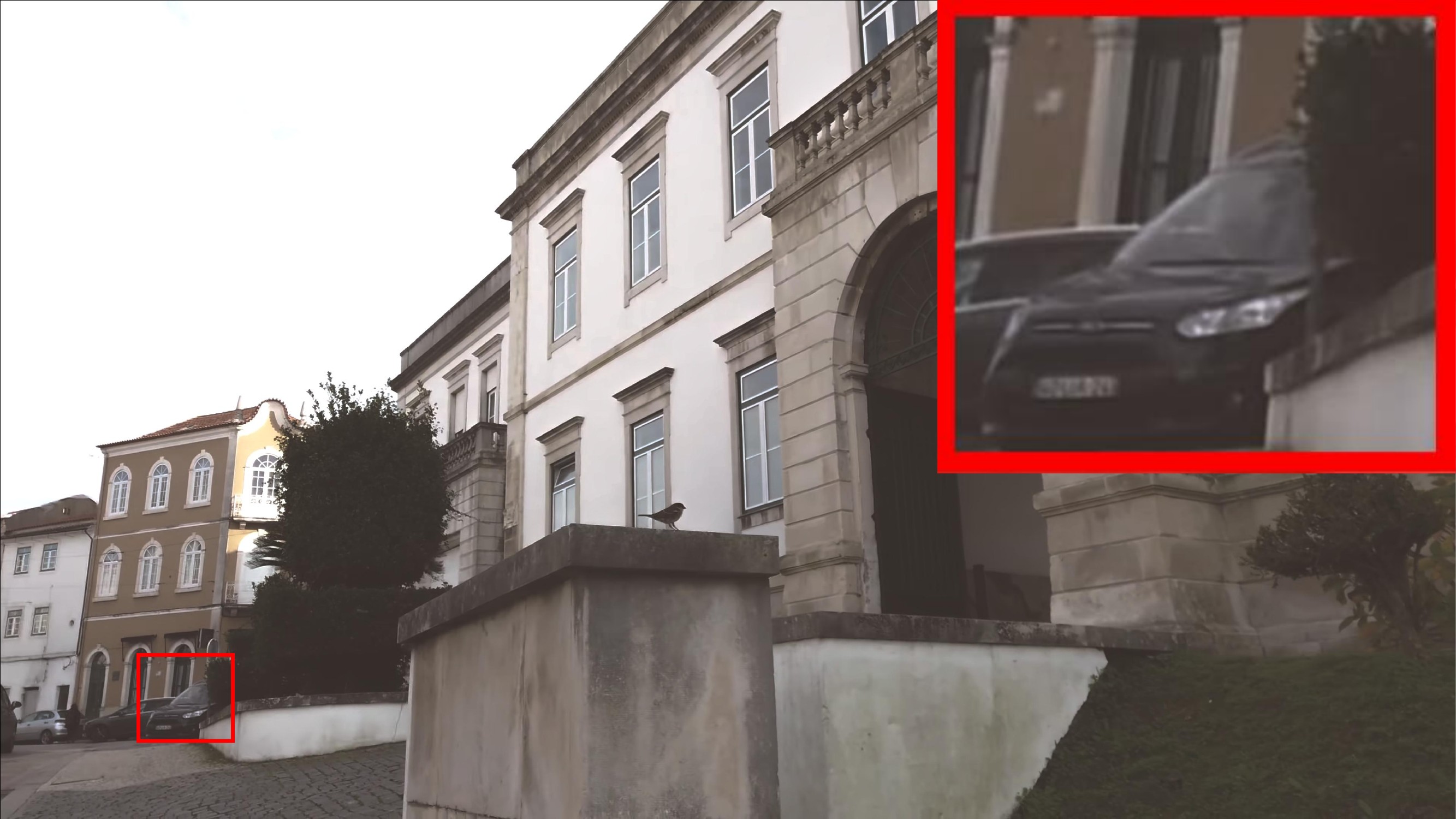}
        &
        \includegraphics[width=0.192\textwidth]
        {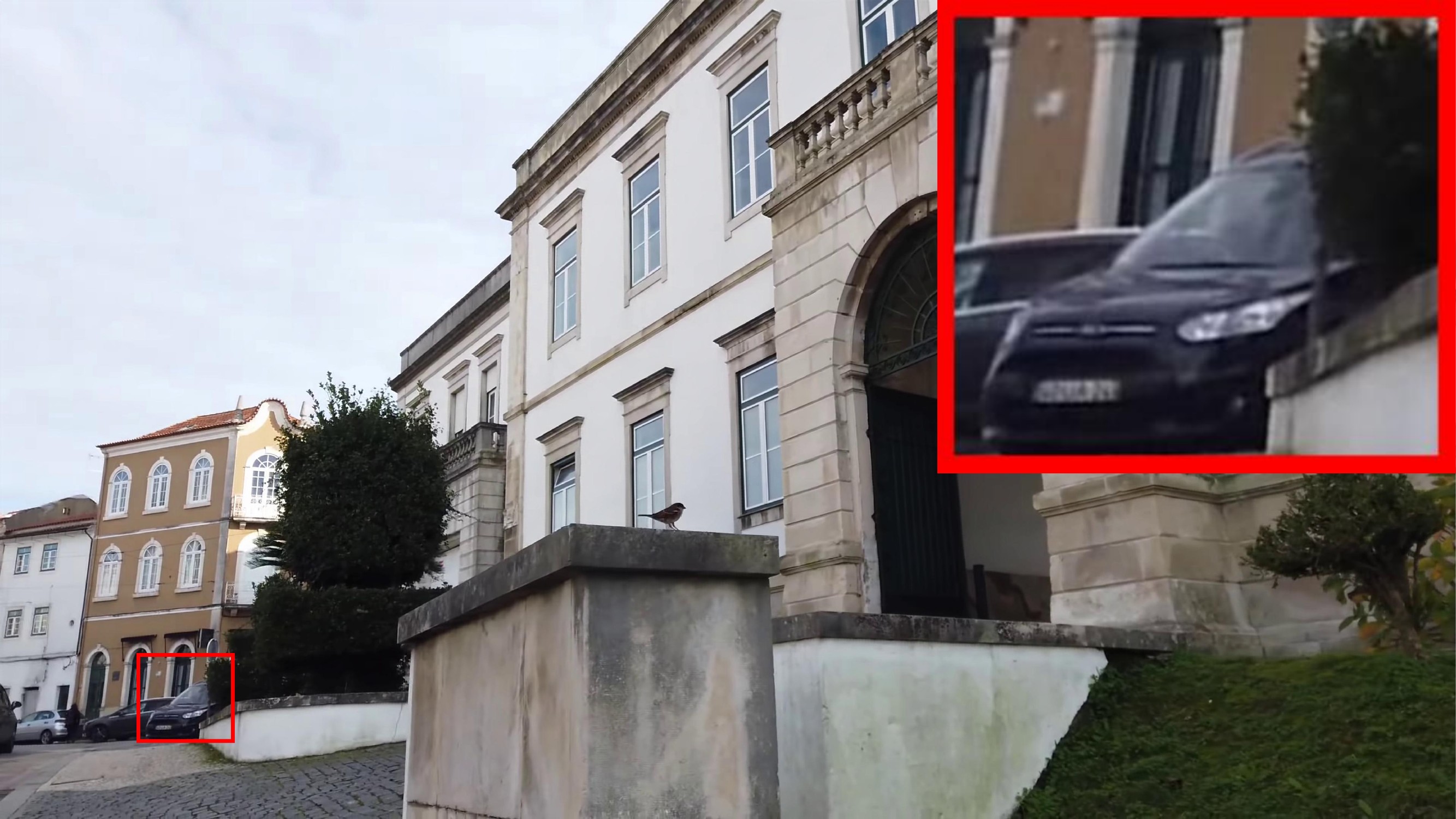}
        \\
        \includegraphics[width=0.192\textwidth]
        {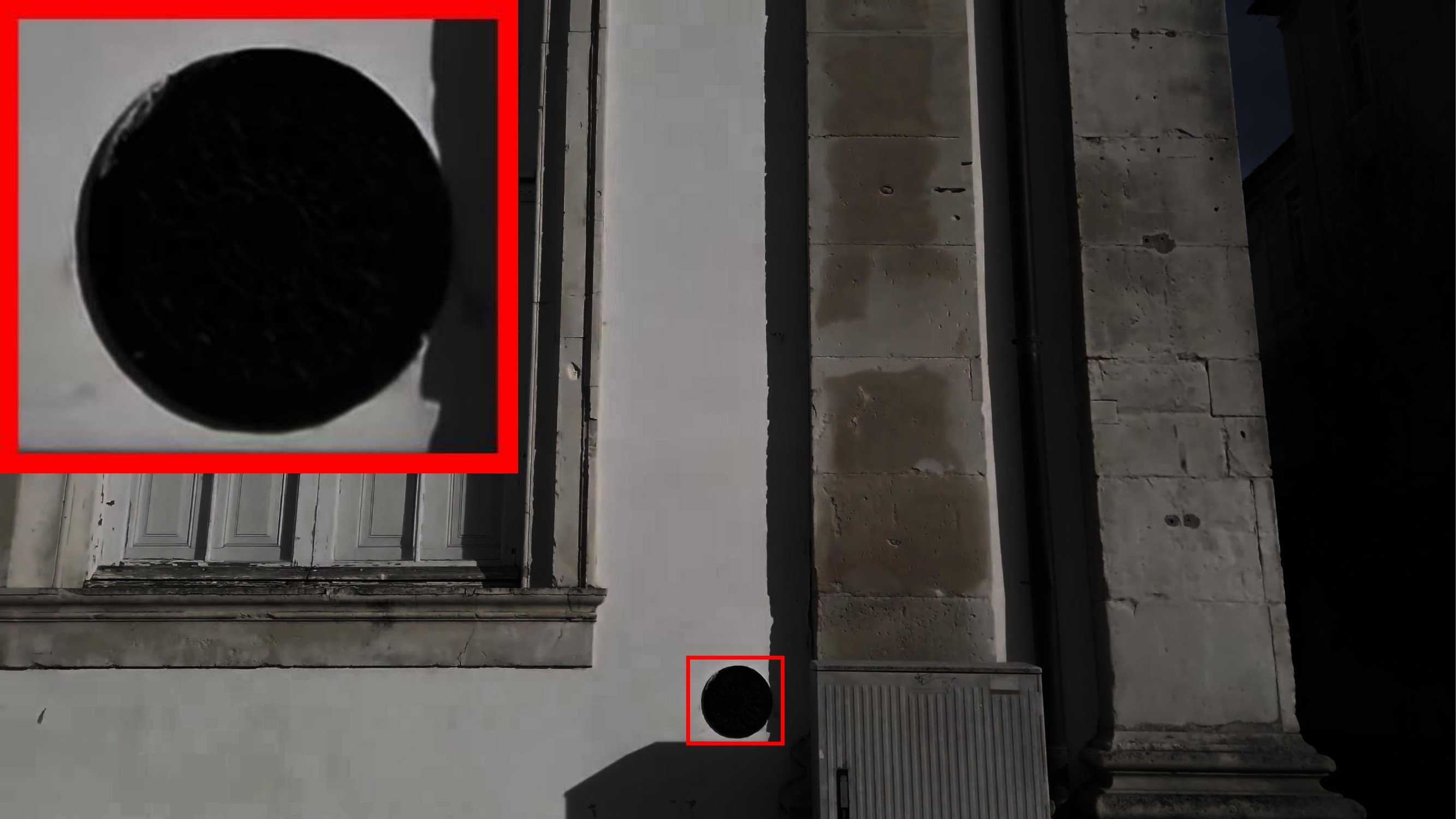}
        &
        \includegraphics[width=0.192\textwidth]
        {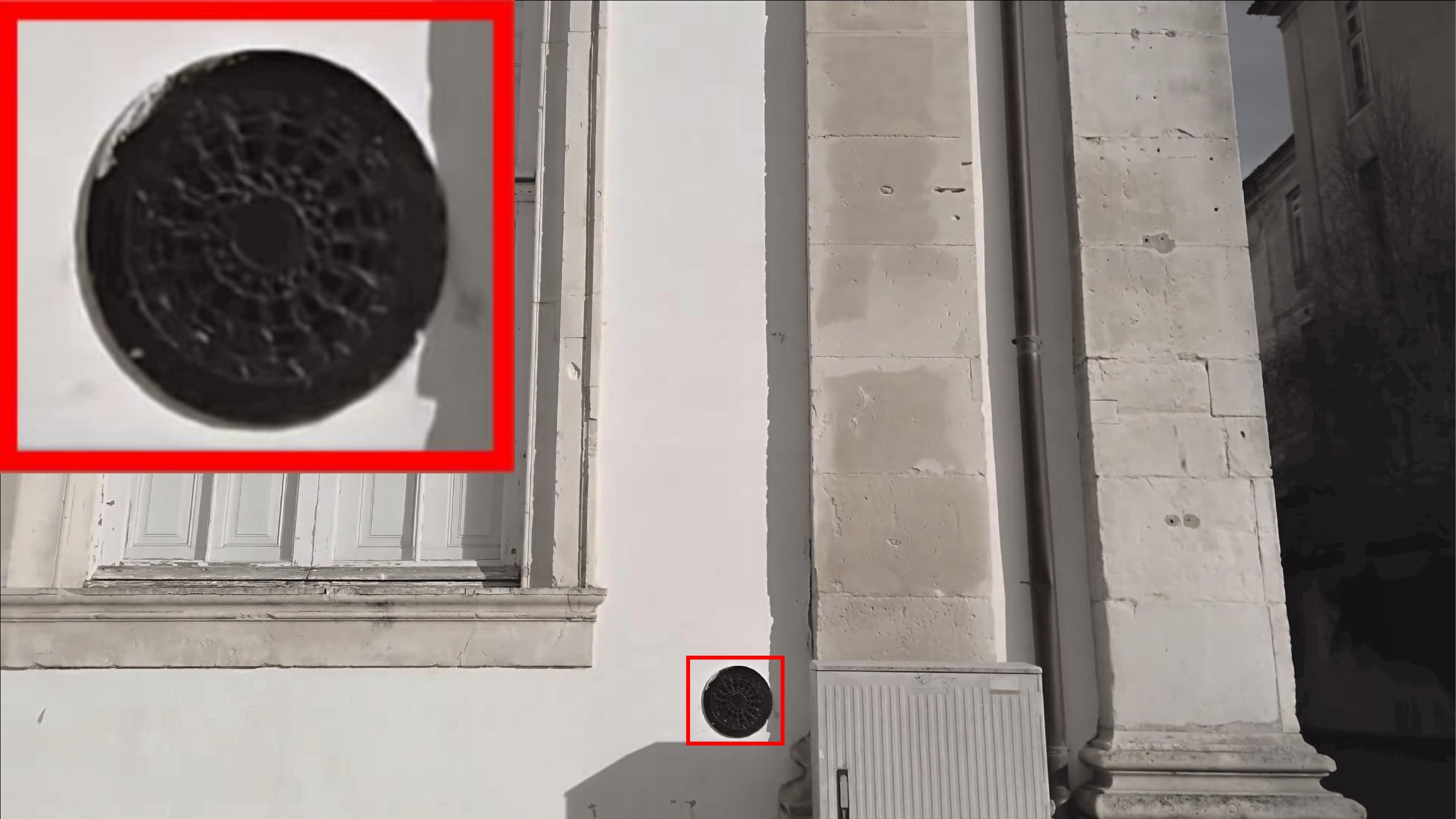}
        &
        \includegraphics[width=0.192\textwidth]
        {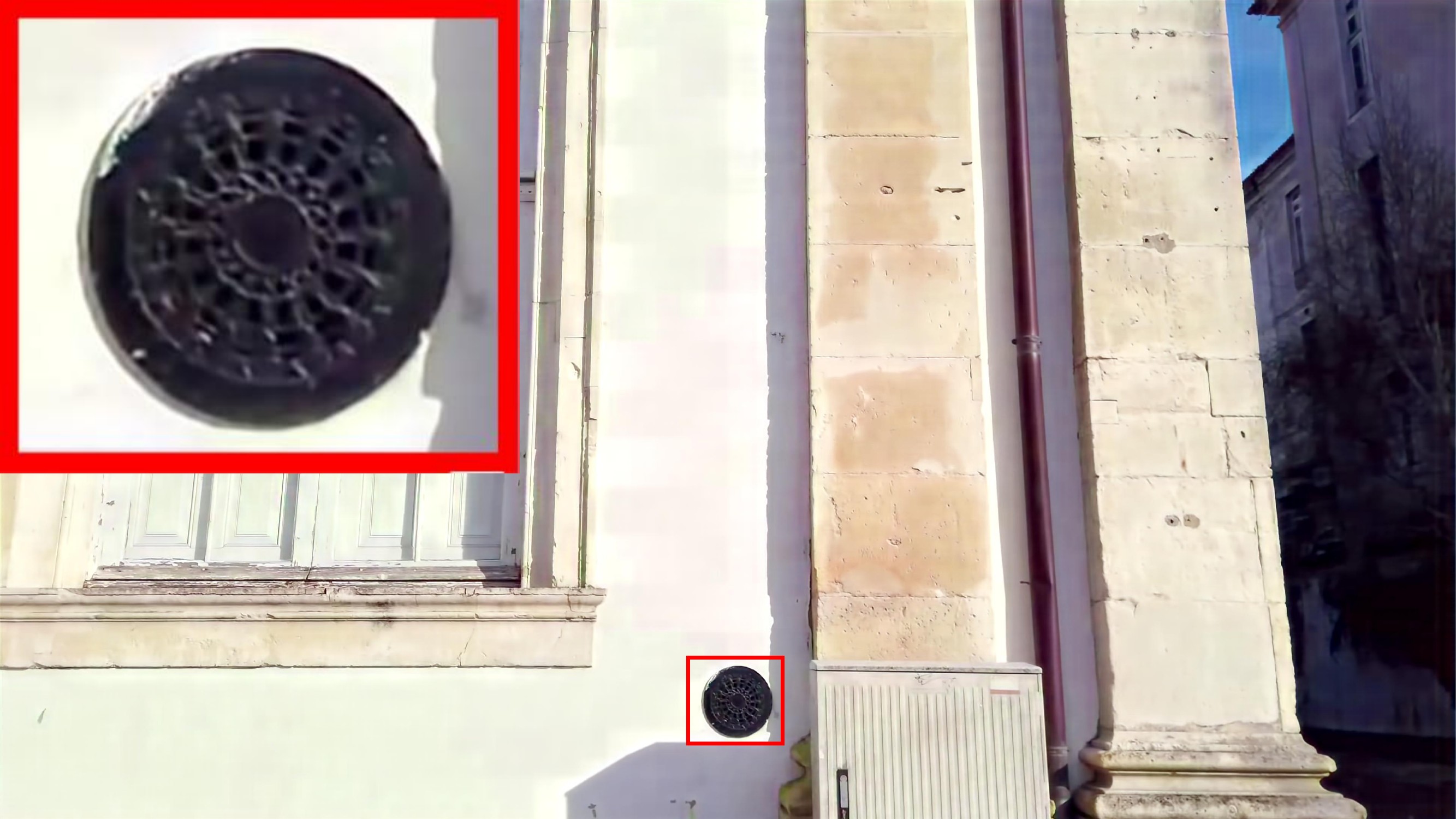}
        &
        \includegraphics[width=0.192\textwidth]
        {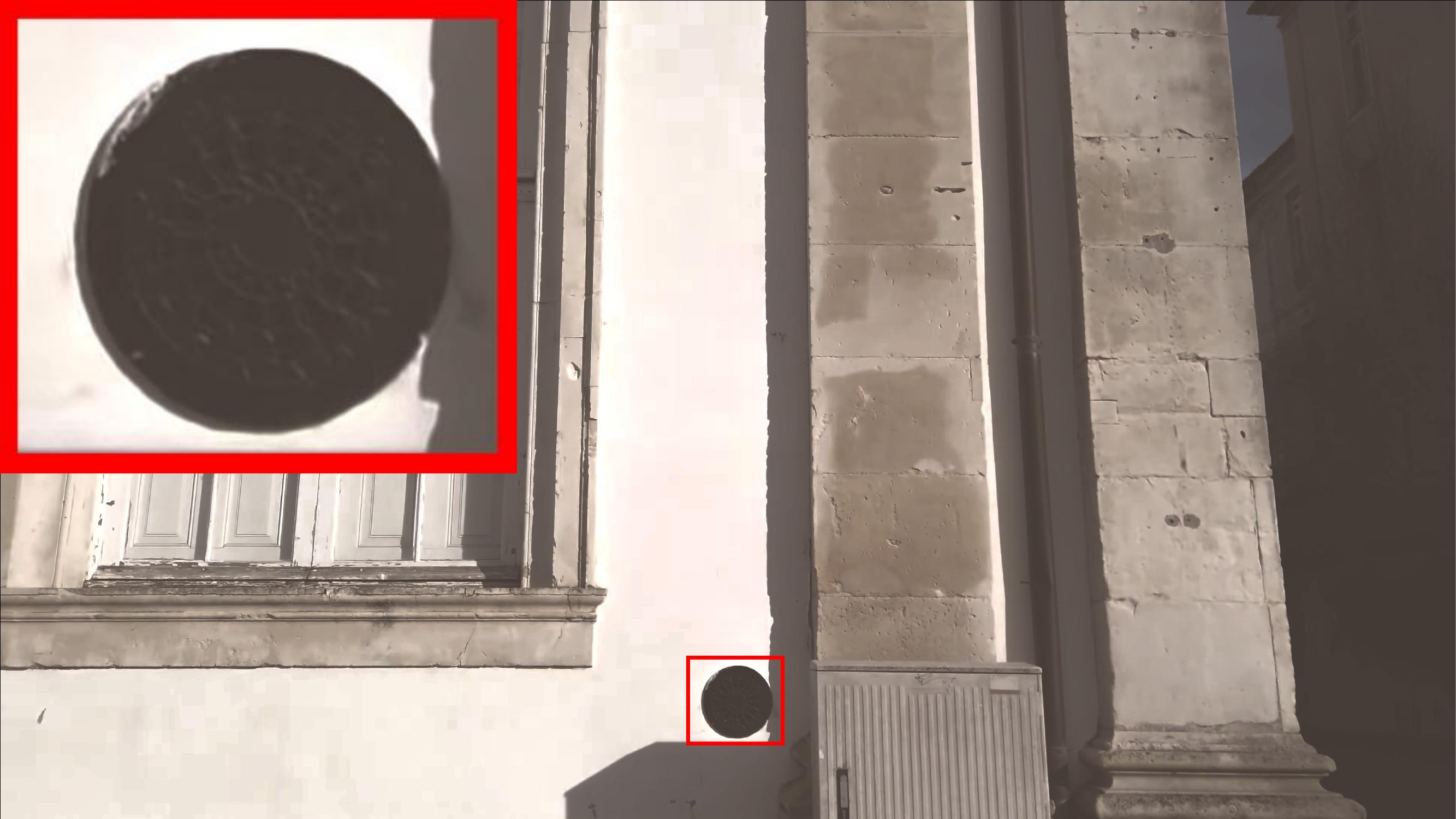}
        &
        \includegraphics[width=0.192\textwidth]
        {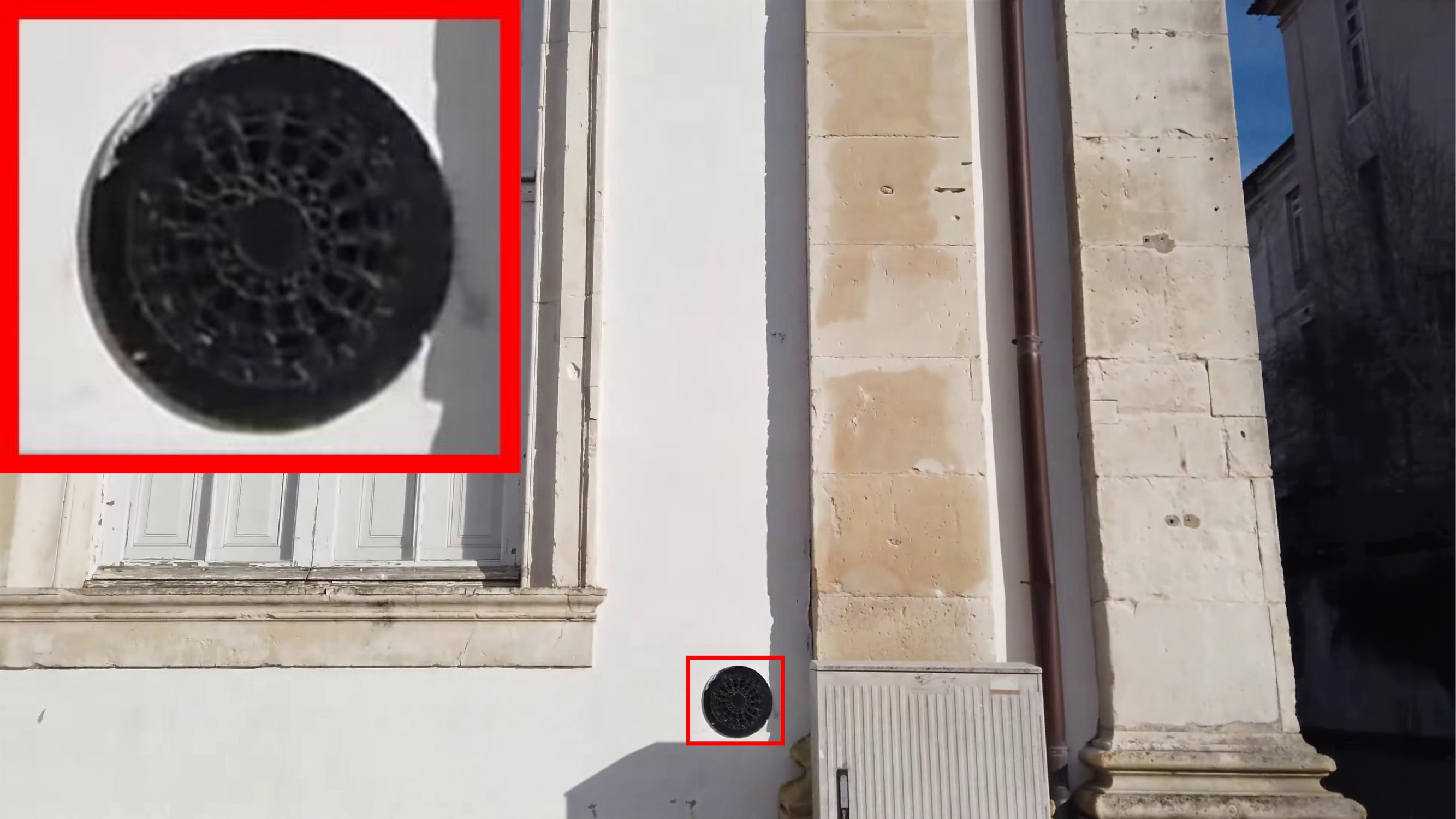}
        \\[-1mm]

        {\footnotesize (a) Input}
        &
        {\footnotesize (b) Uformer~\cite{wang2022uformer}}
        &
        {\footnotesize (c) Reti-Diff~\cite{he2025retidiff}}
        &
        {\footnotesize (d) UHDFour~\cite{li2023uhdfour}}
        &
        {\footnotesize (e) Wave-Mamba~\cite{zou2024wavemamba}}
        \\[1mm]

        \includegraphics[width=0.192\textwidth]
        {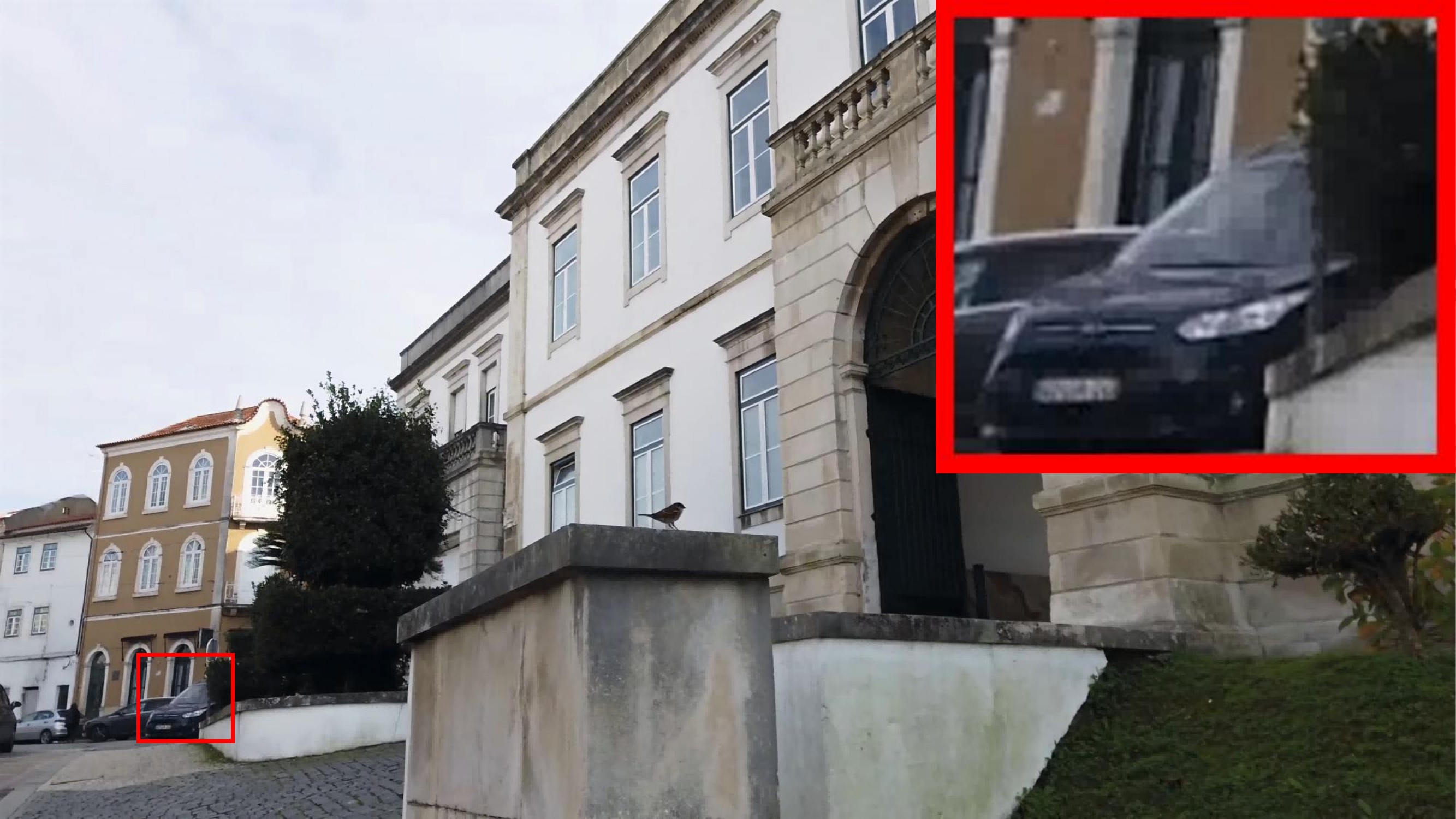}
        &
        \includegraphics[width=0.192\textwidth]
        {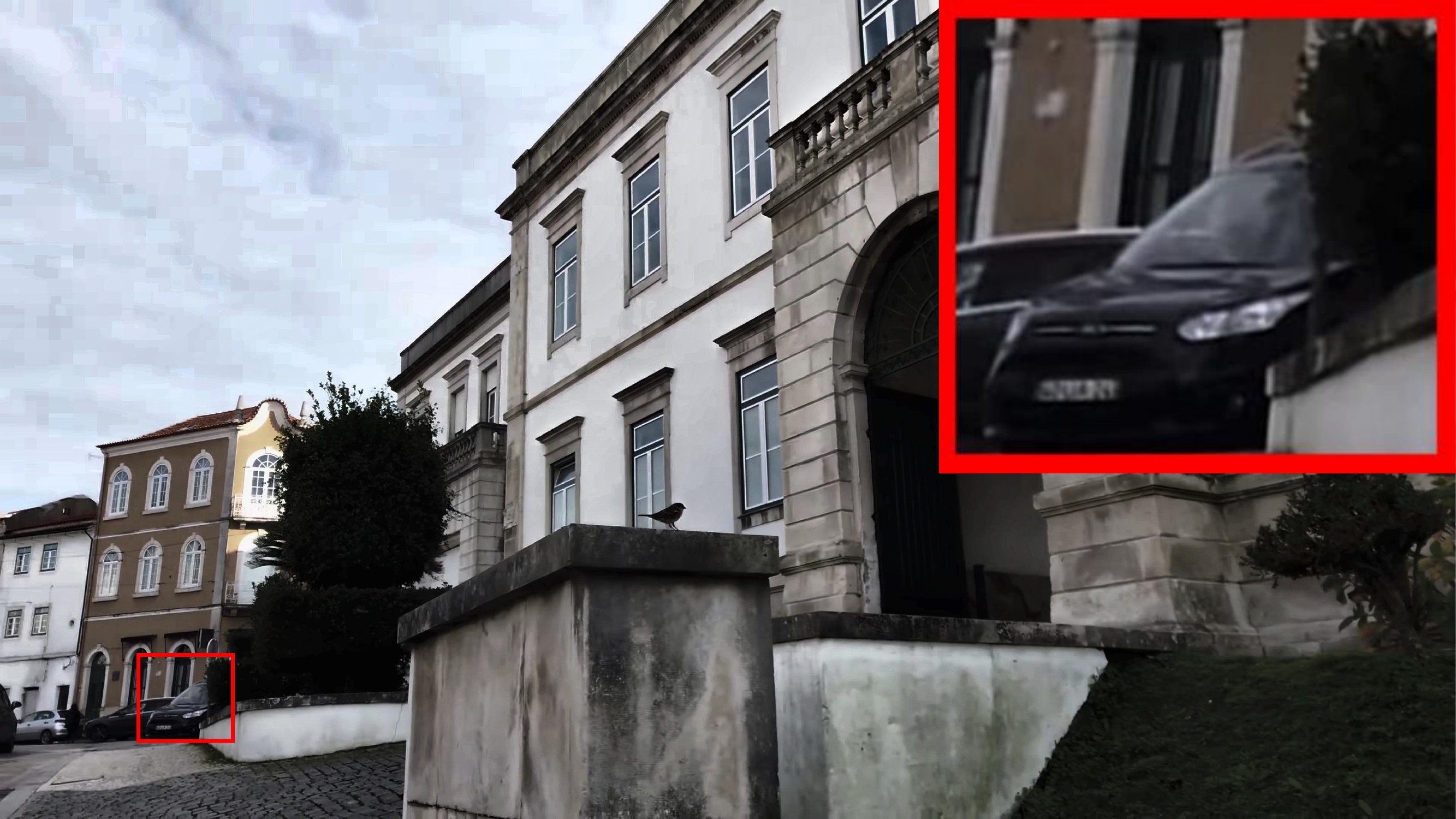}
        &
        \includegraphics[width=0.192\textwidth]
        {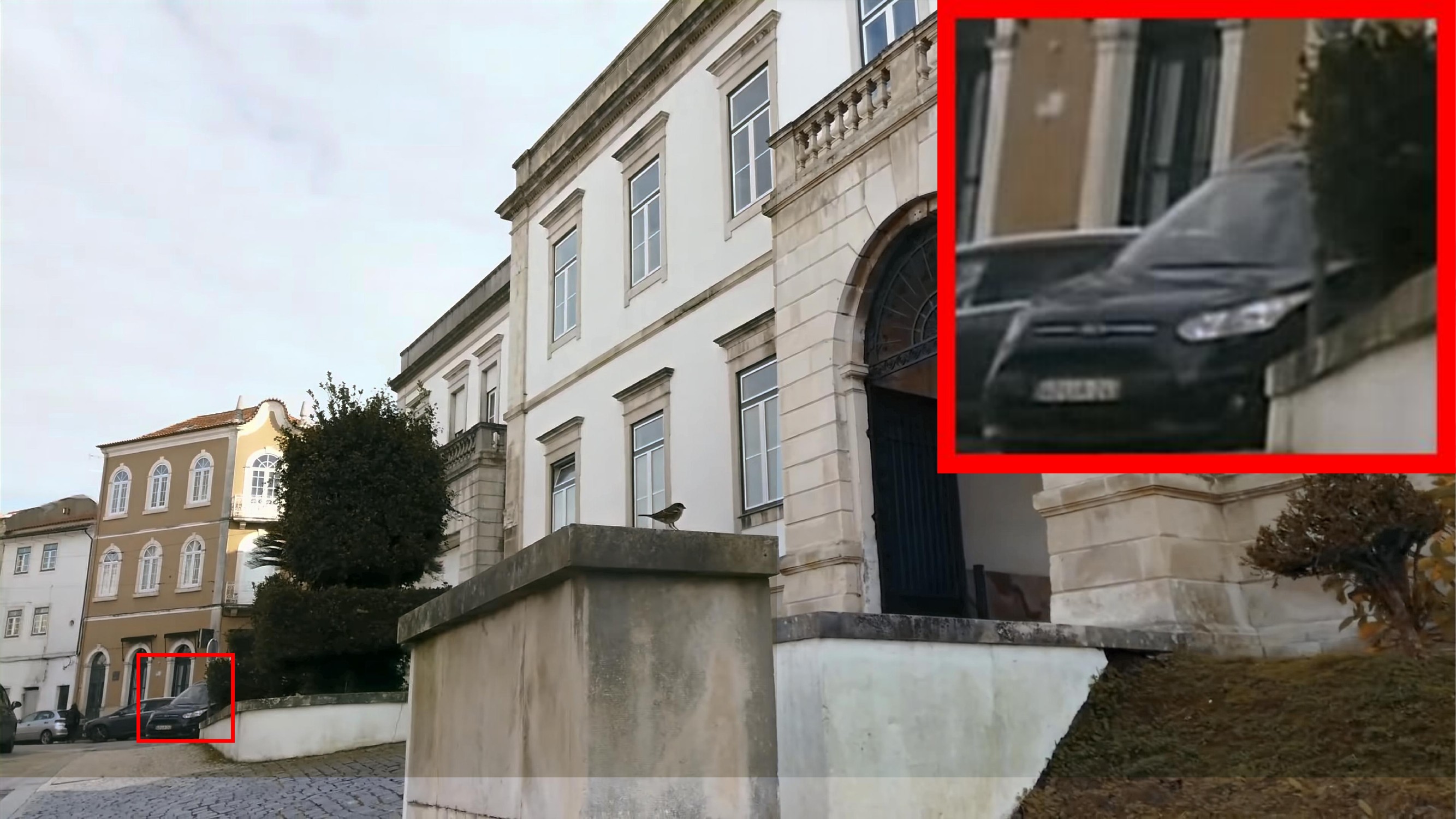}
        &
        \includegraphics[width=0.192\textwidth]
        {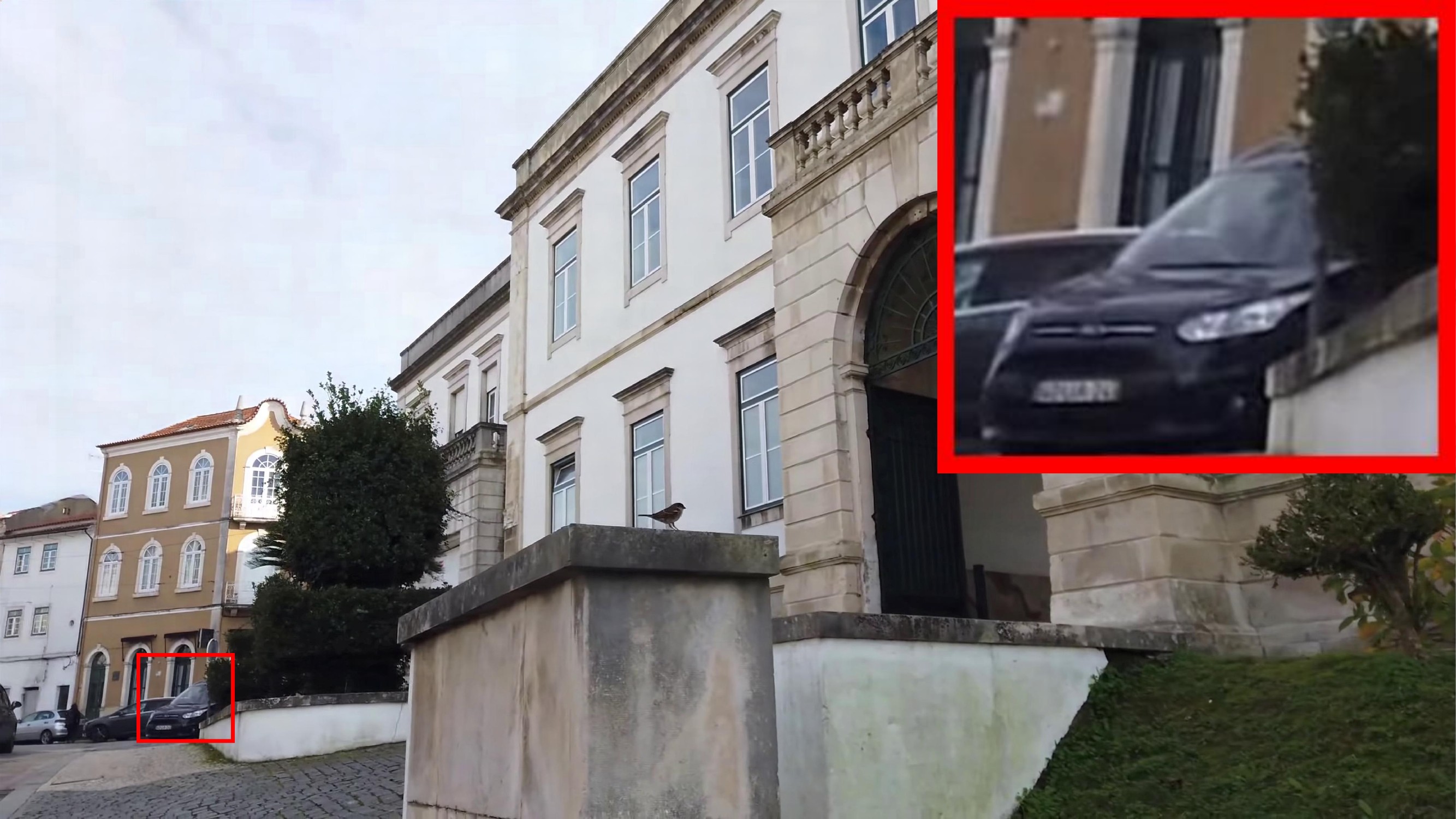}
        &
        \includegraphics[width=0.192\textwidth]
        {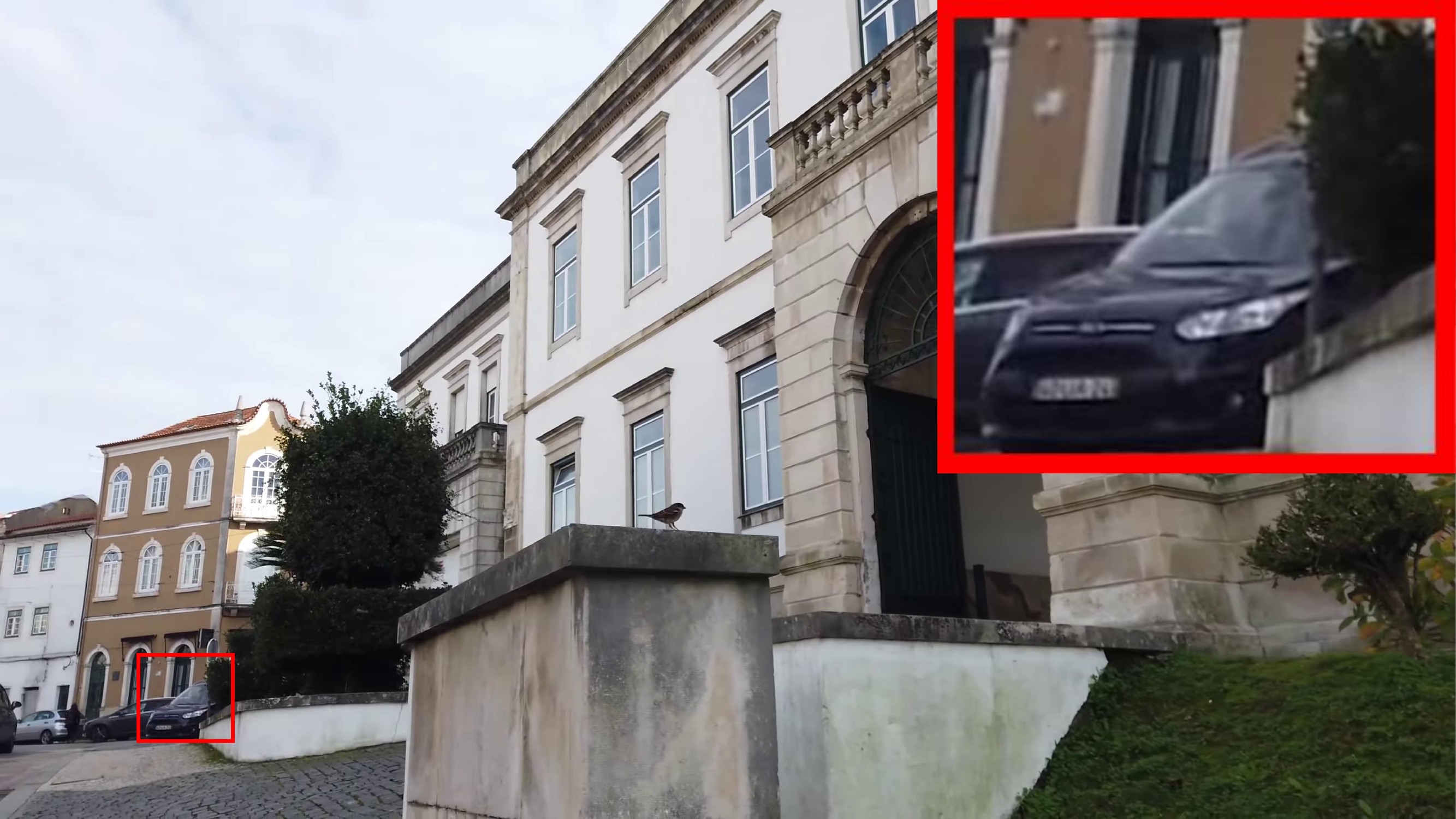}
        \\
        \includegraphics[width=0.192\textwidth]
        {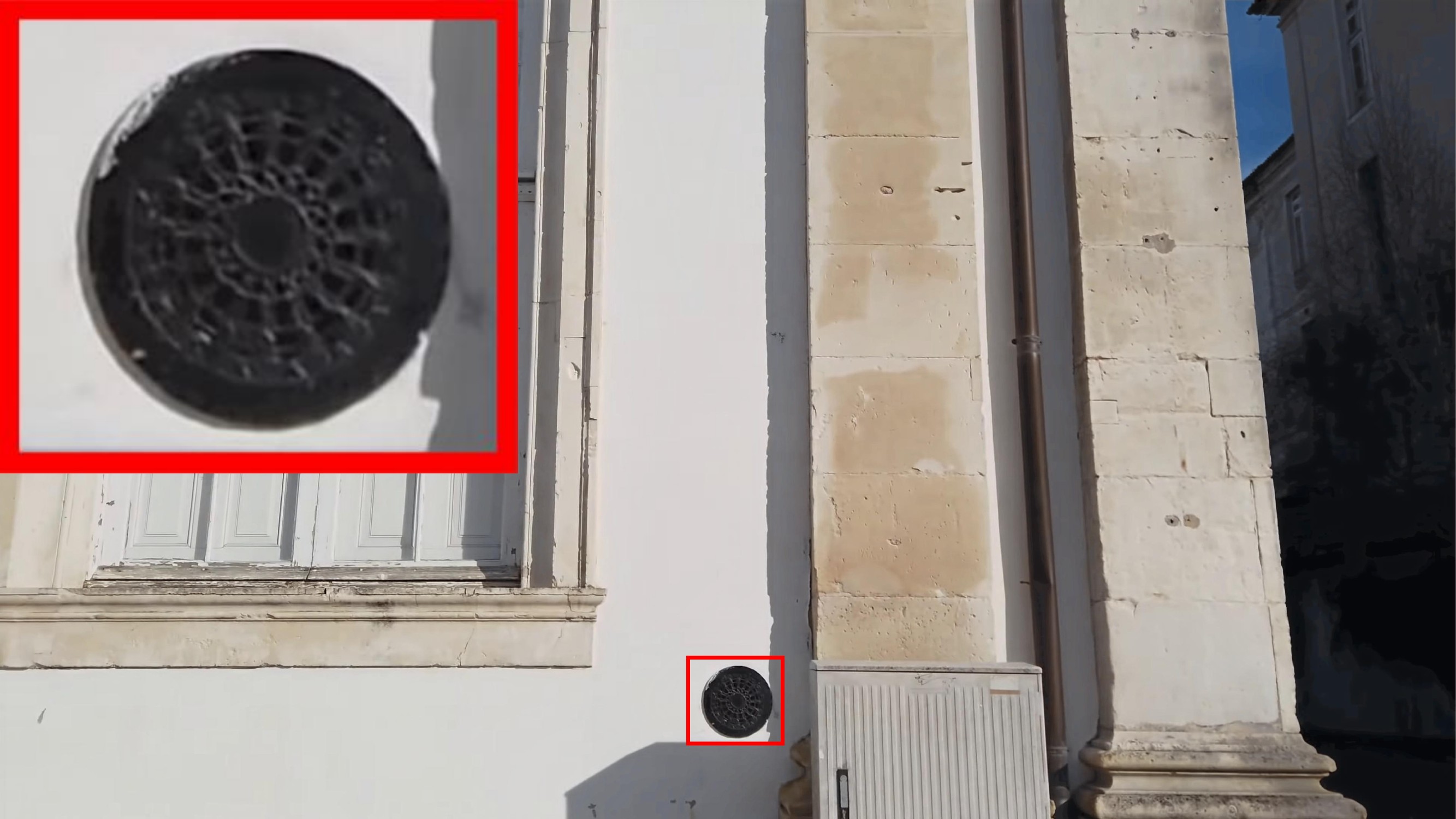}
        &
        \includegraphics[width=0.192\textwidth]
        {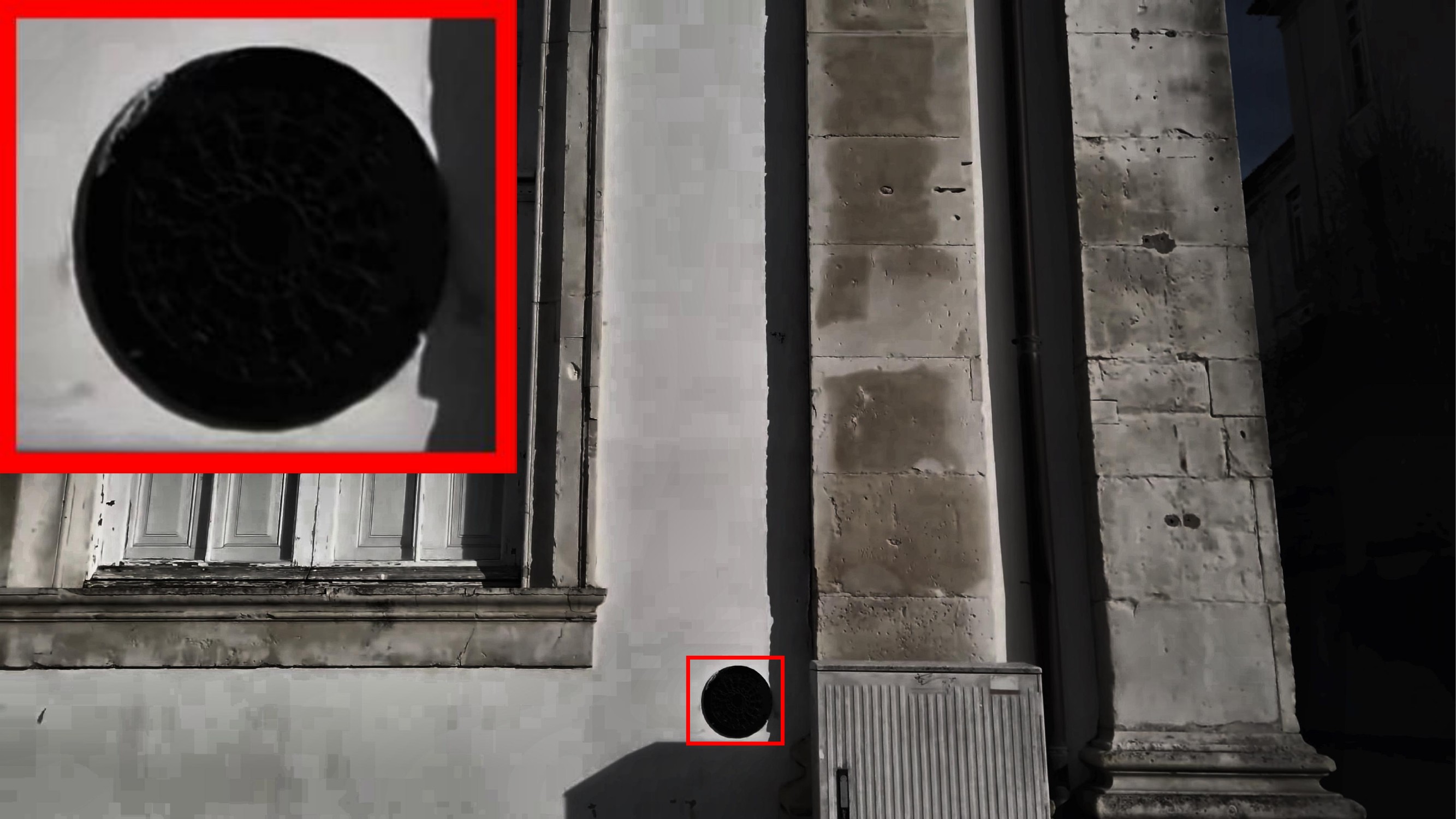}
        &
        \includegraphics[width=0.192\textwidth]
        {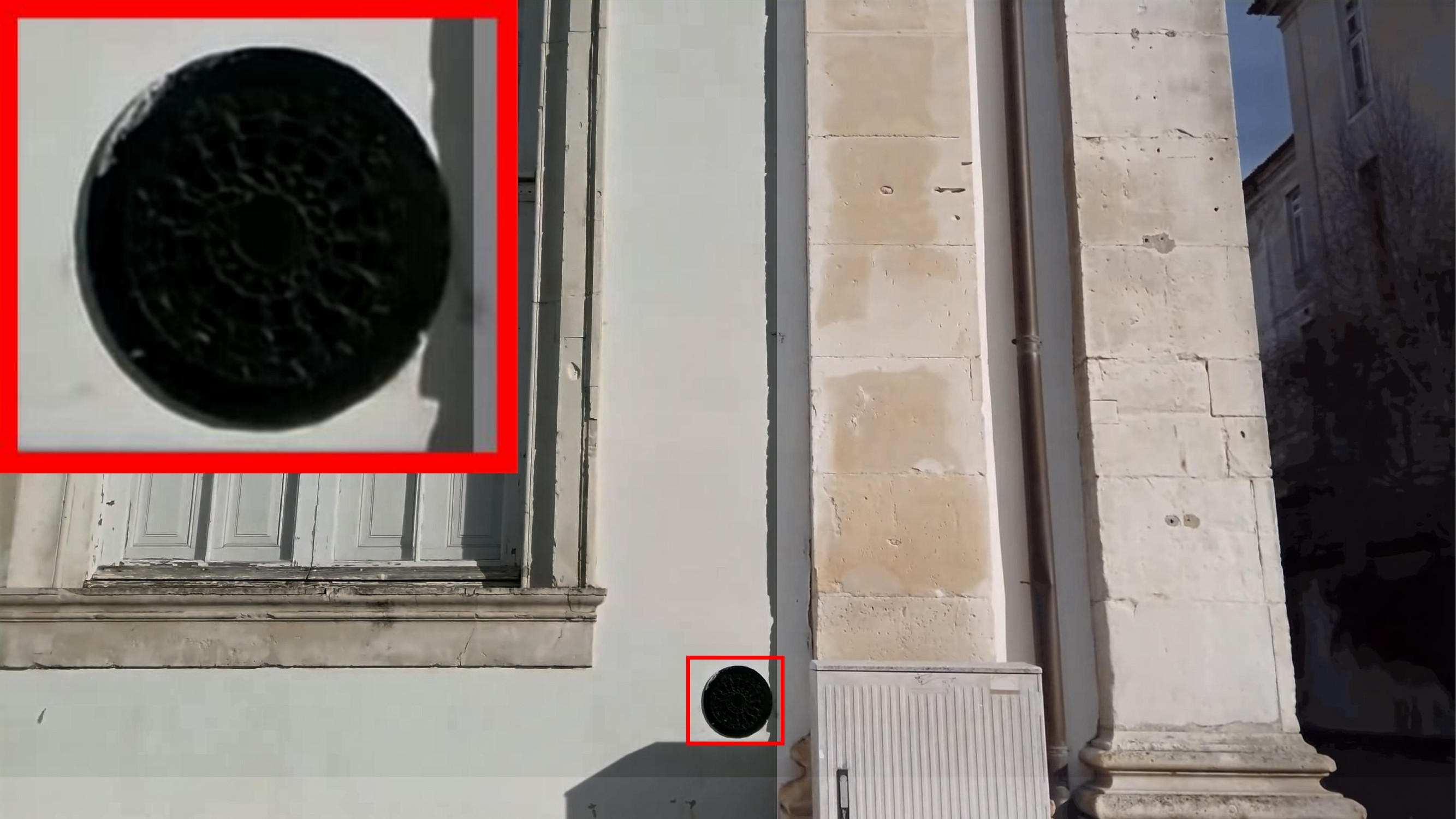}
        &
        \includegraphics[width=0.192\textwidth]
        {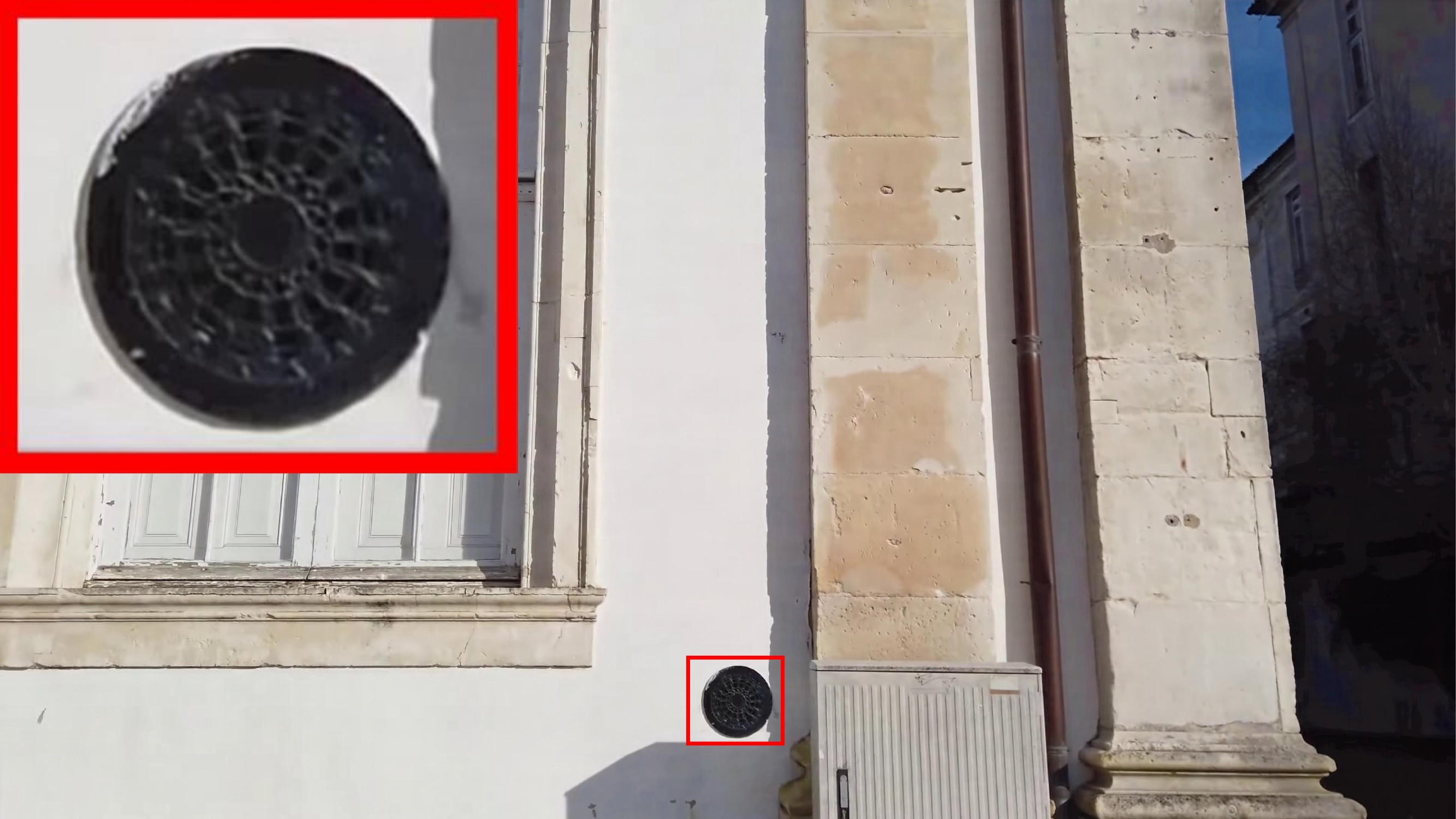}
        &
        \includegraphics[width=0.192\textwidth]
        {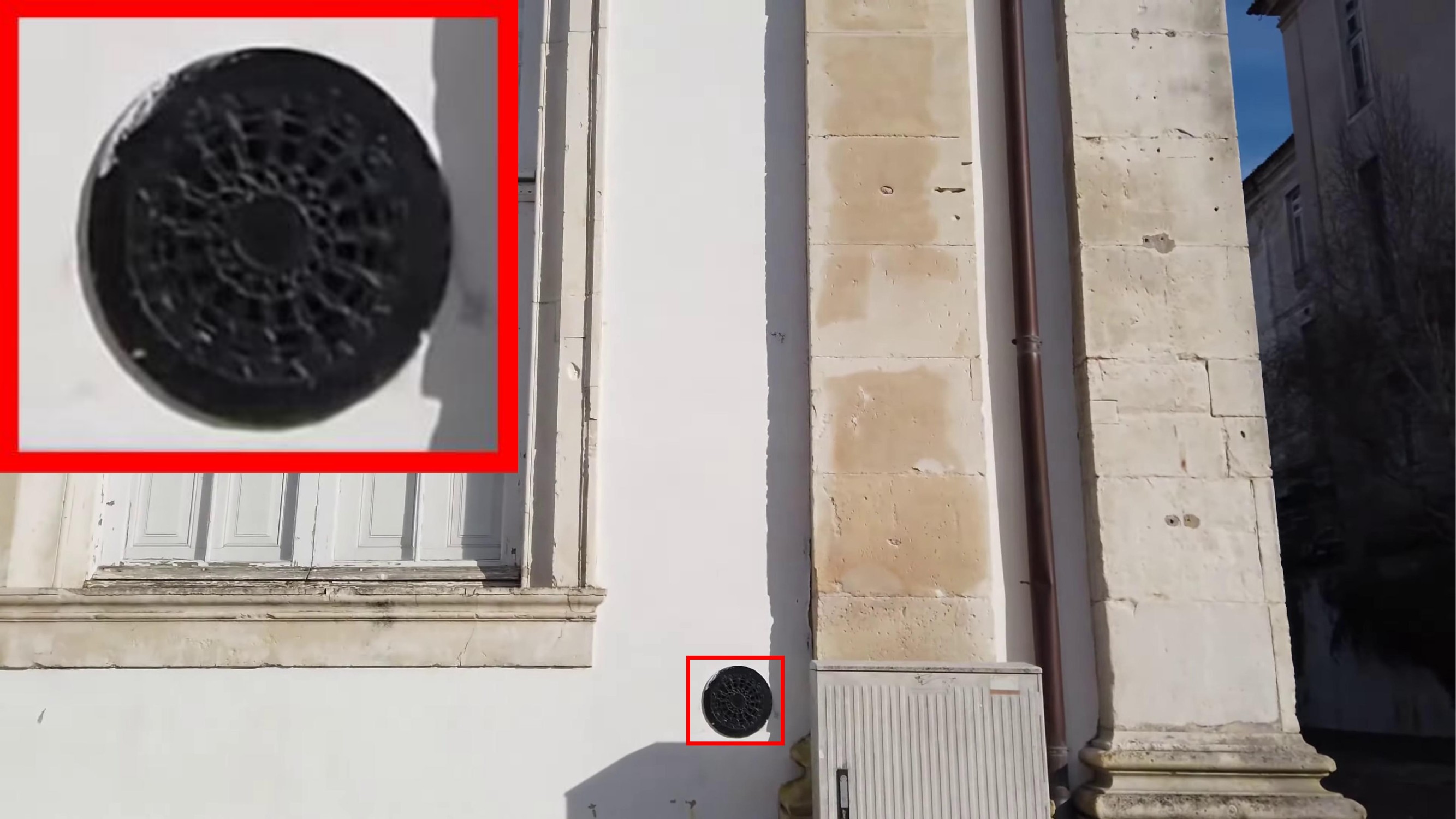}
        \\

        {\footnotesize (f) UHDFormer~\cite{wang2024uhdformer}}
        &
        {\footnotesize (g) UHDDIP~\cite{Wang2026uhddip}}
        &
        {\footnotesize (h) UHD-Processor~\cite{liu2025uhdprocessor}}
        &
        {\footnotesize \textbf{(i) BiFreqNet}}
        &
        {\footnotesize (j) GT}
        \\

    \end{tabular}
    \endgroup

    \resultcaption{
        Visual comparison results on the UHD-LOL4K dataset~\cite{LLFormer}. BiFreqNet produces more visually pleasing results with more faithful illumination and clearer details.
    }
    \label{fig:vis_uhdlol4k}
    \vspace{-2mm}
\end{figure*}
\section{Experiments}
\label{sec:experiments}

\subsection{Experimental Setup}

\noindent\textbf{Implementation Details.}
BiFreqNet adopts a four-scale feature modeling strategy with channel dimensions $\{ C_i \}_{i=1}^{4} = \{ 16, 32, 64, 64\}$.
The numbers of stacked DFI modules at the four scales are set to $\{1,1,1,2\}$, respectively. Within the S$^2$LFE blocks, the numbers of low-frequency Mamba blocks $\{N_i\}_{i=1}^4$ are set to $\{2,4,6,8\}$, and the local scanning length is fixed to 8 for all scales. The VAE follows the architecture of~\cite{Rombach2022LDM} and remains frozen during training so that the backbone provides a stable generative prior for subsequent reconstruction.

BiFreqNet is implemented in PyTorch and trained on a single NVIDIA RTX 3090 GPU. We train the network using AdamW with an initial learning rate of $8\times10^{-4}$ and a batch size of 8. The learning rate is gradually decayed to $2\times10^{-4}$ using the Cosine Annealing scheduler. During training, $512\times512$ patches are randomly cropped, and the network is trained for 600K iterations.

\noindent\textbf{Datasets.}
We evaluate BiFreqNet on two widely used UHD low-light image enhancement datasets, UHD-LL~\cite{li2023uhdfour} and UHD-LOL4K~\cite{LLFormer}, as well as two general low-light image enhancement benchmarks, \ie, LOLv1~\cite{wei2018deep} and LOLv2-Real~\cite{yang2021sparsegradient}.
UHD-LL~\cite{li2023uhdfour} is a real-world paired UHD dataset containing 2,150 pairs of 4K images in 8-bit sRGB format, with 2,000 pairs for training and 150 pairs for testing. UHD-LOL~\cite{LLFormer} contains UHD-LOL4K and UHD-LOL8K subsets; in this work, we use UHD-LOL4K, which consists of 5,999 training pairs and 2,100 testing pairs. We further report results on the conventional-resolution LOL benchmarks~\cite{wei2018deep}. LOLv1 contains 485 training pairs and 15 testing pairs,  while LOLv2-Real~\cite{yang2021sparsegradient} contains 689 training pairs and 100 testing pairs captured under more challenging real low-light conditions.

\noindent\textbf{Baselines.} 
We compare BiFreqNet with 18 state-of-the-art methods, including 8 general LLIE methods (\ie, Zero-DCE++~\cite{li2021zero}, RUAS~\cite{liu2021retinex}, RetinexFormer~\cite{cai2023retinexformer}, DiffLL~\cite{jiang2023diffll}, RetinexMamba~\cite{bai2024retinexmamba}, MambaLLIE~\cite{weng2024mamballie}, Reti-Diff~\cite{he2025retidiff}, and GPP-LLIE~\cite{zhou2025gppllie}), 3 general image restoration methods (\ie, Uformer~\cite{wang2022uformer}, Restormer~\cite{zamir2022restormer}, and MambaIR~\cite{Guo2024MambaIR}), 4 UHD LLIE methods (\ie, UHDFour~\cite{li2023uhdfour}, LLFormer~\cite{LLFormer}, Wave-Mamba~\cite{zou2024wavemamba}, and SCE~\cite{wu2026sce}), and 3 UHD image restoration methods (\ie, UHDFormer~\cite{wang2024uhdformer}, UHD-Processor~\cite{liu2025uhdprocessor}, and UHDDIP~\cite{Wang2026uhddip}). 
For a fair comparison, we use the official source code to train and test all methods using the same data partition for each dataset, except for GPP-LLIE~\cite{zhou2025gppllie}, whose results are taken from the original paper.

\noindent\textbf{Evaluation Metrics.}
We adopt PSNR and SSIM~\cite{wang2004image} as the main quantitative metrics. PSNR measures pixel-level reconstruction fidelity, whereas SSIM evaluates structural similarity in terms of luminance, contrast, and local structure. Higher values indicate better restoration quality.

\subsection{Results on UHD Low-Light Image Enhancement}

The quantitative results on the UHD-LL~\cite{li2023uhdfour} and UHD-LOL4K~\cite{LLFormer} datasets are reported in Table~\ref{tab:sota_uhd}. On the UHD-LL dataset, BiFreqNet achieves the highest PSNR of 27.39~dB and ties for the best SSIM of 0.929. Compared with Wave-Mamba~\cite{zou2024wavemamba}, our method increases the SSIM from 0.913 to 0.929, while outperforming UHDFour~\cite{li2023uhdfour} and UHDDIP~\cite{Wang2026uhddip} by 1.16~dB and 0.64~dB in PSNR, respectively. It also surpasses the strongest non-UHD baseline, Reti-Diff~\cite{he2025retidiff}, by 0.47~dB in PSNR and 0.024 in SSIM. On the UHD-LOL4K dataset, BiFreqNet achieves the highest SSIM of 0.992, indicating better structural preservation, although its PSNR is slightly lower than that of Wave-Mamba~\cite{zou2024wavemamba}. 
%
%
In terms of PSNR, BiFreqNet is only 0.14~dB and 0.04~dB lower than Wave-Mamba~\cite{zou2024wavemamba} and LLFormer~\cite{LLFormer}, respectively, while outperforming UHDFormer~\cite{wang2024uhdformer}, UHDFour~\cite{li2023uhdfour}, UHD-Processor~\cite{liu2025uhdprocessor}, and UHDDIP~\cite{Wang2026uhddip}.
Compared with DiffLL~\cite{jiang2023diffll}, the strongest non-UHD baseline, BiFreqNet improves PSNR from 36.95~dB to 37.29~dB and SSIM from 0.989 to 0.992. Overall, these results demonstrate BiFreqNet's strong UHD low-light restoration capability in both reconstruction fidelity and structural preservation.
The visual comparisons in Figures~\ref{fig:vis_uhdll}--\ref{fig:vis_uhdlol4k} further support the quantitative results. On UHD-LL, existing methods can recover the overall brightness but still suffer from blurred structures, local illumination inconsistency, or color deviations. For example, Uformer~\cite{wang2022uformer} and UHDFour~\cite{li2023uhdfour} weaken the boundaries and fine details, while Reti-Diff~\cite{he2025retidiff} and UHD-Processor~\cite{liu2025uhdprocessor} exhibit noticeable appearance deviations. In contrast, BiFreqNet produces more uniform illumination and preserves clearer text, boundaries, and local textures. On UHD-LOL4K, Uformer~\cite{wang2022uformer} and UHDFour~\cite{li2023uhdfour} tend to weaken local contrast, UHDDIP~\cite{Wang2026uhddip} leaves residual darkness in severely underexposed regions, and Reti-Diff~\cite{he2025retidiff} introduces over-enhancement and color deviation. BiFreqNet better restores the dark arched regions while retaining the fine structures.

\begin{table*}[t]
\centering


\renewcommand{\arraystretch}{1.0}
\setlength{\tabcolsep}{8.8pt}

\caption{
Quantitative comparison with state-of-the-art methods on the LOLv1~\cite{wei2018deep} and LOLv2-Real~\cite{yang2021sparsegradient} datasets. 
Higher PSNR and SSIM values indicate better performance, whereas lower Params and FLOPs values are preferred. FLOPs are computed on a $512\times512$ input.
}
\label{tab:sota_lol}

\footnotesize

\begin{tabular}{c|l|l|cccc|cc}
\Xhline{1pt}

\multirow{2}{*}{\textbf{Type}}
&
\multirow{2}{*}{\textbf{Method}}
&
\multirow{2}{*}{\textbf{Venue}}
&
\multicolumn{2}{c}{\textbf{LOLv1}~\cite{wei2018deep}}
&
\multicolumn{2}{c|}{\textbf{LOLv2-Real}~\cite{yang2021sparsegradient}}
&
\multirow{2}{*}{\textbf{Params (M) $\downarrow$}}
&
\multirow{2}{*}{\textbf{FLOPs (G) $\downarrow$}}
\\[-1pt]

\cline{4-5}
\cline{6-7}

& & &
\textbf{PSNR}~$\uparrow$
&
\textbf{SSIM}~$\uparrow$
&
\textbf{PSNR}~$\uparrow$
&
\textbf{SSIM}~$\uparrow$
&
&
\\

\Xhline{0.5pt}

\multirow{9}{*}{non-UHD}

& Uformer~\cite{wang2022uformer}
& CVPR'22
& 18.55 & 0.721 & 18.44 & 0.759
& 20.63 & 43.86 \\

& Restormer~\cite{zamir2022restormer}
& CVPR'22
& 22.37 & 0.816 & 24.91 & 0.851
& 26.13 & 141.75 \\

& Retinexformer~\cite{cai2023retinexformer}
& ICCV'23
& 25.16 & 0.845 & 22.80 & 0.840
& 1.61 & 15.57 \\

& MambaIR~\cite{Guo2024MambaIR}
& ECCV'24
& 22.31 & 0.826 & 21.25 & 0.831
& 4.30 & 60.66 \\

& RetinexMamba~\cite{bai2024retinexmamba}
& ICONIP'24
& 24.03 & 0.827 & 22.45 & 0.844
& 4.59 & 42.82 \\

& MambaLLIE~\cite{weng2024mamballie}
& NeurIPS'24
& 25.14 & 0.850 & 22.95 & 0.847
& 2.28 & 20.85 \\

& DiffLL~\cite{jiang2023diffll}
& SIGGRAPH'23
& 26.34 & 0.845 & 28.86 & 0.876
& 17.29 & 29.74 \\

& GPP-LLIE~\cite{zhou2025gppllie}
& AAAI'25
& 27.51 & - & \underline{29.23} & -
& - & - \\

& Reti-Diff~\cite{he2025retidiff}
& ICLR'25
& 25.35 & 0.866 & 22.97 & 0.858
& 26.11 & 87.63 \\

\Xhline{0.5pt}

\multirow{6}{*}{UHD}
& LLFormer~\cite{LLFormer}
& AAAI'23
& 23.65 & 0.816 & 27.75 & 0.861
& 24.52 & 22.52 \\

& UHDFour~\cite{li2023uhdfour}
& ICLR'23
& 23.09 & 0.871 & 21.78 & 0.870
& 17.54 & 75.63 \\

& Wave-Mamba~\cite{zou2024wavemamba}
& ACM MM'24
& 26.54 & 0.883 & 29.04 & \underline{0.901}
& 1.26 & 0.88 \\

& UHD-Processor~\cite{liu2025uhdprocessor}
& CVPR'25
& 19.80 & 0.795 & 17.06 & 0.723
& 1.60 & 4.17 \\

& UHDDIP~\cite{Wang2026uhddip}
& TCSVT'25
& \underline{27.62} & \underline{0.903}
& 27.78 & 0.851
& 0.81 & 34.73 \\
& \cellcolor{gray!30}\textbf{BiFreqNet (Ours)}
& \cellcolor{gray!30}--
& \cellcolor{gray!30}\textbf{28.26} 
& \cellcolor{gray!30}\textbf{0.910}
& \cellcolor{gray!30}\textbf{29.59} 
& \cellcolor{gray!30}\textbf{0.903}
& \cellcolor{gray!30}9.26 
& \cellcolor{gray!30}55.33 \\
\Xhline{1pt}
\end{tabular}
\vspace{-2mm}
\end{table*}
\begin{figure*}[!t]
    \centering
    \begingroup
    \setlength{\tabcolsep}{2.5pt}
    \renewcommand{\arraystretch}{1.0}
    \begin{tabular}{@{}ccccc@{}}
        \includegraphics[width=0.192\textwidth]
        {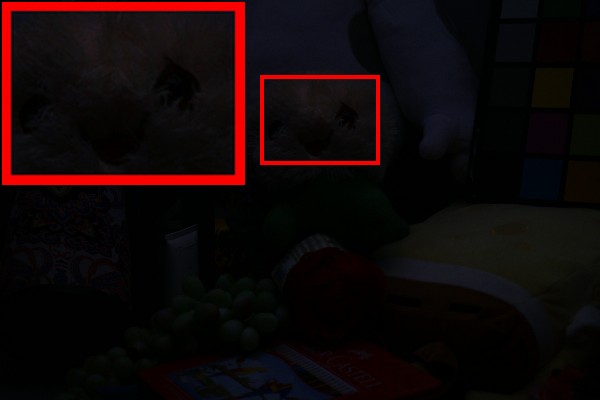}
        &
        \includegraphics[width=0.192\textwidth]
        {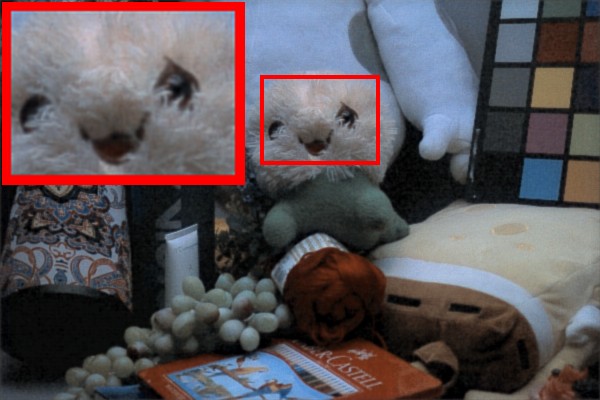}
        &
        \includegraphics[width=0.192\textwidth]
        {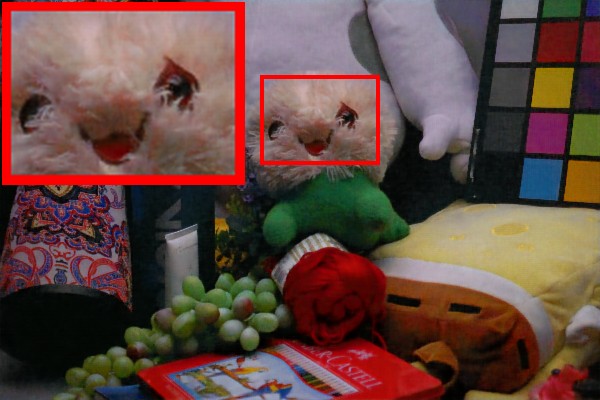}
        &
        \includegraphics[width=0.192\textwidth]
        {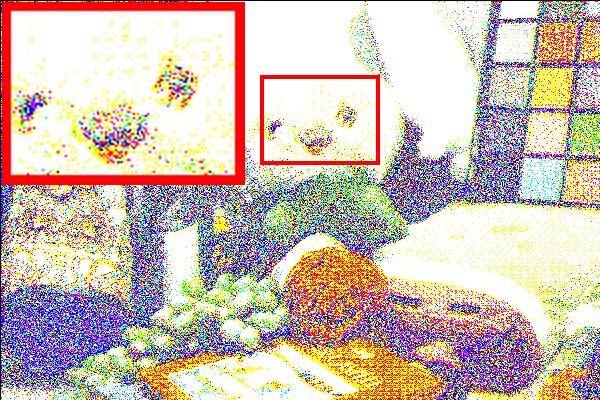}
        &
        \includegraphics[width=0.192\textwidth]
        {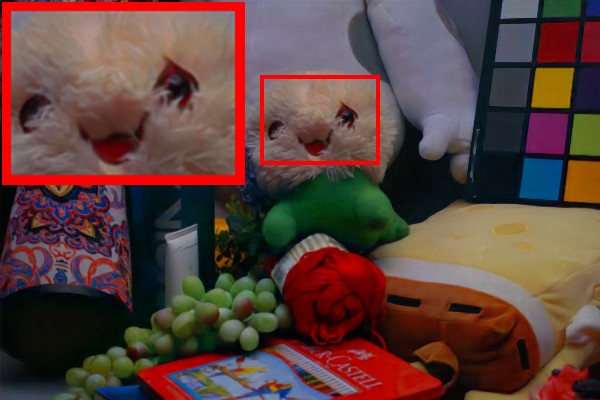}
        \\[-1mm]

        {\footnotesize (a) Input}
        &
        {\footnotesize (b) Uformer~\cite{wang2022uformer}}
        &
        {\footnotesize (c) RetinexFormer~\cite{cai2023retinexformer}}
        &
        {\footnotesize (d) Reti-Diff~\cite{he2025retidiff}}
        &
        {\footnotesize (e) Wave-Mamba~\cite{zou2024wavemamba}}
        \\[1mm]

        \includegraphics[width=0.192\textwidth]
        {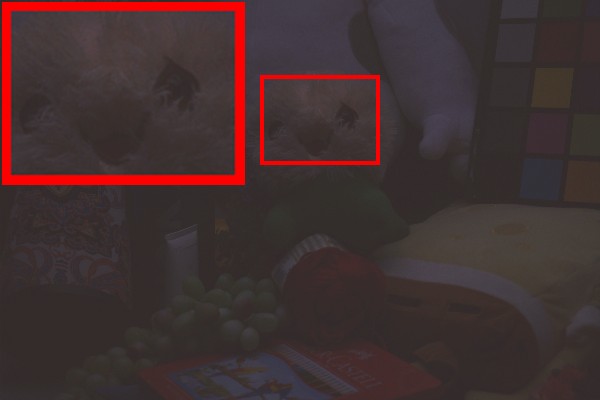}
        &
        \includegraphics[width=0.192\textwidth]
        {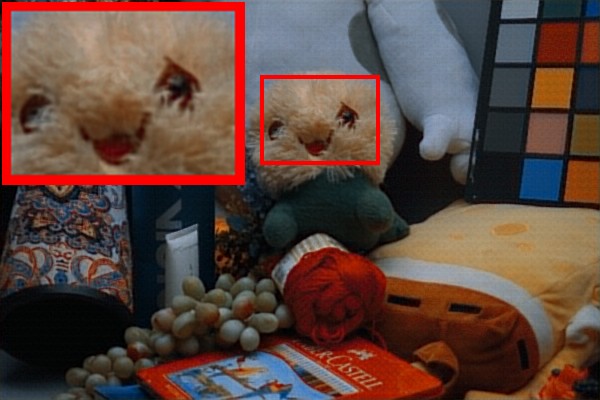}
        &
        \includegraphics[width=0.192\textwidth]
        {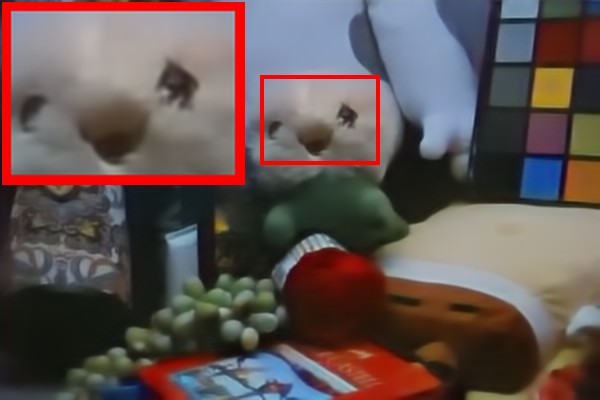}
        &
        \includegraphics[width=0.192\textwidth]
        {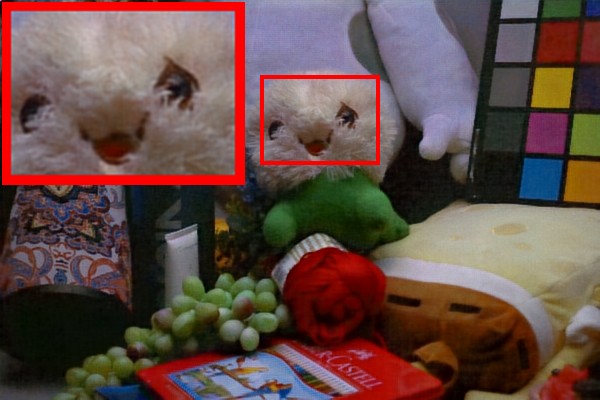}
        &
        \includegraphics[width=0.192\textwidth]
        {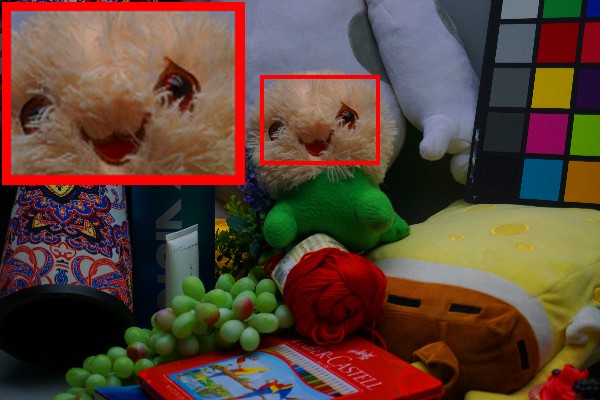}
        \\[-1mm]

        {\footnotesize (f) UHDFour~\cite{li2023uhdfour}}
        &
        {\footnotesize (g) UHDDIP~\cite{Wang2026uhddip}}
        &
        {\footnotesize (h) UHD-Processor~\cite{liu2025uhdprocessor}}
        &
        {\footnotesize \textbf{(i) BiFreqNet}}
        &
        {\footnotesize (j) GT}

    \end{tabular}
    \endgroup
    \caption{Visual comparison results on the LOLv1 dataset~\cite{wei2018deep}. BiFreqNet produces more visually pleasing results with more faithful illumination, clearer structures, and finer details.
    }
    \label{fig:vis_uhdlolv1}
    \vspace{-2mm}
\end{figure*}
\begin{figure*}[!t]
    \centering
    \begingroup
    \setlength{\tabcolsep}{2pt}
    \renewcommand{\arraystretch}{1.0}

    \begin{tabular}{@{}ccccc@{}}
        \includegraphics[width=0.194\textwidth,keepaspectratio]
        {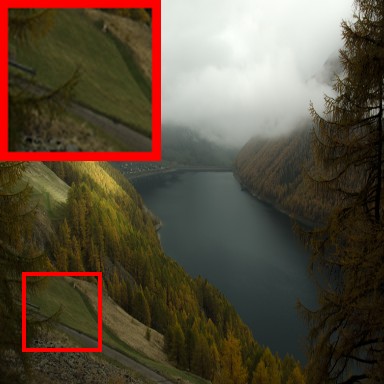}
        &
        \includegraphics[width=0.194\textwidth,keepaspectratio]
        {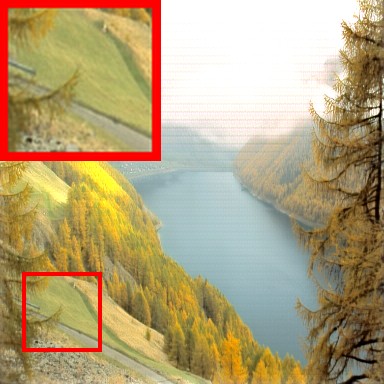}
        &
        \includegraphics[width=0.194\textwidth,keepaspectratio]
        {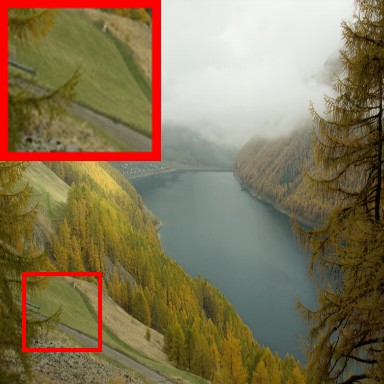}
        &
        \includegraphics[width=0.194\textwidth,keepaspectratio]
        {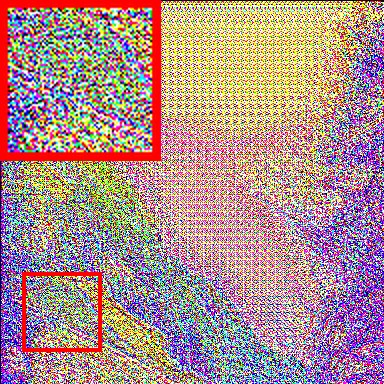}
        &
        \includegraphics[width=0.194\textwidth,keepaspectratio]
        {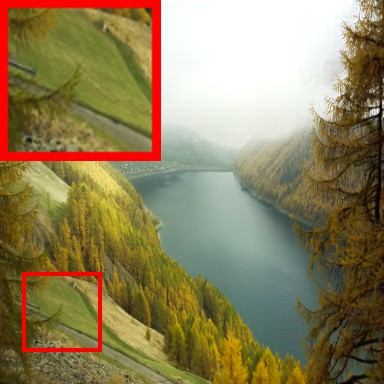}
        \\[-1mm]

        {\footnotesize (a) Input}
        &
        {\footnotesize (b) Uformer~\cite{wang2022uformer}}
        &
        {\footnotesize (c) RetinexFormer~\cite{cai2023retinexformer}}
        &
        {\footnotesize (d) Reti-Diff~\cite{he2025retidiff}}
        &
        {\footnotesize (e) Wave-Mamba~\cite{zou2024wavemamba}}
        \\[1mm]

        \includegraphics[width=0.194\textwidth,keepaspectratio]
        {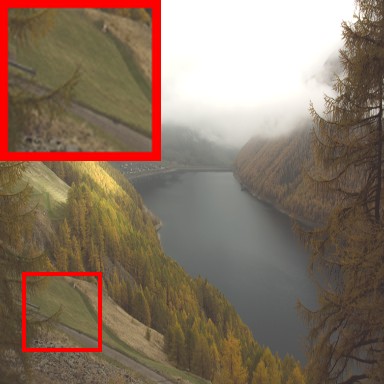}
        &
        \includegraphics[width=0.194\textwidth,keepaspectratio]
        {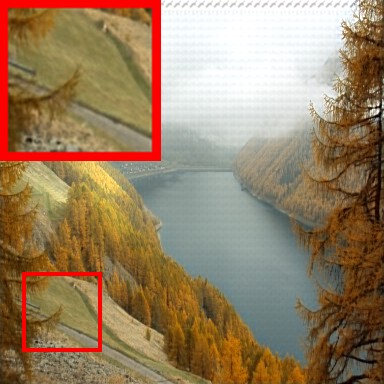}
        &
        \includegraphics[width=0.194\textwidth,keepaspectratio]
        {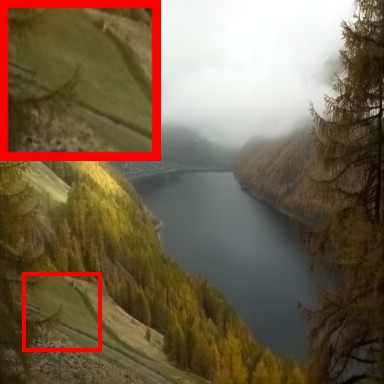}
        &
        \includegraphics[width=0.194\textwidth,keepaspectratio]
        {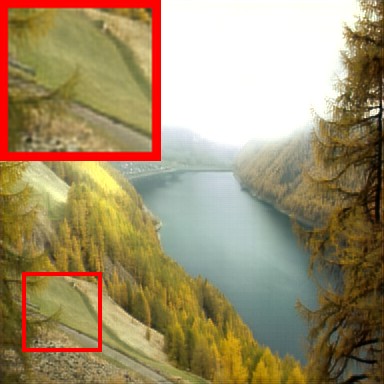}
        &
        \includegraphics[width=0.194\textwidth,keepaspectratio]
        {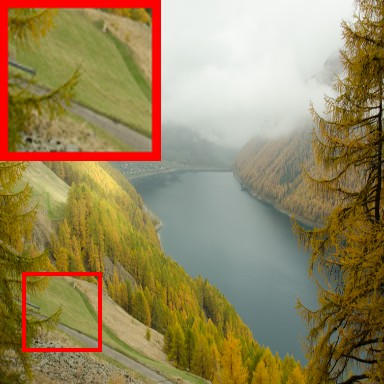}
        \\[-1mm]

        {\footnotesize (f) UHDFour~\cite{li2023uhdfour}}
        &
        {\footnotesize (g) UHDDIP~\cite{Wang2026uhddip}}
        &
        {\footnotesize (h) UHD-Processor~\cite{liu2025uhdprocessor}}
        &
        {\footnotesize \textbf{(i) BiFreqNet}}
        &
        {\footnotesize (j) GT}

    \end{tabular}
    \endgroup
    \caption{Visual comparison results on the LOLv2-Real dataset~\cite{yang2021sparsegradient}. BiFreqNet produces more visually pleasing results with more faithful illumination, clearer structures, and finer details.}
    \label{fig:vis_uhdlolv2real}
    \vspace{-2mm}
\end{figure*}

\subsection{Results on General Low-Light Image Enhancement}

We further evaluate BiFreqNet on the conventional-resolution LOLv1~\cite{wei2018deep} and LOLv2-Real~\cite{yang2021sparsegradient} benchmarks to assess its effectiveness beyond UHD inputs. As shown in Table~\ref{tab:sota_lol}, BiFreqNet achieves the best reported PSNR and SSIM on both datasets. On the LOLv1 dataset, it obtains 28.26~dB PSNR and 0.910 SSIM, outperforming the second-best UHDDIP~\cite{Wang2026uhddip} by 0.64~dB and 0.007, respectively. On the LOLv2-Real dataset, our method achieves 29.59~dB PSNR and 0.903 SSIM, exceeding GPP-LLIE~\cite{zhou2025gppllie} by 0.36~dB in PSNR and Wave-Mamba~\cite{zou2024wavemamba} by 0.002 in SSIM. These results demonstrate that BiFreqNet maintains strong restoration performance across both UHD and conventional-resolution low-light scenes.

The qualitative comparisons in Figures~\ref{fig:vis_uhdlolv1}--\ref{fig:vis_uhdlolv2real} show a consistent trend. On LOLv1, Uformer~\cite{wang2022uformer} tends to produce desaturated results with weakened local contrast, while RetinexFormer~\cite{cai2023retinexformer} and Wave-Mamba exhibit noticeable color and illumination deviations. UHDFour~\cite{li2023uhdfour} remains insufficiently enhanced in severely dark regions, Reti-Diff~\cite{he2025retidiff} introduces severe noise and color artifacts, and UHD-Processor~\cite{liu2025uhdprocessor} over-smooths local structures. In contrast, BiFreqNet recovers more natural illumination and faithful colors while preserving fine details, such as the cat fur, object boundaries, and textures in the highlighted regions. Similar observations can be made on LOLv2-Real. Although Wave-Mamba~\cite{zou2024wavemamba} and UHDDIP~\cite{Wang2026uhddip} produce competitive results, some local color or structural deviations remain. BiFreqNet achieves a more balanced illumination distribution and better preserves fine structures, including the grass textures and branch boundaries in the enlarged regions, yielding results visually closer to the ground truth.

\subsection{Ablation Study}
In this section, we conduct ablation experiments on the UHD low-light image enhancement task to investigate the individual contributions of the S$^2$LFE and CFE blocks within the proposed DFI module. All experiments are conducted on the UHD-LL dataset~\cite{li2023uhdfour}.

\subsubsection{Effect of DFI Module}
We first investigate the effectiveness of the two components in DFI, \ie, S$^2$LFE and CFE, by removing them individually or simultaneously.
As shown in Table~\ref{tab:DFI}, removing both the S$^2$LFE and CFE blocks (a) significantly degrades performance to 22.81~dB PSNR and 0.784 SSIM.  Retaining only CFE results (b), \ie, directly leveraging the original low-frequency feature $\mathbf{F}^{LL}_i$ instead of $\hat{\mathbf{F}}^{LL}_i$, in a 2.23~dB increase in PSNR, while retaining only S$^2$LFE (c) improves PSNR by 4.22~dB, indicating that S$^2$LFE provides the major restoration capability through spatial--spectral modeling of low-frequency representations.  Further introducing CFE improves PSNR/SSIM to 27.39~dB/0.929 with only 0.69M additional parameters, demonstrating its complementary role in refining high-frequency details under the guidance of enhanced low-frequency features. 
Fig.~\ref{fig:ablation of DFI} presents a visual comparison, where the model with the full DFI module achieves better illumination and detail restoration.
Overall, the two components are complementary: S$^2$LFE first restores reliable low-frequency illumination, upon which CFE refines high-frequency details, jointly yielding the best restoration performance.

\begin{table}[!t]
\centering
\begingroup
\footnotesize
\caption{Effect of the DFI module on UHD-LL~\cite{li2023uhdfour}. The best performance is marked in \textbf{bold}.}
\centering
\label{tab:DFI}

\renewcommand{\arraystretch}{1.01}
\setlength{\tabcolsep}{5.4pt}
\resizebox{1.008\columnwidth}{!}{%
    \begin{tabular}{c| c | c c |c c }
    \Xhline{1pt}
\textbf{ID} & \textbf{Options} & \textbf{PSNR~$\uparrow$} & \textbf{SSIM~$\uparrow$}  &\textbf{Params~$\downarrow$} 
& \textbf{FLOPs~$\downarrow$} \\
    \Xhline{0.5pt}
        (a) & w/o DFI 
        & 22.81 & 0.784 
        & 1.18M 
        & 21.53G 
        \\

        (b) & w/o S$^2$LFE 
        & 25.04 & 0.871 
        & 1.87M
        & 24.30G 
         \\

        (c) & w/o CFE 
        & 27.03 & 0.923 
        & 8.57M
        & 52.63G 
         \\

        \rowcolor{gray!30}
        \textbf{(d)} & \textbf{w/ DFI (Ours)} 
        & \textbf{27.39} & \textbf{0.929} 
        & 9.26M
        & 55.33G
         \\
   \Xhline{1pt}
    \end{tabular}
    }
\endgroup
\vspace{-2mm}
\end{table}
\begin{figure}[!t]
\centering
\setlength{\tabcolsep}{1pt}
\begin{tabular}{@{}ccc@{}}
    \includegraphics[width=0.326\linewidth]{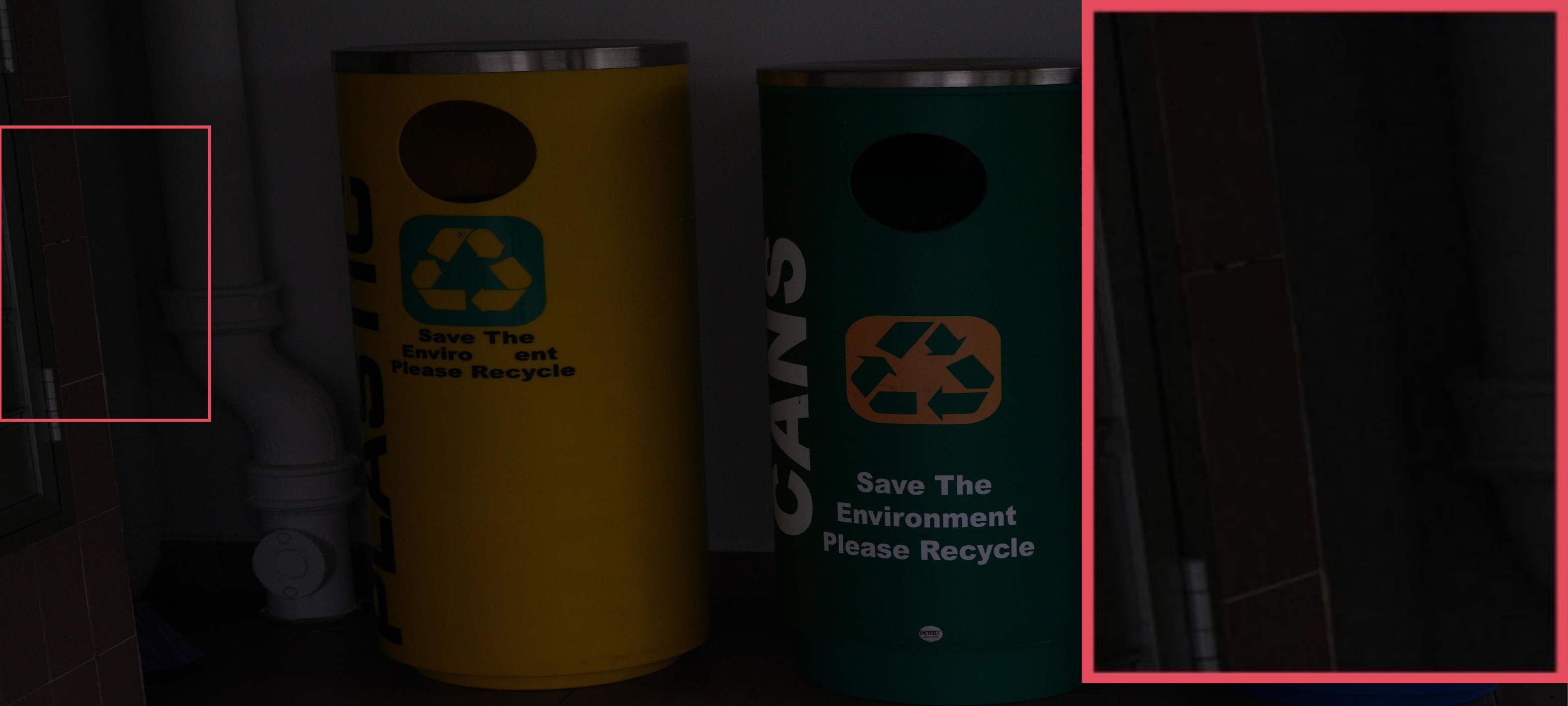} &
    \includegraphics[width=0.326\linewidth]{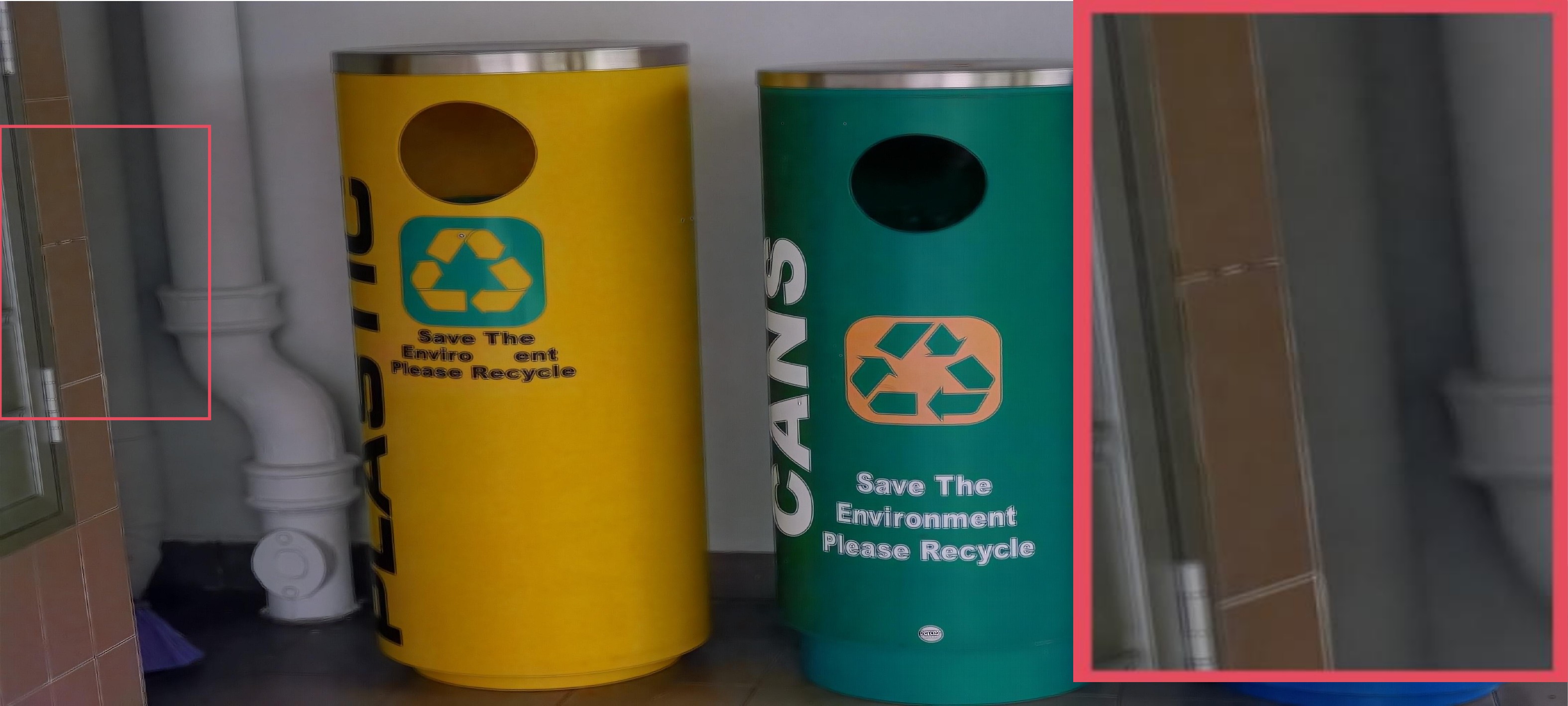} &
    \includegraphics[width=0.326\linewidth]{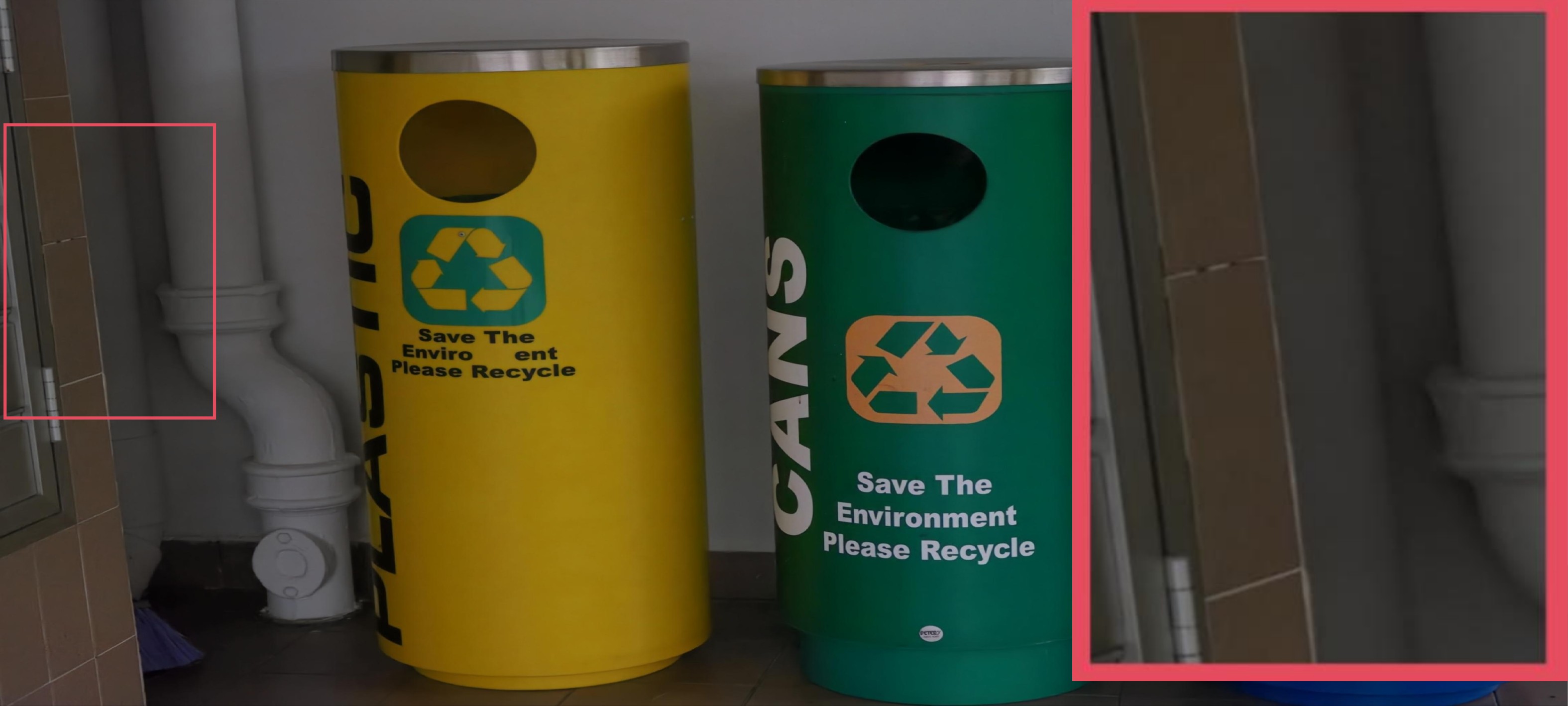} \\[-2.5pt]

    \scalebox{0.85}{(a) Input} &
    \scalebox{0.85}{(b) w/o DFI} &
    \scalebox{0.85}{(c) w/o S$^2$LFE} \\[-1pt]

    \includegraphics[width=0.326\linewidth]{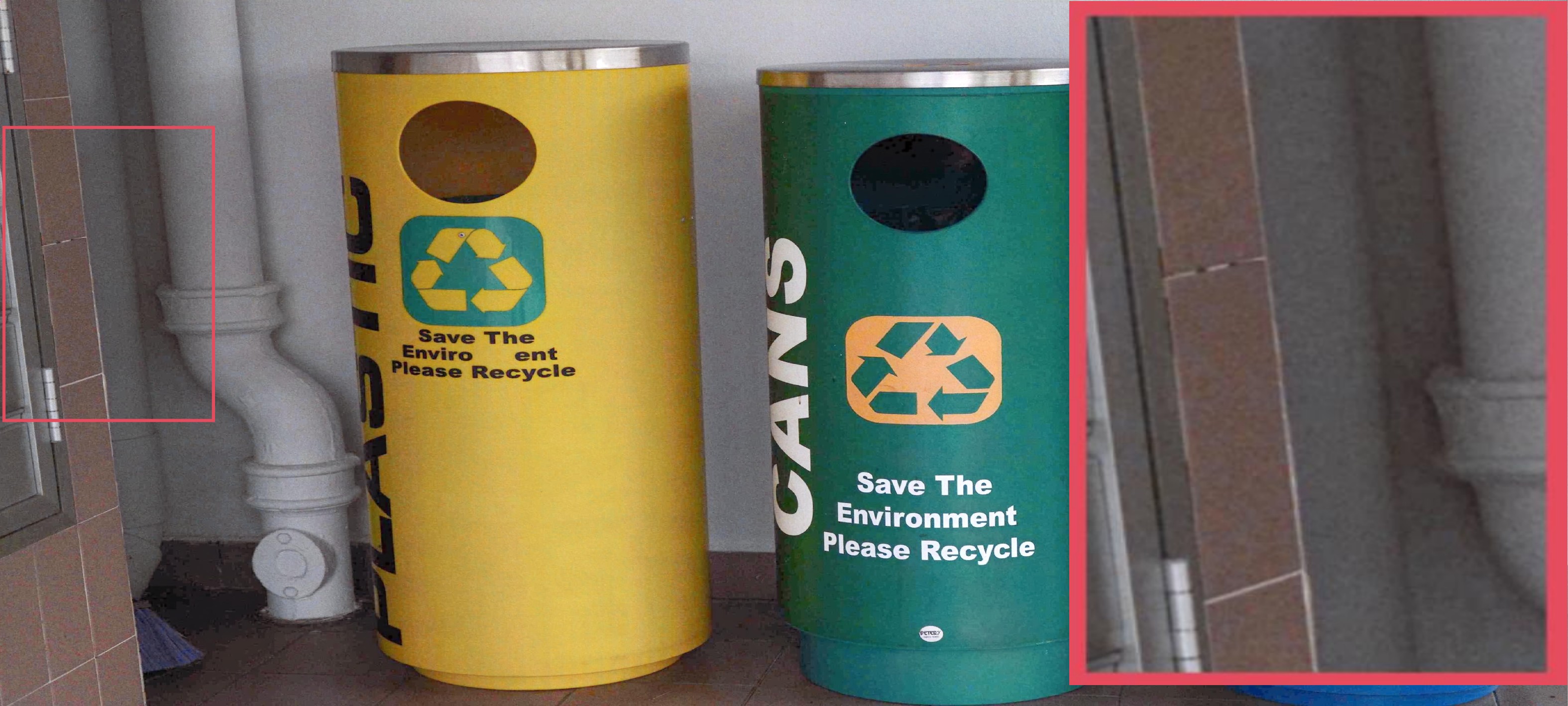} &
    \includegraphics[width=0.326\linewidth]{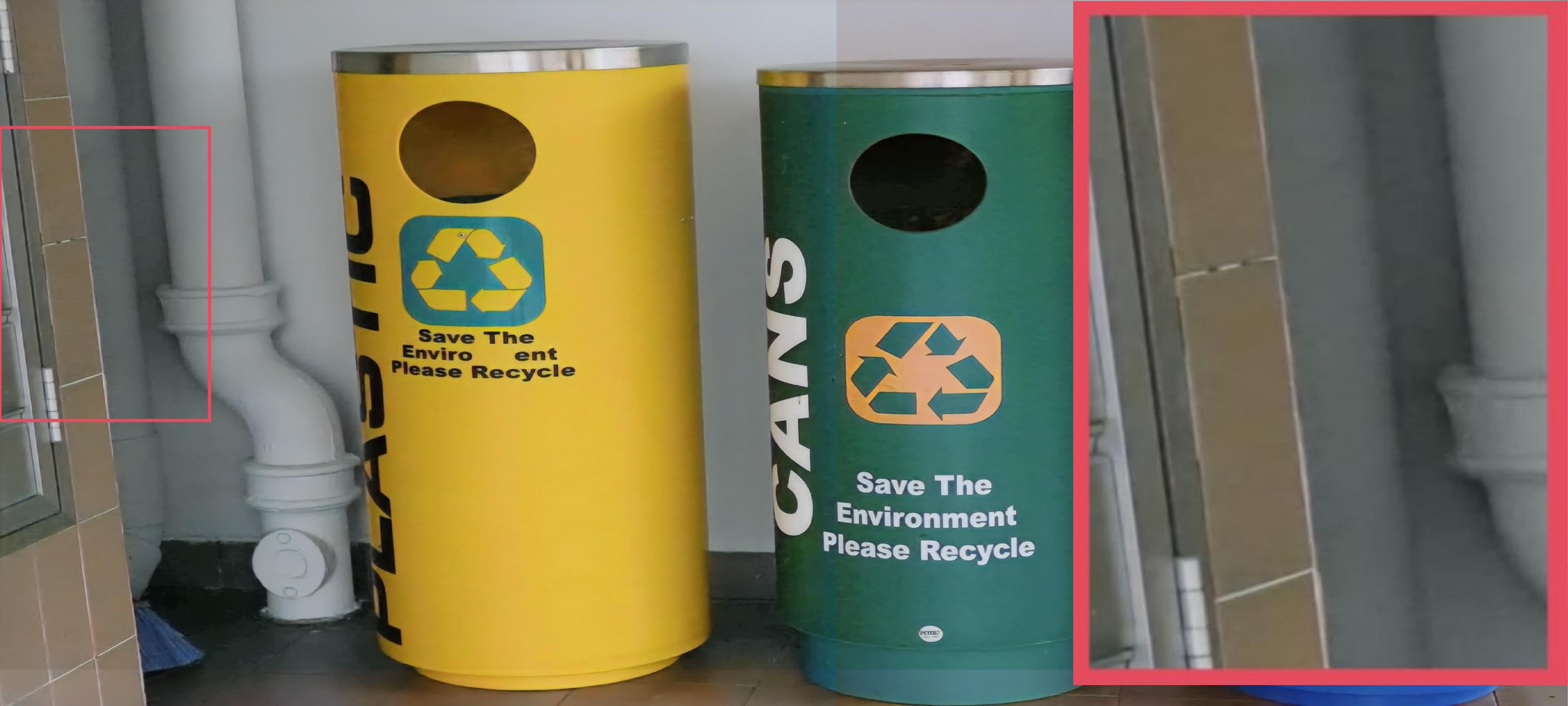} &
    \includegraphics[width=0.326\linewidth]{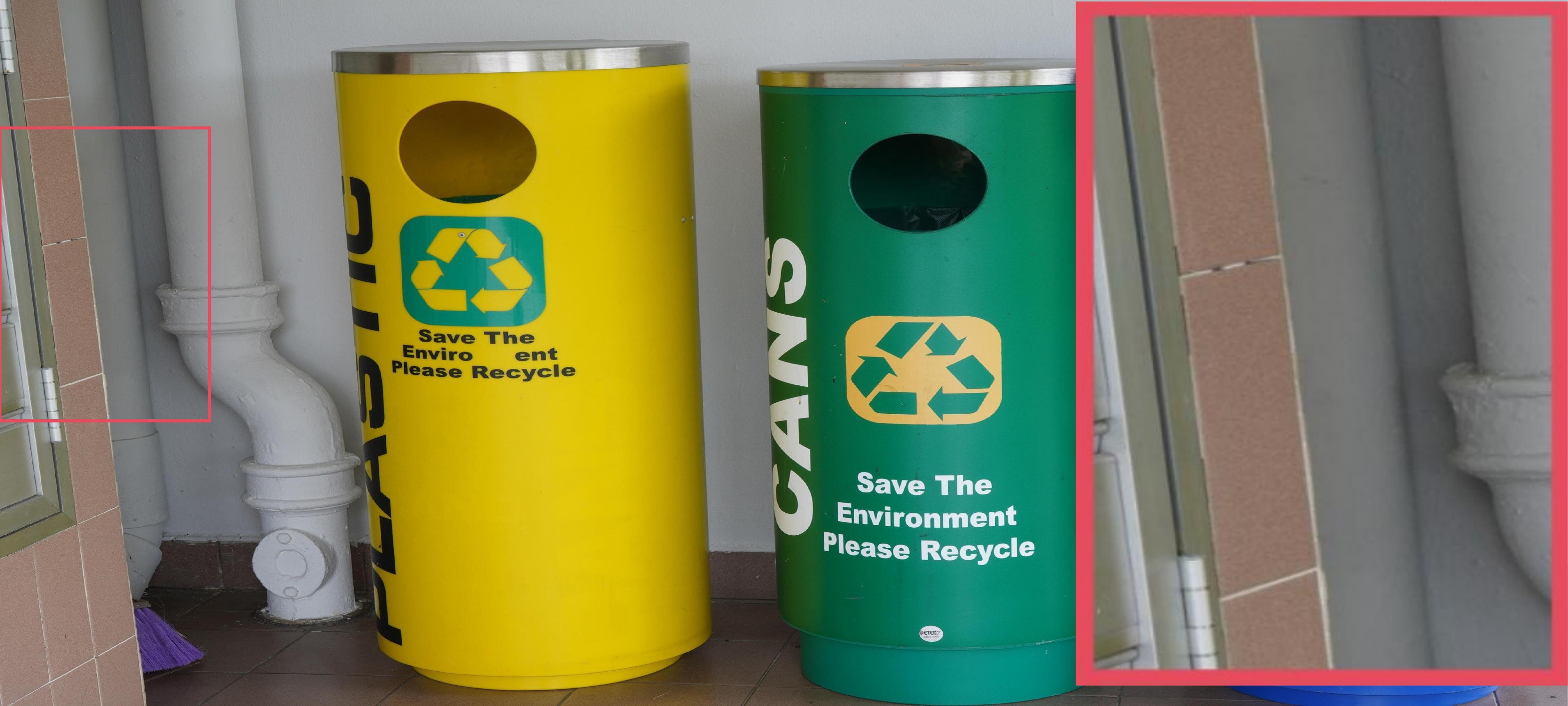} \\[-2.5pt]
    \scalebox{0.85}{(d) w/o CFE} &
    \scalebox{0.85}{\textbf{(e) w/ DFI (Ours)}} &
    \scalebox{0.85}{(f) GT}
\end{tabular}

\caption{Visual ablation of DFI, illustrating the complementary roles of S$^2$LFE and CFE.}
\label{fig:ablation of DFI}

\vspace{-4mm}
\end{figure}
\subsubsection{Internal Analysis of S$^2$LFE Block}

We further analyze the internal structure of S$^2$LFE, focusing on the effects of MaIR-based spatial modeling, GSR-based frequency modeling, and high-frequency guidance on low-frequency restoration. The corresponding results are reported in Table~\ref{tab:s2lfe}.
Compared with removing both branches (a), retaining only the spatial branch (b) improves PSNR from 24.96~dB to 27.21~dB, whereas retaining only the frequency branch (c) achieves a higher SSIM of 0.923. This indicates that spatial modeling contributes more to reconstruction fidelity, while frequency modeling is more effective in preserving structural information. Combining both branches (f) further achieves the best PSNR of 27.39~dB and the best SSIM of 0.929 (tied with (d)), validating their complementary integration. Moreover, replacing our spatial modeling with LFSS~\cite{zou2024wavemamba} or a Transformer block~\cite{LLFormer} leads to lower PSNR while introducing higher computational or parameter costs. Fig.~\ref{fig:ablation_s2lfe} shows the visual ablation results of S$^2$LFE, highlighting its contribution to global illumination recovery.

We further evaluate different allocations of low-frequency Mamba blocks across feature scales.
\begin{table}[t]
\centering
\begingroup
\caption{Internal analysis of the S$^2$LFE block on UHD-LL~\cite{li2023uhdfour}. FLOPs are measured with a $512\times512$ input. The best performance is marked in \textbf{bold}.}
\label{tab:s2lfe}
\centering
\small
\setlength{\tabcolsep}{1.7pt}
\resizebox{1.01\columnwidth}{!}{%
\begin{tabular}{c|c|cc|cc}
\Xhline{1pt}
\textbf{ID} & \textbf{Options} & \textbf{PSNR~$\uparrow$} & \textbf{SSIM~$\uparrow$} & \textbf{Params~$\downarrow$} & \textbf{FLOPs~$\downarrow$} \\
\Xhline{0.5pt}
(a) & w/o GSR; w/o MaIR & 24.96 & 0.899 & 3.54M & 29.12G  \\
(b) & w/o GSR & 27.21 & 0.916 & 4.32M & 39.53G  \\
(c) & w/o MaIR & 26.71 & 0.923 & 8.48M & 44.92G  \\
\Xhline{0.5pt}
(d) & MaIR $\rightarrow$ LFSS~\cite{zou2024wavemamba} & 26.35 & \textbf{0.929} & 10.47M & 59.01G  \\
(e) & MaIR $\rightarrow$ Transformer Block~\cite{LLFormer} & 26.70 & 0.911 & 12.48M & 57.06G  \\
\Xhline{0.5pt}
\rowcolor{gray!30}
\textbf{(f)} & \textbf{w/ S$^2$LFE (Ours)} & \textbf{27.39} & \textbf{0.929} & 9.26M  & 55.33G \\
\Xhline{1pt}
\end{tabular}
}
\endgroup
\vspace{-2mm}
\end{table}
\begin{figure}[!t]
\centering
\begingroup

\setlength{\tabcolsep}{1pt}

\begin{tabular}{@{}cccc@{}}
\includegraphics[width=0.2435\columnwidth]{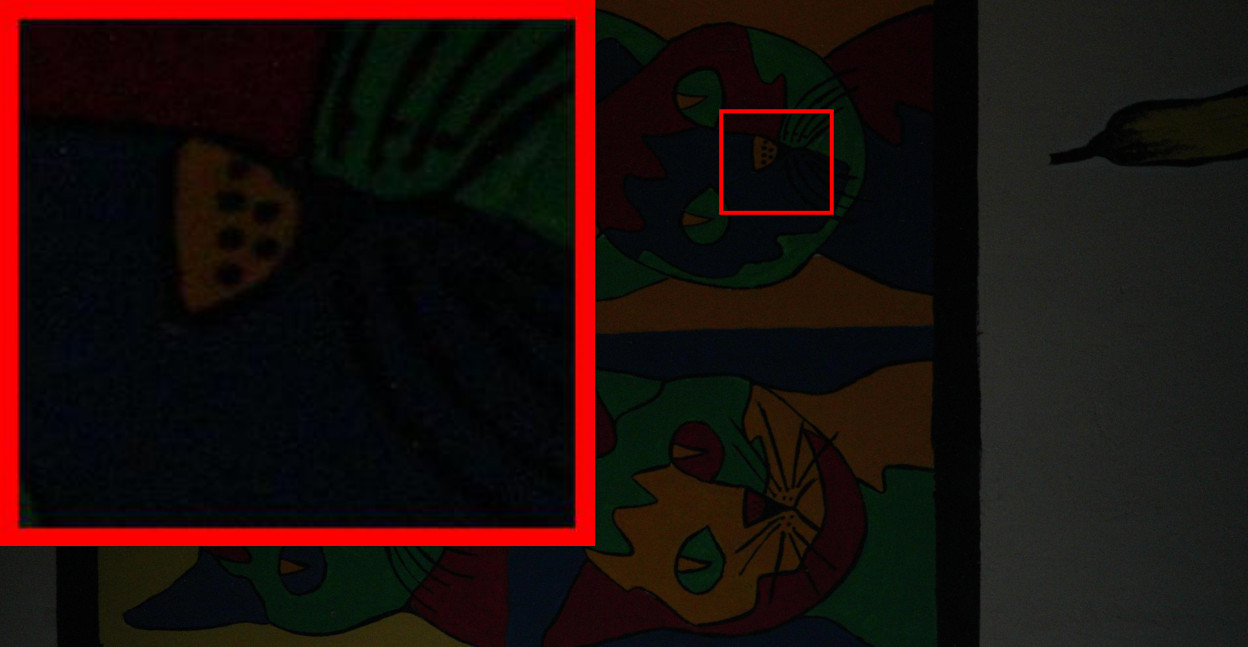}
&
\includegraphics[width=0.2435\columnwidth]{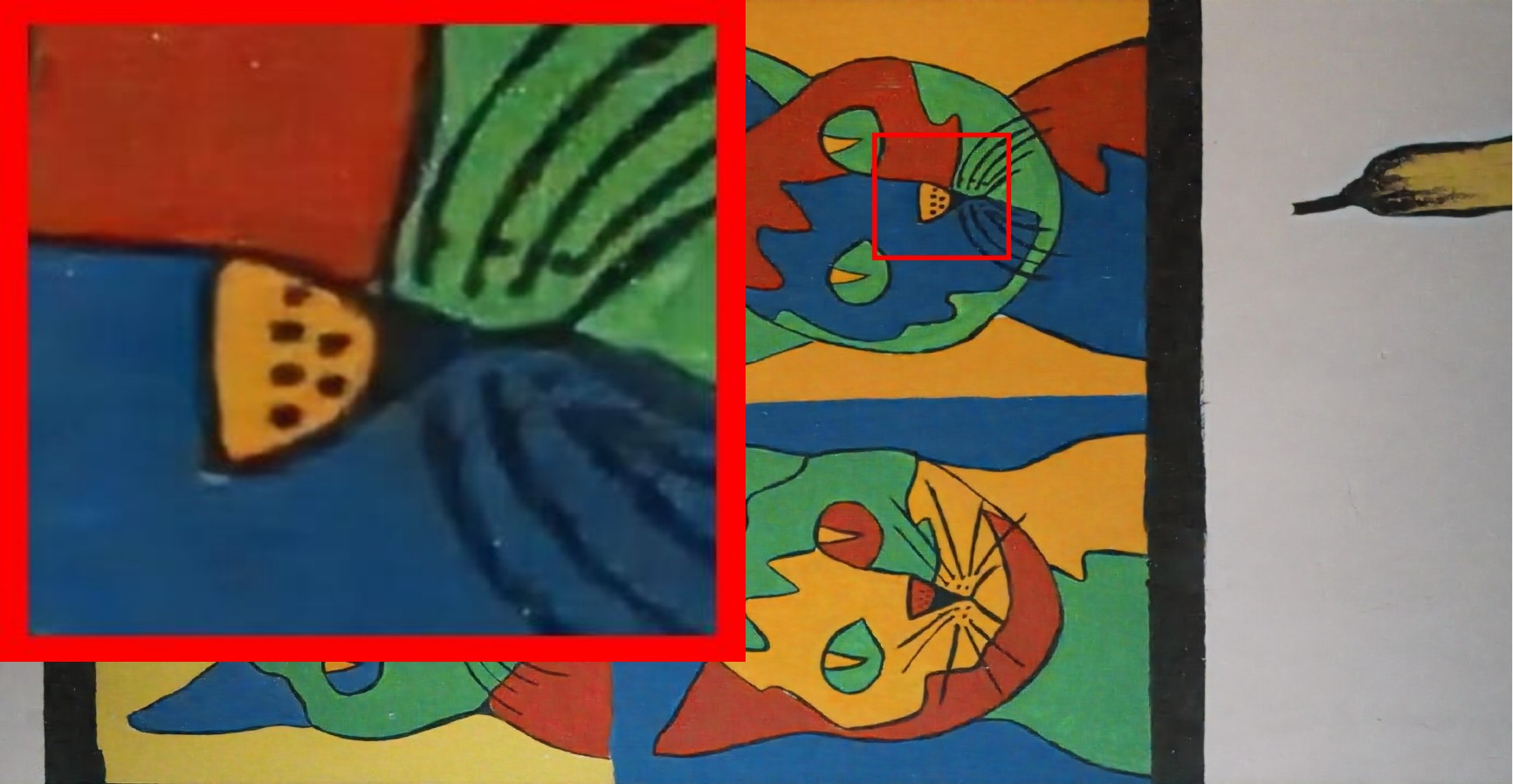}
&
\includegraphics[width=0.2435\columnwidth]{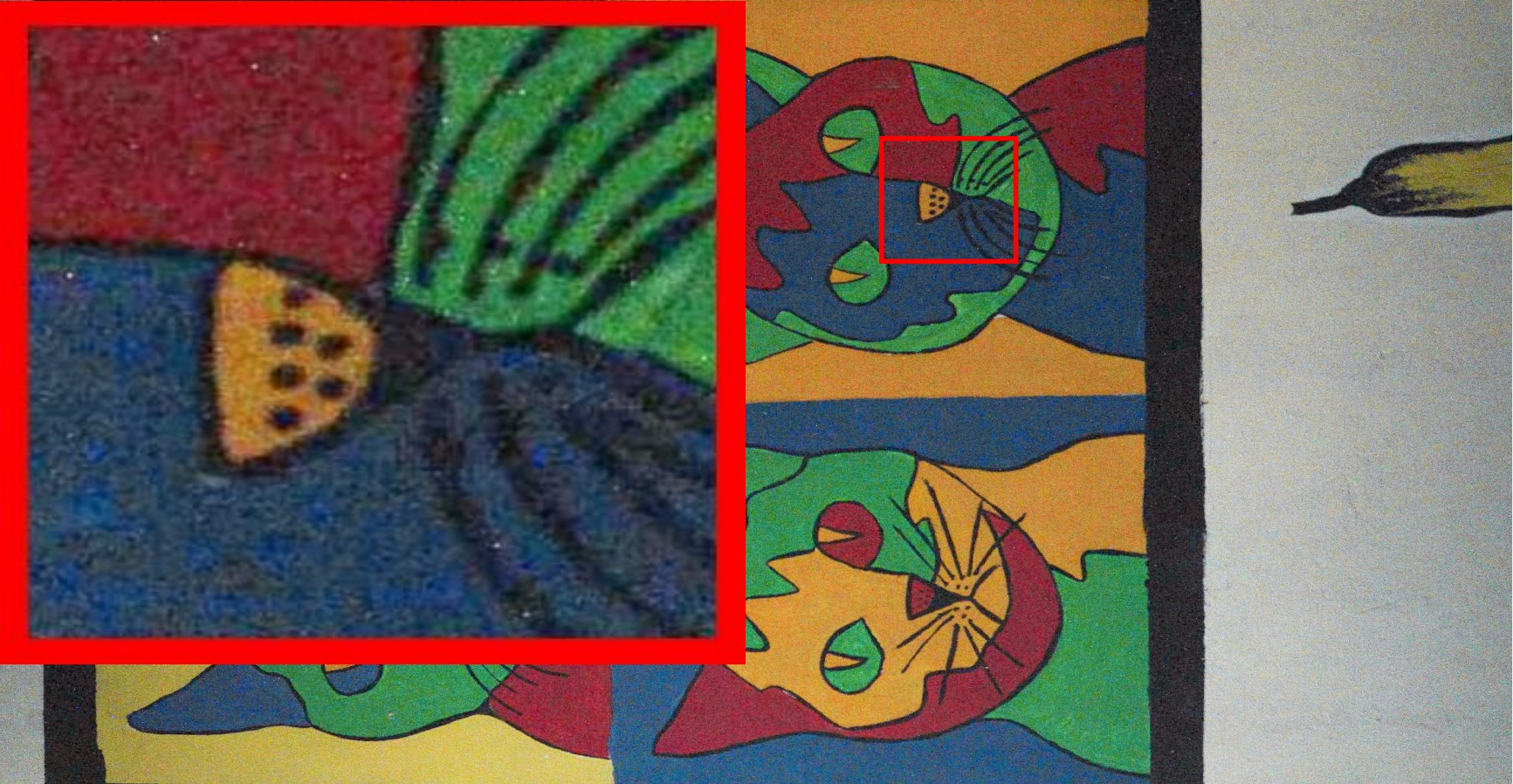}
&
\includegraphics[width=0.2435\columnwidth]{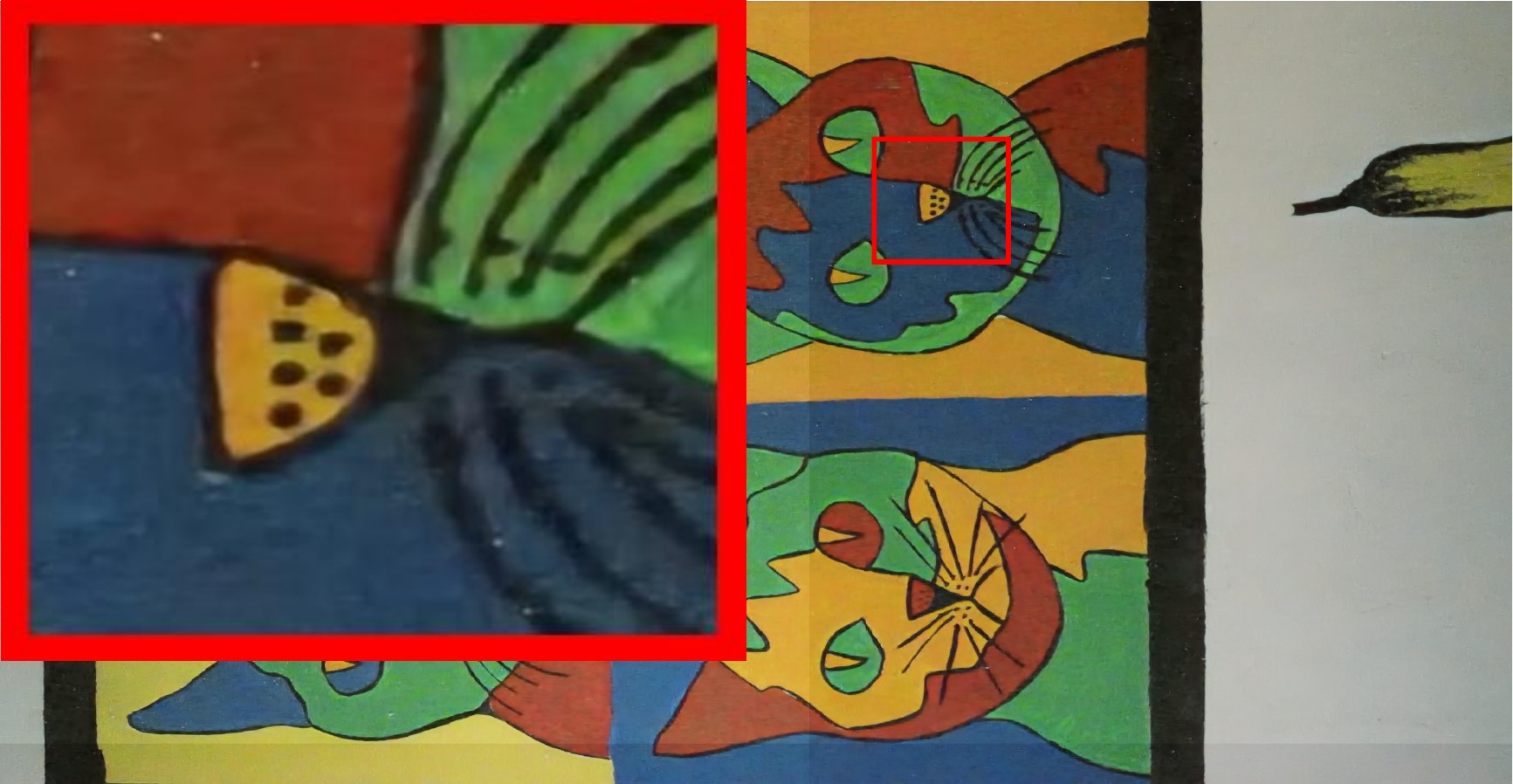}
\\[-5pt]

\parbox[t]{0.2435\columnwidth}{\centering
\fontsize{6pt}{6pt}\selectfont
(a) Input}
&
\parbox[t]{0.2435\columnwidth}{\centering
\fontsize{6pt}{6pt}\selectfont
(b) w/o MaIR; w/o GSR}
&
\parbox[t]{0.2435\columnwidth}{\centering
\fontsize{6pt}{6pt}\selectfont
(c) w/o GSR}
&
\parbox[t]{0.2435\columnwidth}{\centering
\fontsize{6pt}{6pt}\selectfont
(d) w/o MaIR}
\\[-0pt]

\includegraphics[width=0.2435\columnwidth]{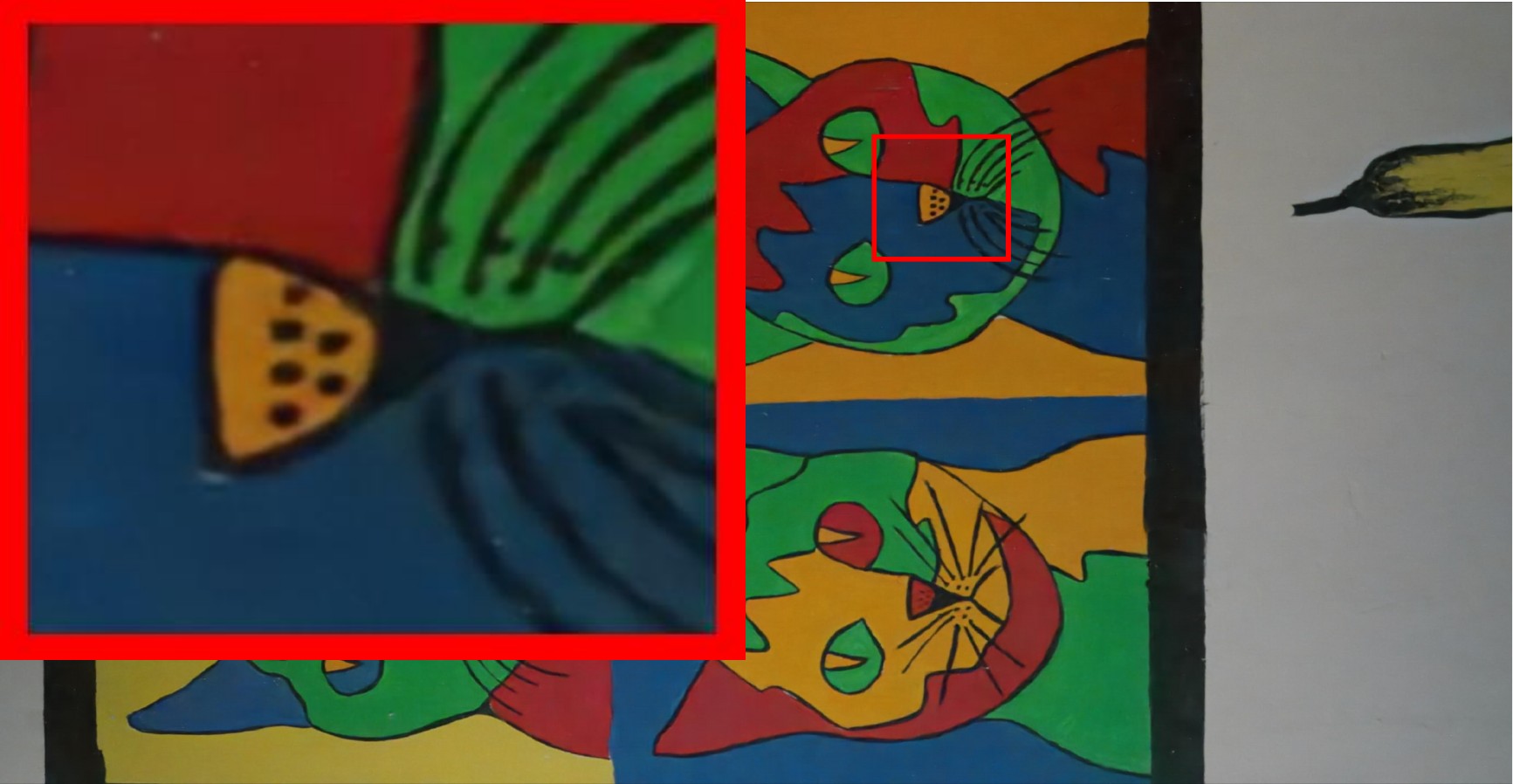}
&
\includegraphics[width=0.2435\columnwidth]{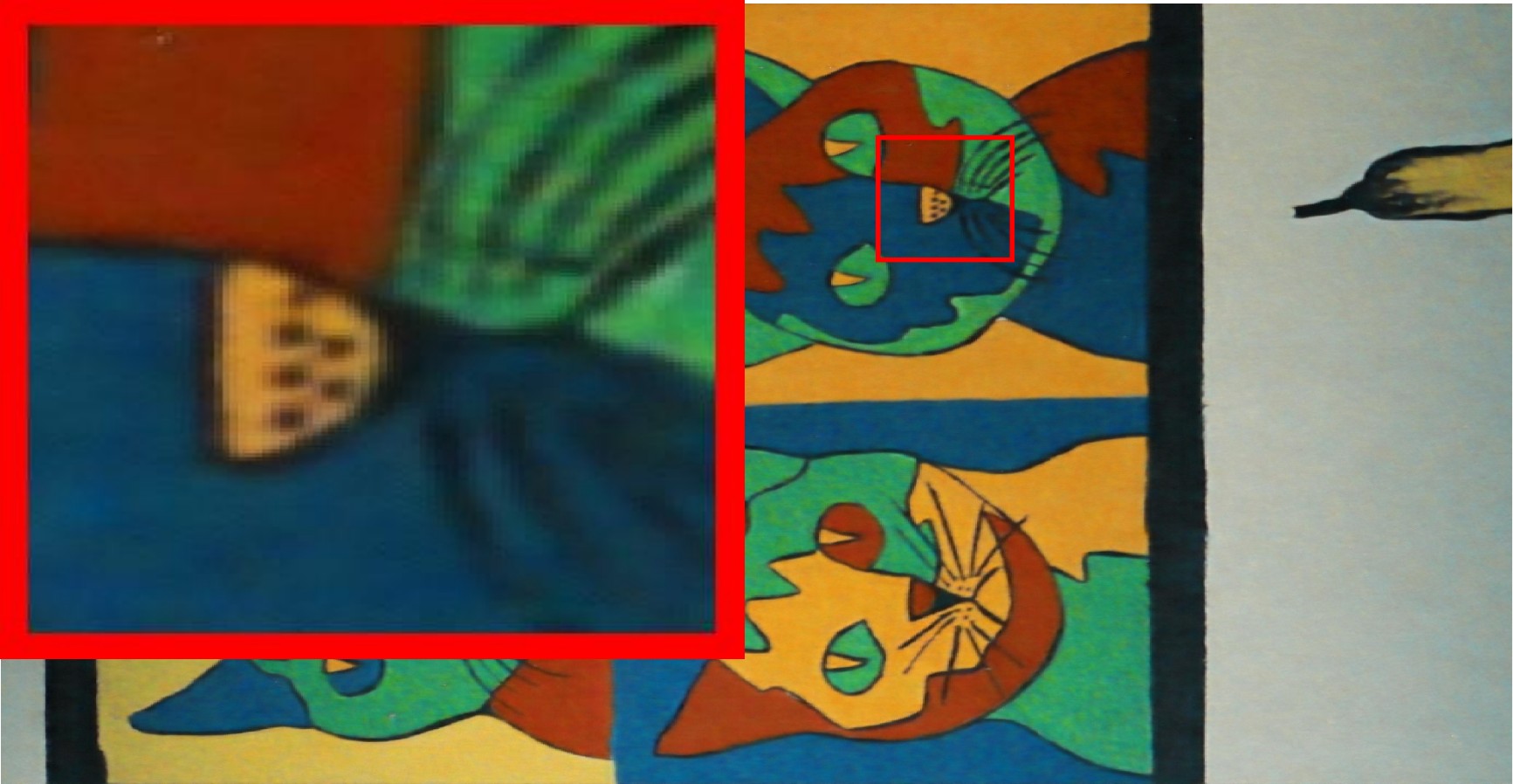}
&
\includegraphics[width=0.2435\columnwidth]{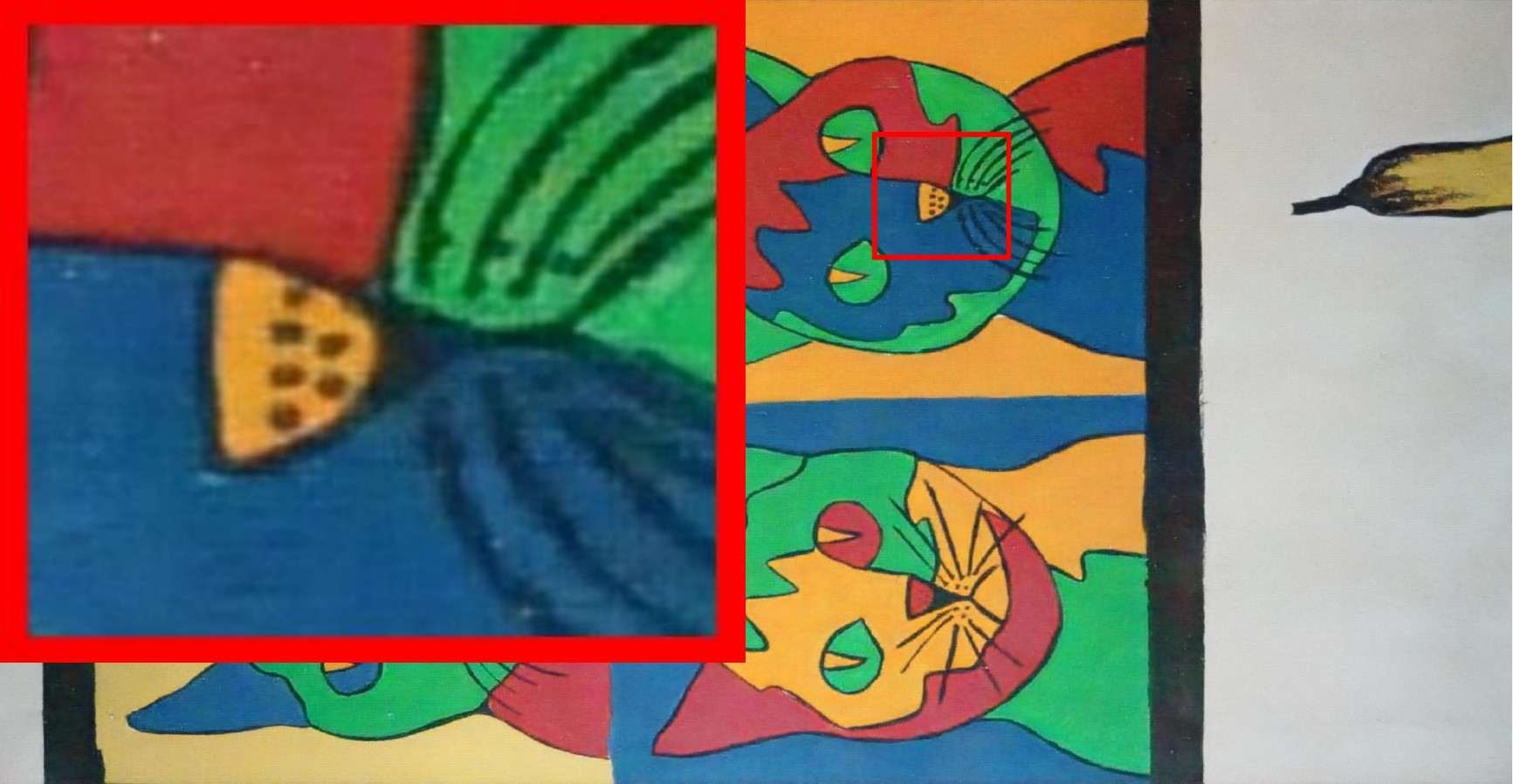}
&
\includegraphics[width=0.2435\columnwidth]{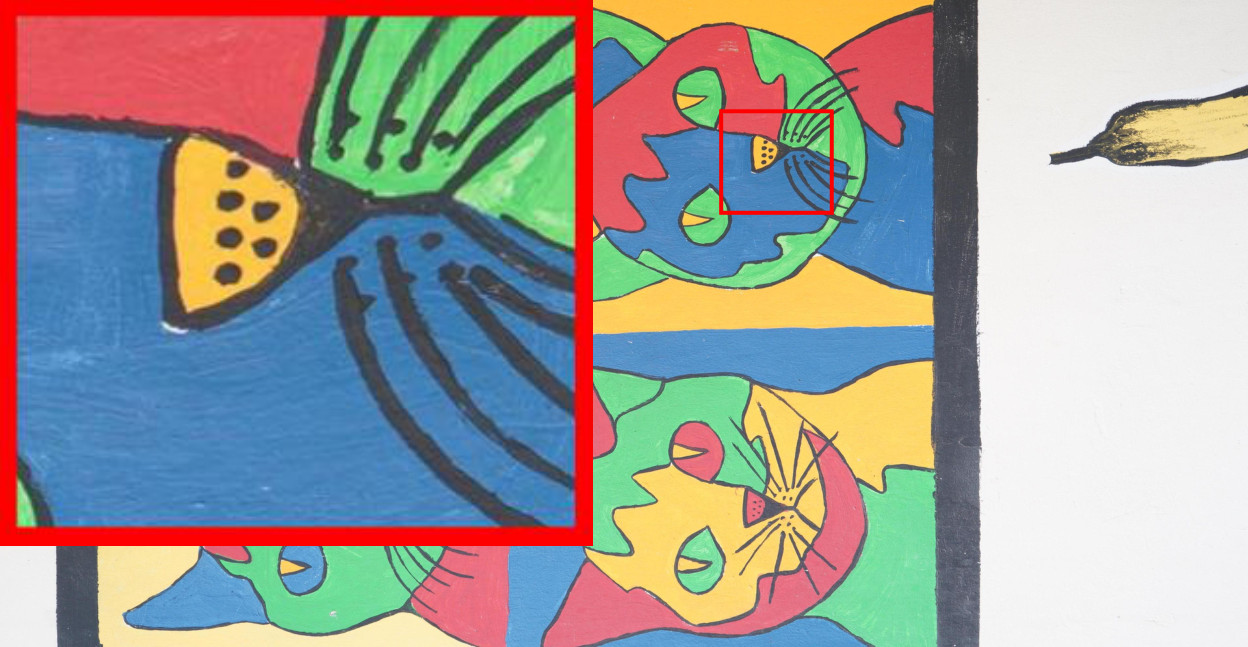}
\\[-5pt]

\parbox[t]{0.2435\columnwidth}{\centering
\fontsize{6pt}{6pt}\selectfont
(e) MaIR $\rightarrow$ LFSS}
&
\parbox[t]{0.2435\columnwidth}{\centering
\fontsize{6pt}{6pt}\selectfont
(f) MaIR $\rightarrow$ Transformer Block}
&
\parbox[t]{0.2435\columnwidth}{\centering
\fontsize{6pt}{6pt}\selectfont
\boldmath
(g) \textbf{w/ S$^2$LFE (Ours)}}
&
\parbox[t]{0.2435\columnwidth}{\centering
\fontsize{6pt}{6pt}\selectfont
(h) GT}
\\
\end{tabular}
\endgroup
\caption{Visual ablation of the S$^2$LFE block, highlighting its contribution to spatial--frequency feature modeling and detail restoration.}
\label{fig:ablation_s2lfe}
\end{figure}
\begin{table}[!t]
\centering

\begingroup
\caption{Ablation study on the number of low-frequency Mamba blocks on UHD-LL~\cite{li2023uhdfour}. FLOPs are measured with a $512\times512$ input. The best performance is marked in \textbf{bold}.}
\label{tab:dfi_blocks}
\centering
\small
\setlength{\tabcolsep}{7pt}
\resizebox{\columnwidth}{!}{%
\begin{tabular}{c|cc|cc}
\Xhline{1pt}
\textbf{Mamba Blocks} & \textbf{PSNR~$\uparrow$} & \textbf{SSIM~$\uparrow$} & \textbf{Params~$\downarrow$} & \textbf{FLOPs~$\downarrow$} \\
\Xhline{0.5pt}
$\{2,2,4,8\}$ & 26.67 & 0.920 & 8.45M & 47.84G  \\
$\{2,4,4,8\}$ & 26.49 & 0.919 & 8.62M & 52.08G  \\
\rowcolor{gray!30}
$\{2,4,6,8\}$ & \textbf{27.39} & \textbf{0.929} & 9.26M & 55.33G  \\
$\{2,6,6,8\}$ & 27.04 & 0.926 & 9.43M & 59.56G  \\
\Xhline{1pt}
\end{tabular}%
}
\vspace{-4mm}
\endgroup
\end{table}
\begin{figure}[!t]
\vspace{2mm}
\centering
\begingroup
\setlength{\tabcolsep}{1pt}
\begin{tabular}{ccc}
\includegraphics[width=0.327\columnwidth]{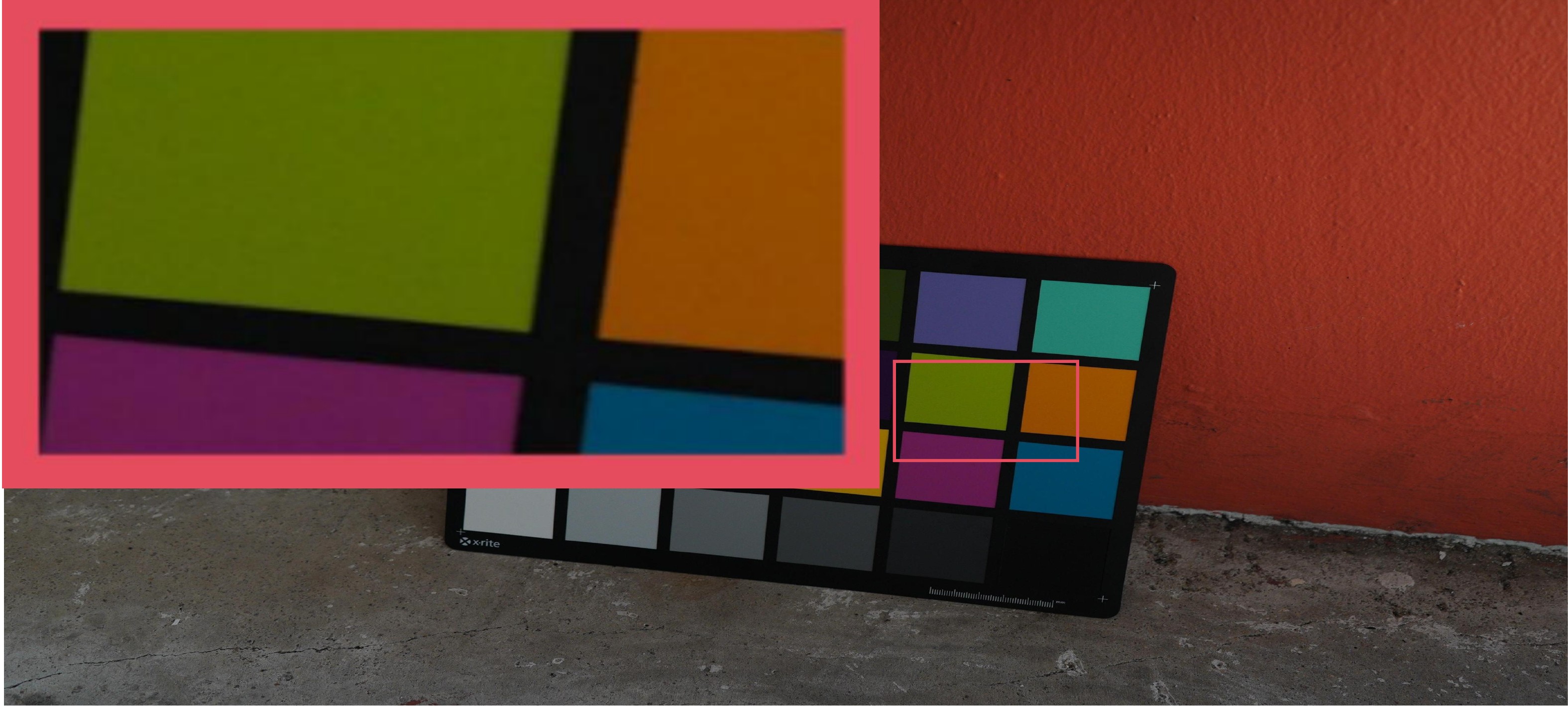}
&
\includegraphics[width=0.327\columnwidth]{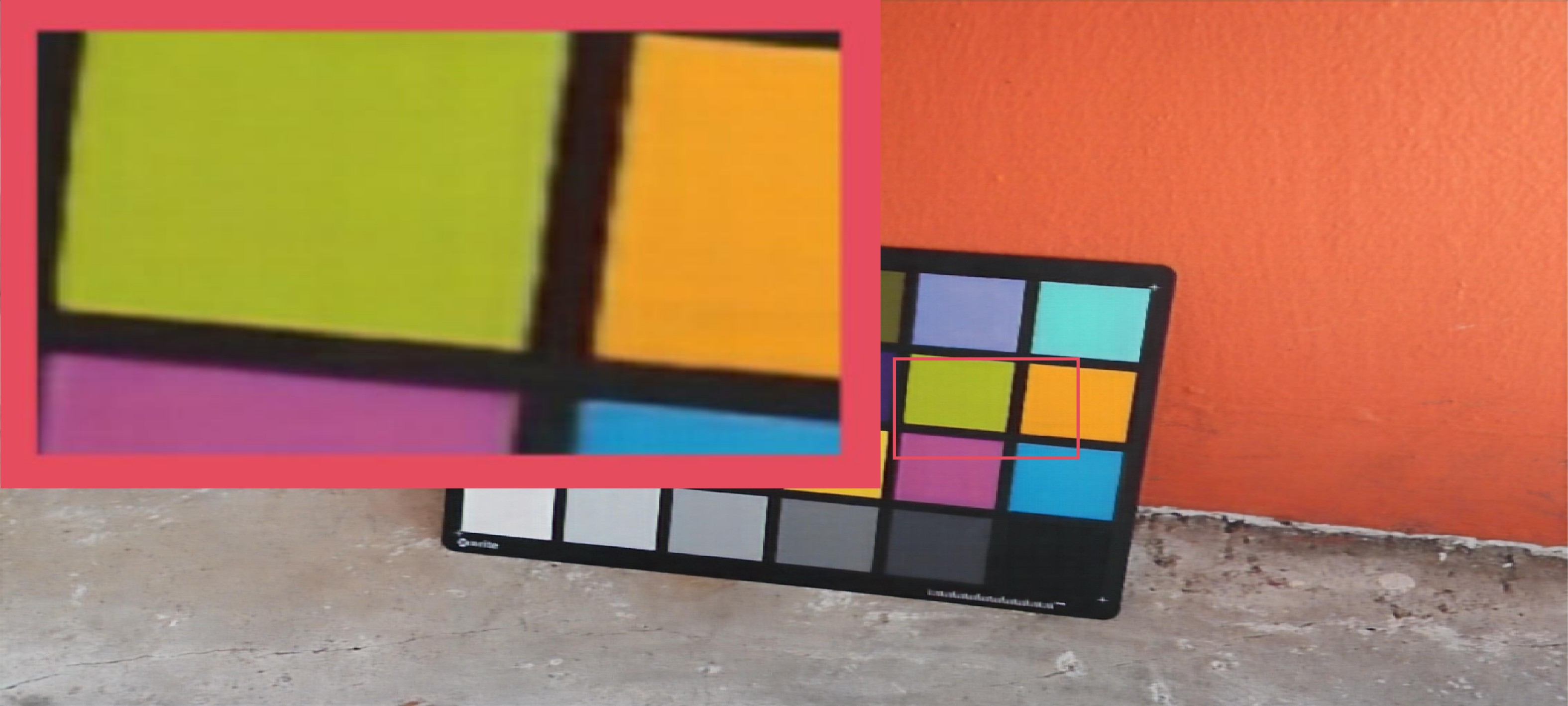}
&
\includegraphics[width=0.327\columnwidth]{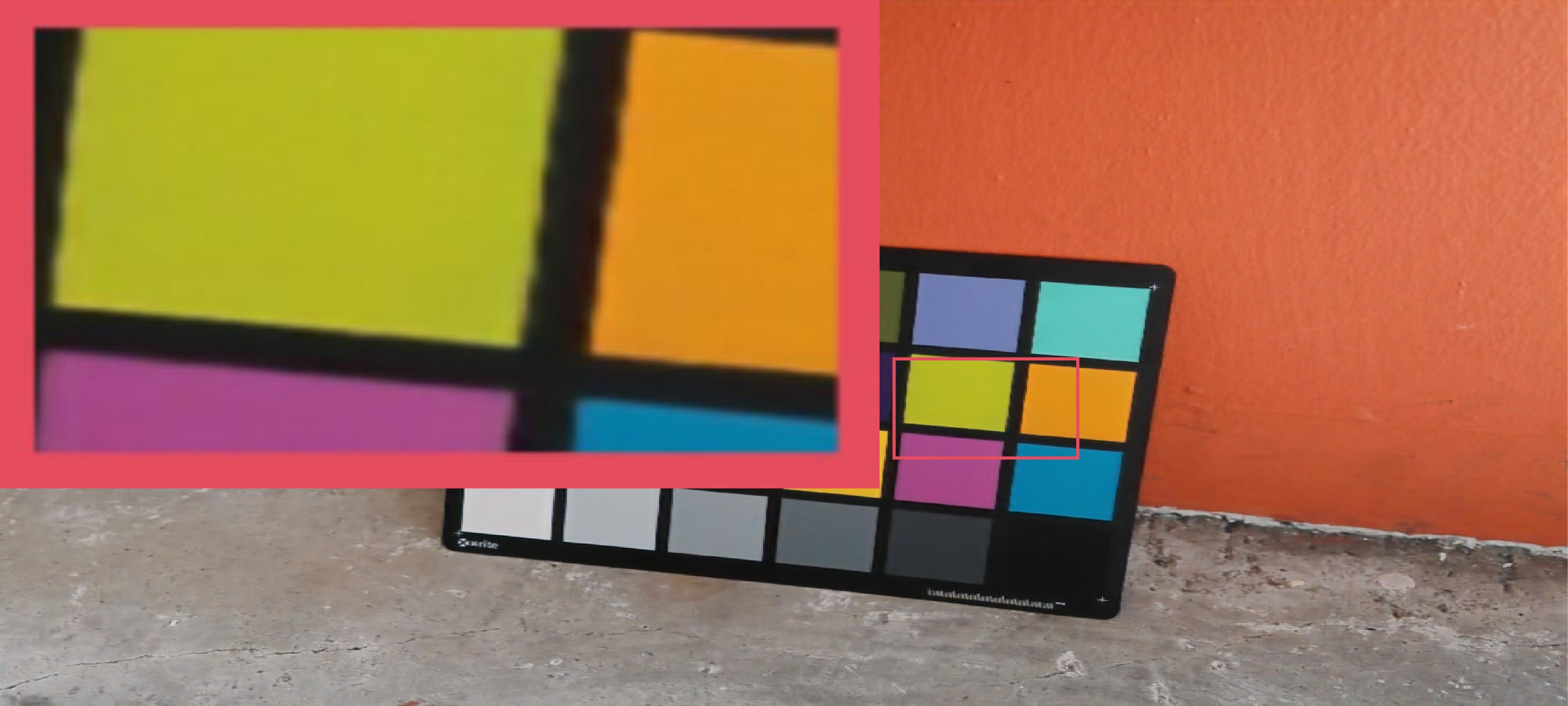}
\\[-4pt]
\parbox[t]{0.327\columnwidth}{\centering
\fontsize{8pt}{6pt}\selectfont
(a) Input}
&
\parbox[t]{0.327\columnwidth}{\centering
\fontsize{8pt}{6pt}\selectfont
(b) \{2, 2, 4, 8\} }
&
\parbox[t]{0.327\columnwidth}{\centering
\fontsize{8pt}{6pt}\selectfont
(c) \{2, 4, 4, 8\} }
\\[1pt]
\includegraphics[width=0.327\columnwidth]{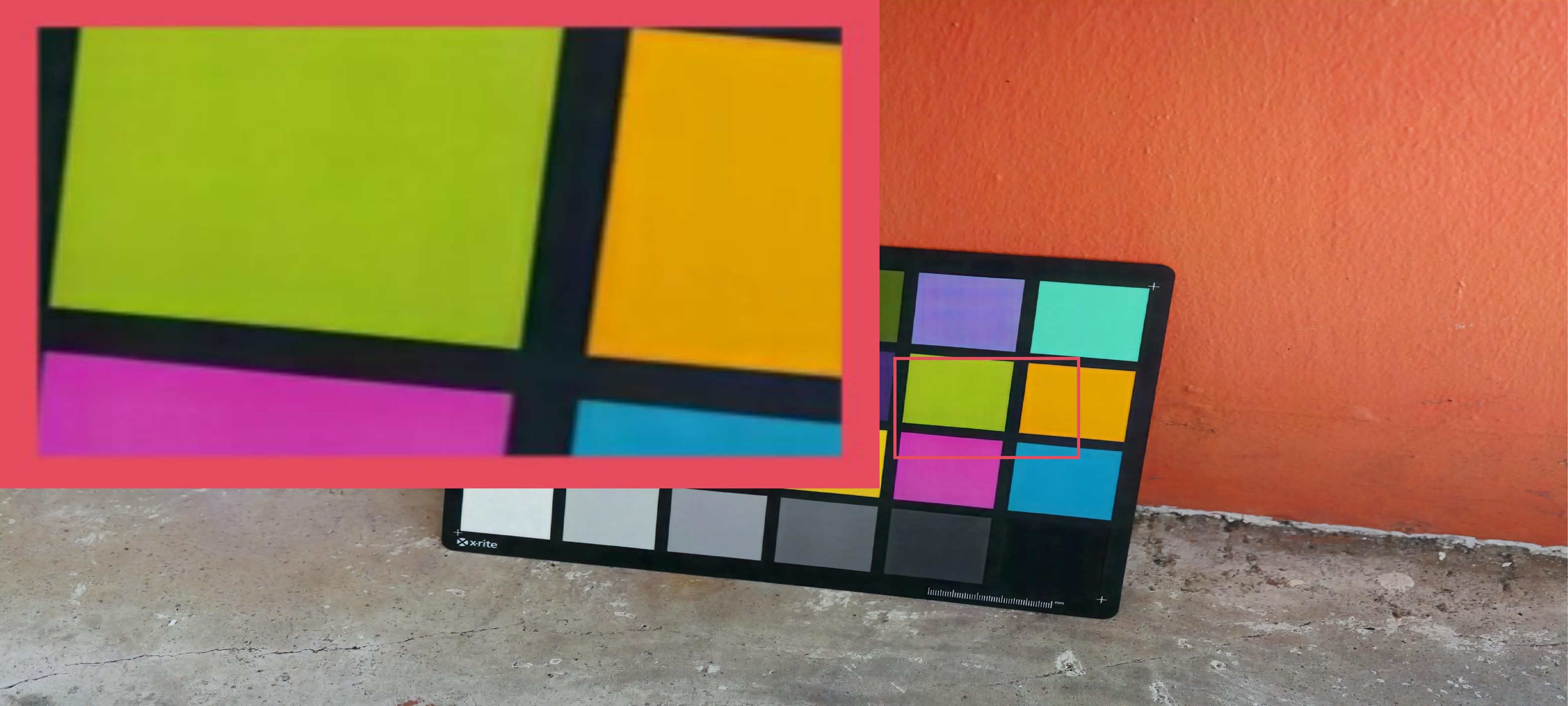}
&
\includegraphics[width=0.327\columnwidth]{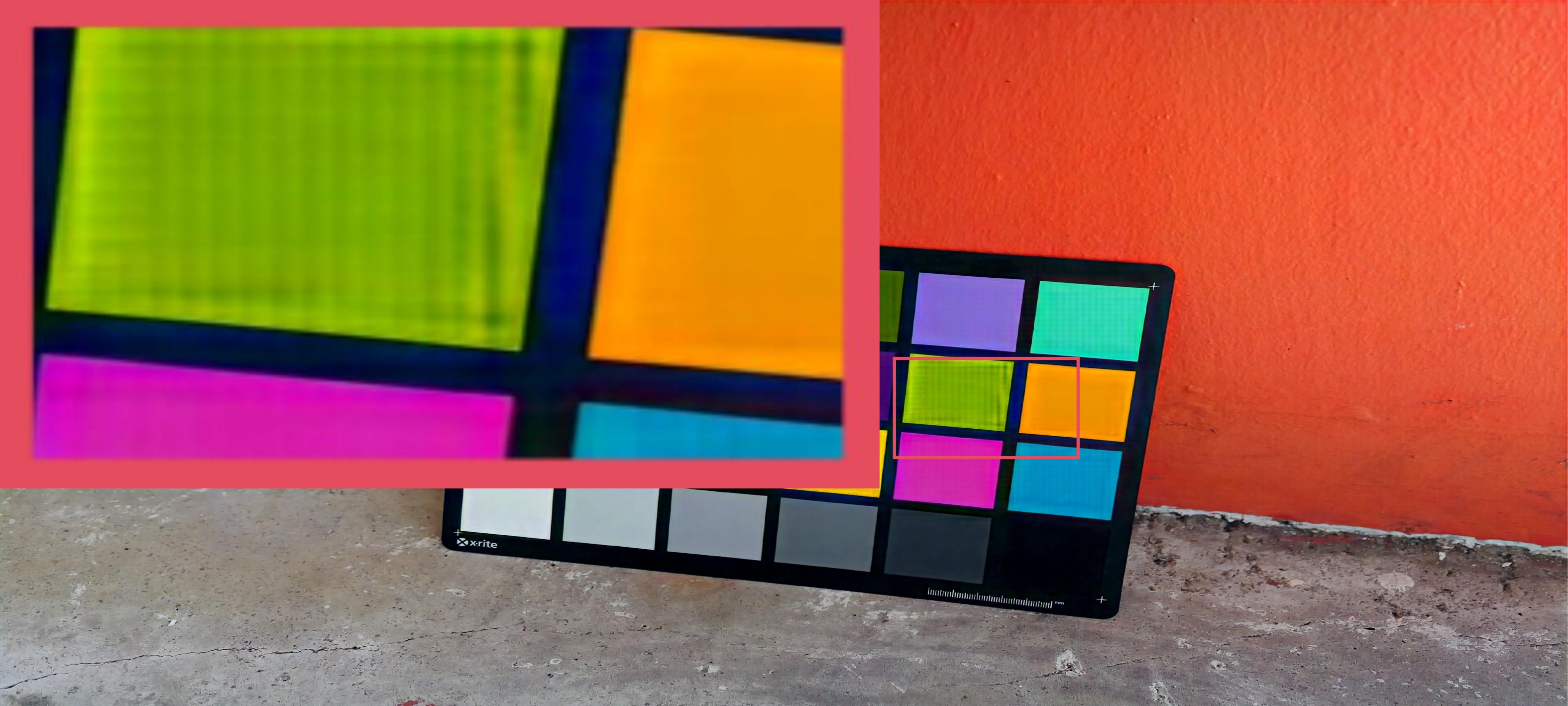}
&
\includegraphics[width=0.327\columnwidth]{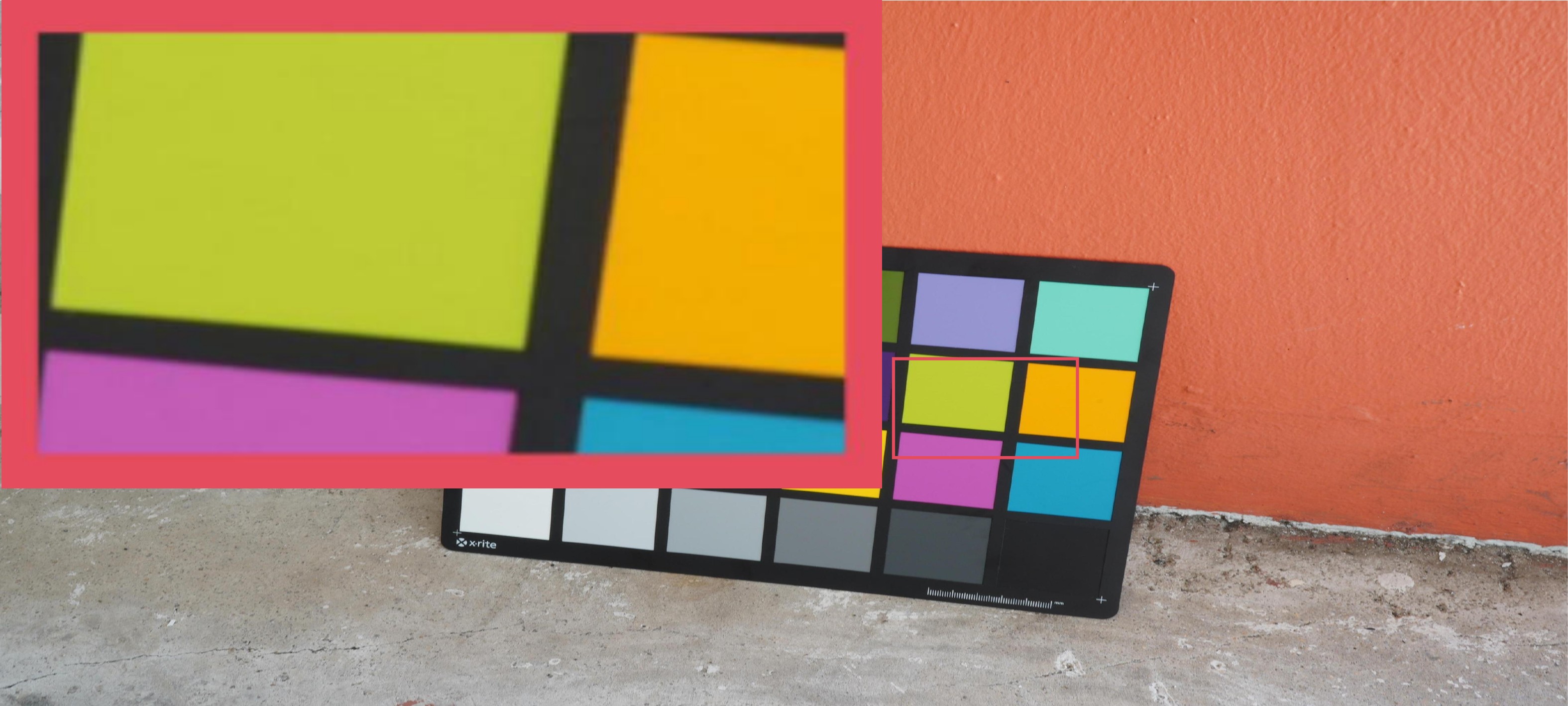}
\\[-5pt]
\parbox[t]{0.327\columnwidth}{\centering
\fontsize{8pt}{6pt}\selectfont
(d) \textbf{\{2, 4, 6, 8\} (Ours)}}
&
\parbox[t]{0.327\columnwidth}{\centering
\fontsize{8pt}{6pt}\selectfont
(e) \{2, 6, 6, 8\}}
&
\parbox[t]{0.327\columnwidth}{\centering
\fontsize{8pt}{6pt}\selectfont
(f) GT}
\\
\end{tabular}
\endgroup
\caption{Visual ablation of the number of MaIR blocks, highlighting their contribution to spatial--frequency feature modeling and detail restoration.}
\label{fig:ablation of block}
\end{figure}
As shown in Table~\ref{tab:dfi_blocks} and Fig.~\ref{fig:ablation of block}, the $\{2,4,6,8\}$ configuration achieves the best performance, reaching 27.39~dB PSNR and 0.929 SSIM. Increasing the number of blocks does not consistently improve restoration quality; for example, $\{2,6,6,8\}$ incurs higher computational cost but yields lower performance. This suggests that a balanced multi-scale allocation is more important than simply increasing network depth.
Notably, the PSNR gap among different allocations is considerably larger than the SSIM gap, suggesting that the block allocation mainly affects global illumination and color accuracy rather than structural fidelity.
Therefore, $\{2,4,6,8\}$ is adopted as the final configuration.

\subsubsection{Internal Analysis of CFE Block}

For the CFE block, we further compare different high-frequency refinement
strategies and analyze the effect of low-frequency guidance on high-frequency
reconstruction. The corresponding results are shown in Table~\ref{tab:cfe}.
\begin{table}[!t]
\centering

\begingroup
\caption{Internal analysis of the CFE block on UHD-LL~\cite{li2023uhdfour}. The best performance is marked in \textbf{bold}.}
\label{tab:cfe}
\centering
\footnotesize
\setlength{\tabcolsep}{4.8pt}
\resizebox{1.01\columnwidth}{!}{%
\begin{tabular}{c|c|cc|cc}
\Xhline{1pt}
\textbf{ID} & \textbf{Options} & \textbf{PSNR~$\uparrow$} & \textbf{SSIM~$\uparrow$} & \textbf{Params~$\downarrow$} & \textbf{FLOPs~$\downarrow$}  \\
\Xhline{0.5pt}
(a) & CFE $\rightarrow$ HFE~\cite{zou2024wavemamba} & 27.13 & 0.925 & 9.16M & 56.27G  \\
(b) & CFE $\rightarrow$ HFRM~\cite{jiang2023diffll} & 26.96 & 0.922 & 9.25M & 55.20G  \\
\Xhline{0.5pt}
\rowcolor{gray!30}
\textbf{(c)} & \textbf{w/ CFE (Ours)} & \textbf{27.39} & \textbf{0.929} & 9.26M & 55.33G  \\
\Xhline{1pt}
\end{tabular}
}
\vspace{-2mm}
\endgroup
\end{table}
%
%
%
\begin{figure}[!t]
\vspace{2mm}
\centering
\begingroup
\setlength{\tabcolsep}{1pt}
\begin{tabular}{ccccc}
\includegraphics[width=0.1925\columnwidth]{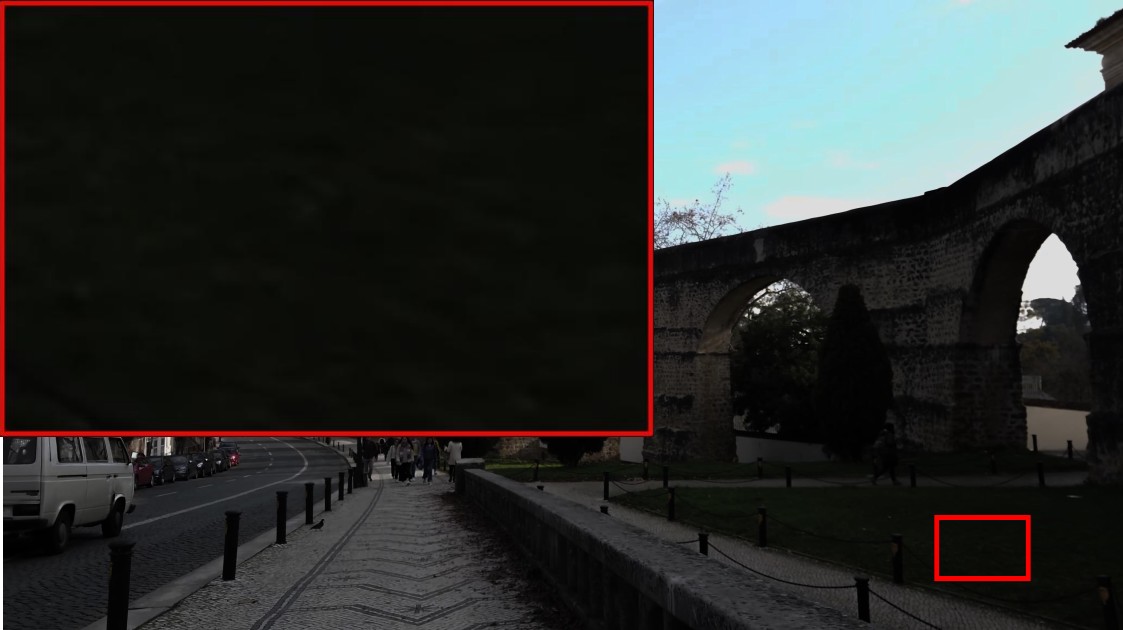}
&
\includegraphics[width=0.1925\columnwidth]{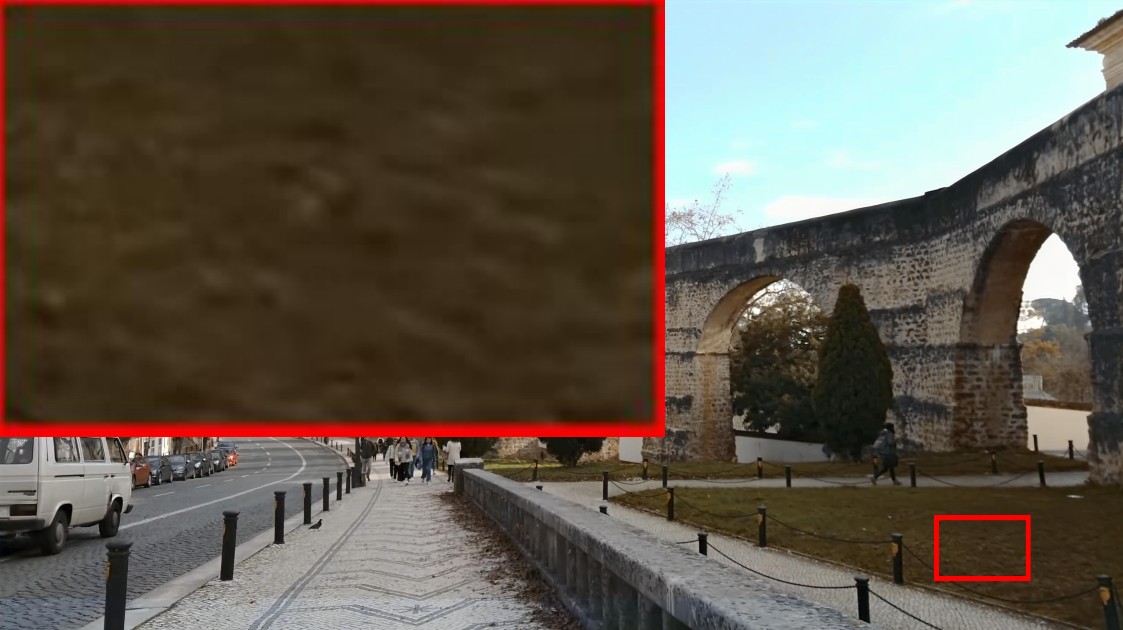}
&
\includegraphics[width=0.1925\columnwidth]{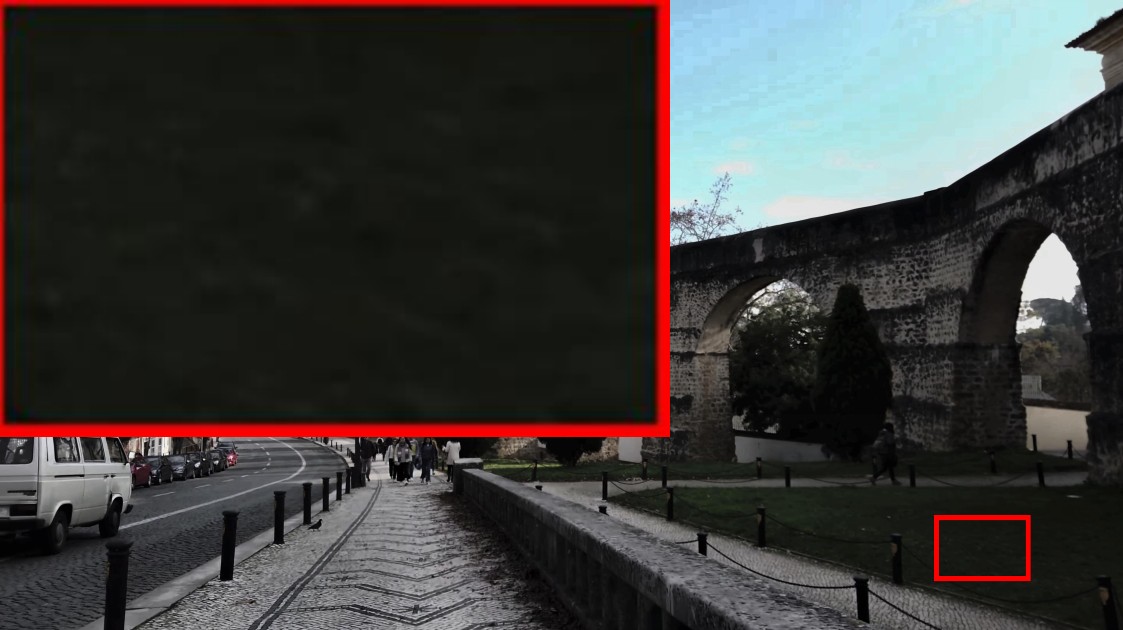}
&
\includegraphics[width=0.1925\linewidth]{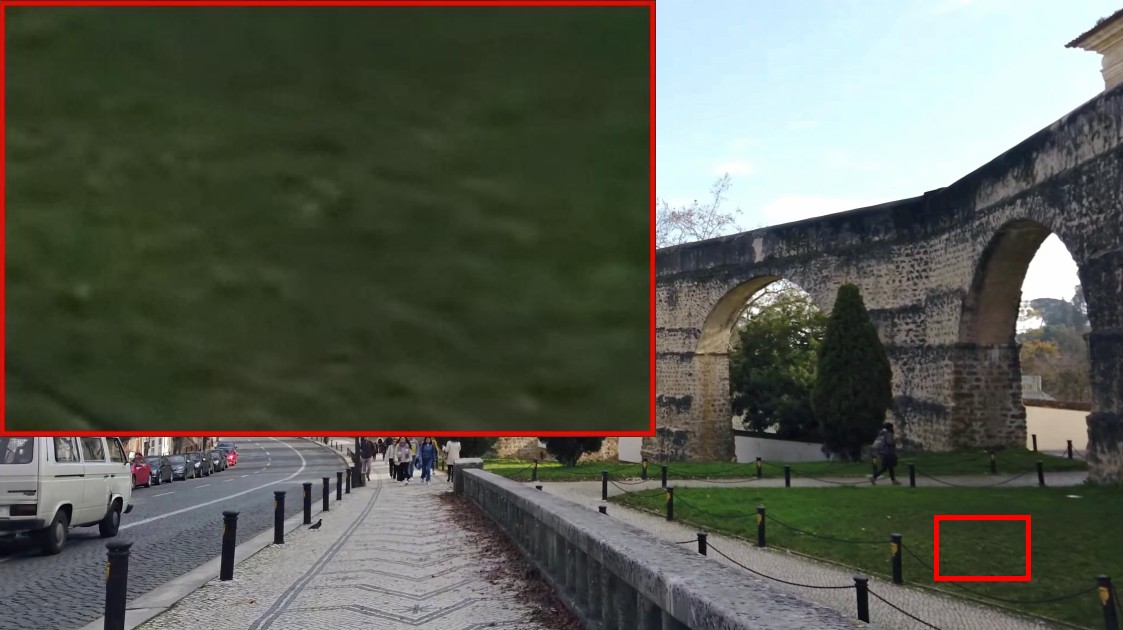}
&
\includegraphics[width=0.1925\linewidth]{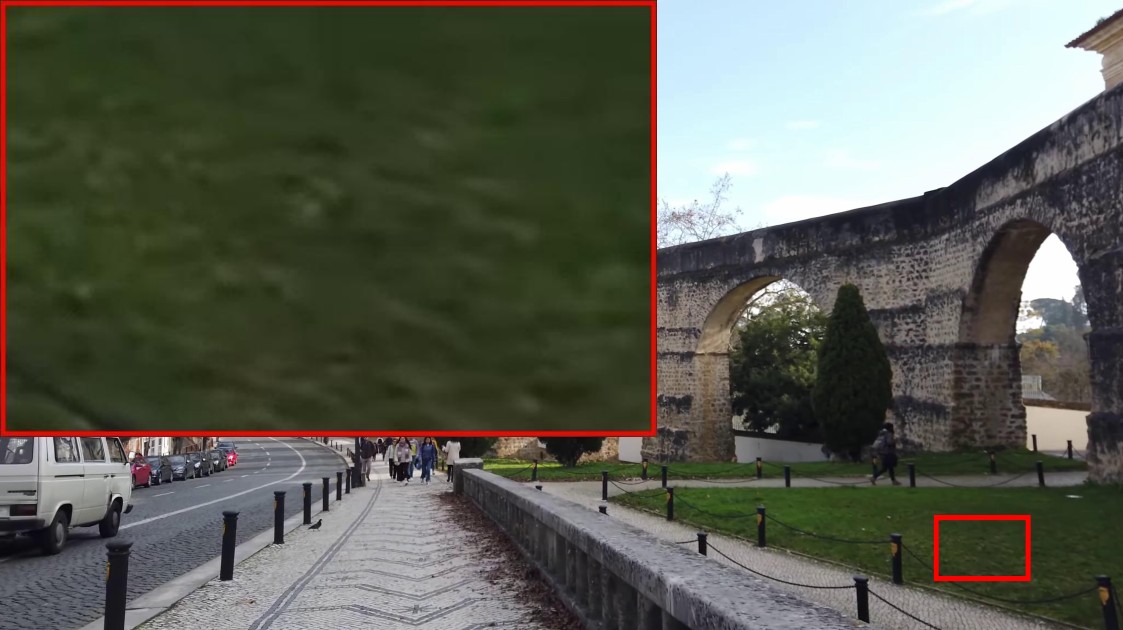}
\\[-4pt]
\parbox[t]{0.1925\columnwidth}{\centering
\fontsize{6.2pt}{6pt}\selectfont
(a) Input}
&
\parbox[t]{0.1925\columnwidth}{\centering
\fontsize{6.2pt}{6pt}\selectfont
(b) CFE $\rightarrow$ HFE }
&
\parbox[t]{0.1925\columnwidth}{\centering
\fontsize{6.2pt}{6pt}\selectfont
(c) CFE $\rightarrow$ HFRM }
&
\parbox[t]{0.1925\columnwidth}{\centering
\fontsize{6.2pt}{6pt}\selectfont
(d) \textbf{w/ CFE (Ours)} }
&
\parbox[t]{0.1925\columnwidth}{\centering
\fontsize{6.2pt}{6pt}\selectfont
(e) GT }
\end{tabular}
\endgroup
\caption{Visual ablation of CFE, highlighting its contribution to low-frequency-guided high-frequency detail refinement.}
\label{fig:ablation of CFE}
\end{figure}
The proposed CFE achieves the best performance while maintaining comparable parameter complexity to the HFE~\cite{zou2024wavemamba} and HFRM~\cite{jiang2023diffll} alternatives ((a)--(b)), demonstrating the effectiveness of its cross-frequency refinement strategy. Fig.~\ref{fig:ablation of CFE} shows the visual ablation results of CFE, demonstrating its effectiveness in high-frequency detail refinement. These results confirm that explicitly leveraging restored low-frequency representations is beneficial for more reliable high-frequency detail reconstruction.

\subsubsection{Effect of the Bidirectional Guidance Mechanism }

\begin{table}[!t]
\centering

\begingroup
\caption{Effect of the bidirectional guidance mechanism on UHD-LL~\cite{li2023uhdfour} dataset. The best performance is marked in \textbf{bold}.}
\label{tab:biguidance}
\centering
\footnotesize
\setlength{\tabcolsep}{4.5pt}
\resizebox{\columnwidth}{!}{%
\begin{tabular}{c|c|cc|cc}
\Xhline{1pt}
\textbf{ID} & \textbf{Options} & \textbf{PSNR~$\uparrow$} & \textbf{SSIM~$\uparrow$} & \textbf{Params~$\downarrow$} & \textbf{FLOPs~$\downarrow$}  \\
\Xhline{0.5pt}
(a) & w/o High-Freq Guidance & 27.08 & 0.924 & 8.50M & 53.18G  \\
(b) & w/o Low-Freq Guidance & 27.17 & 0.925 & 9.26M & 55.27G  \\
\Xhline{0.5pt}
\rowcolor{gray!30}
\textbf{(c)} & \textbf{Bidirectional Guidance} & \textbf{27.39} & \textbf{0.929} & 9.26M  & 55.33G \\
\Xhline{1pt}
\end{tabular}%
}
\vspace{-2mm}
\endgroup
\end{table}







\begin{figure}[!t]
\vspace{2mm}
\centering
\begingroup
\setlength{\tabcolsep}{1pt}
\begin{tabular}{ccccc}
\includegraphics[width=0.193\columnwidth]{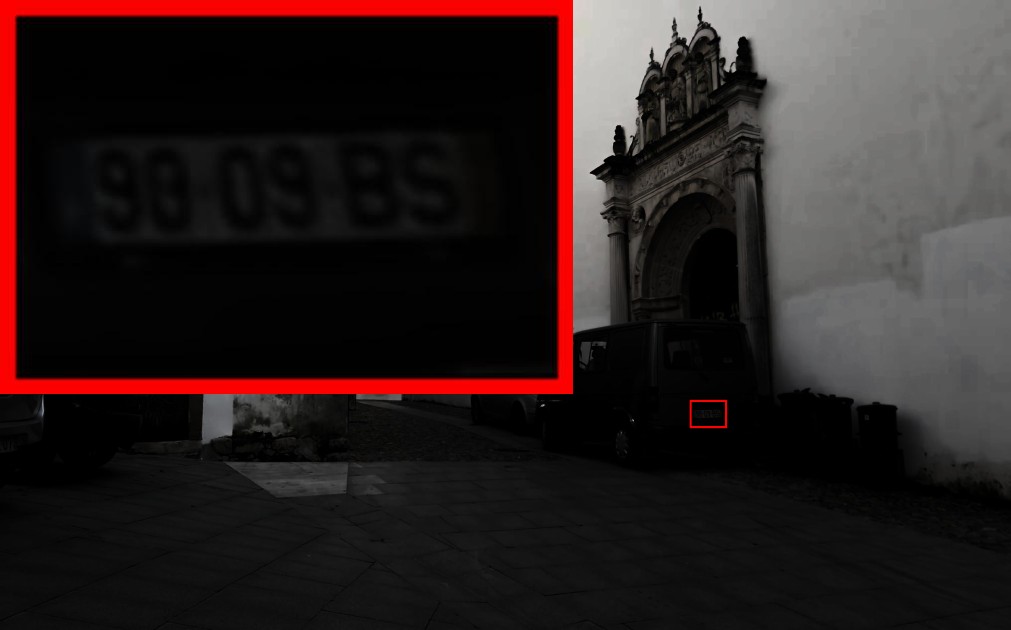}
&
\includegraphics[width=0.193\columnwidth]{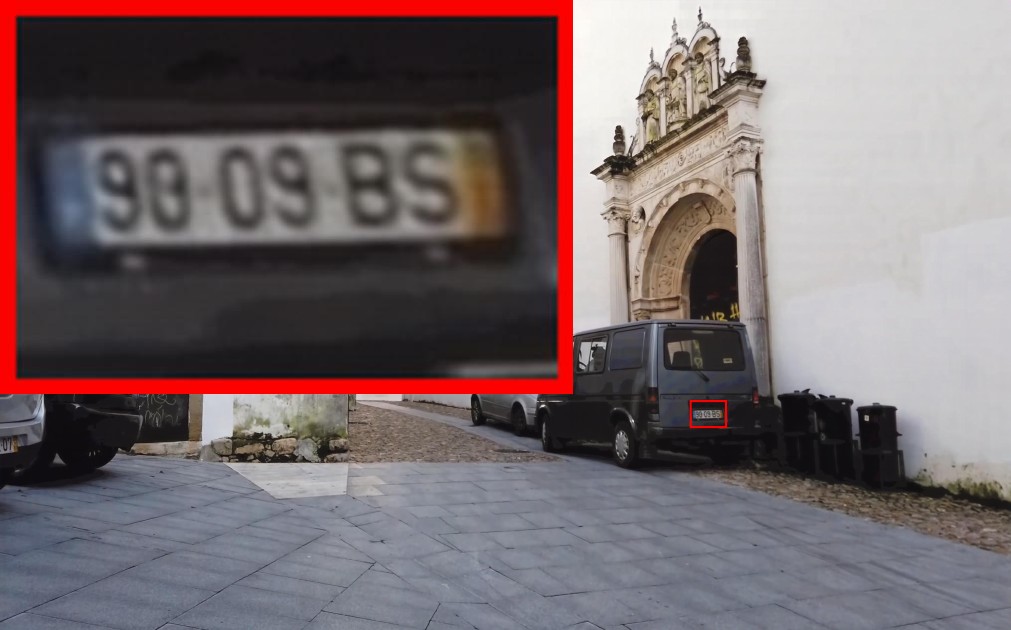}
&
\includegraphics[width=0.193\columnwidth]{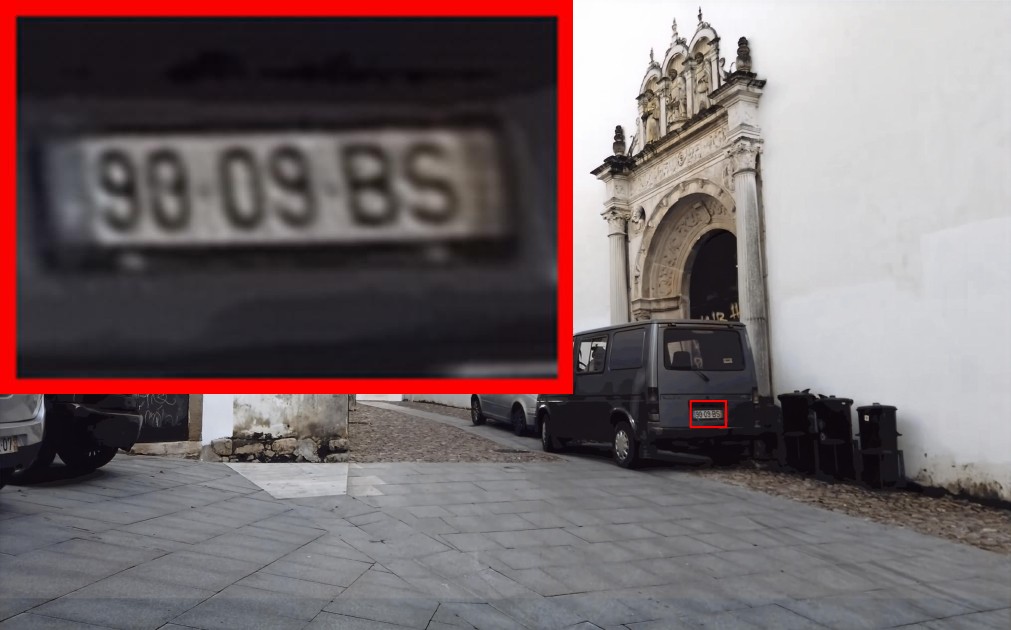}
&
\includegraphics[width=0.193\linewidth]{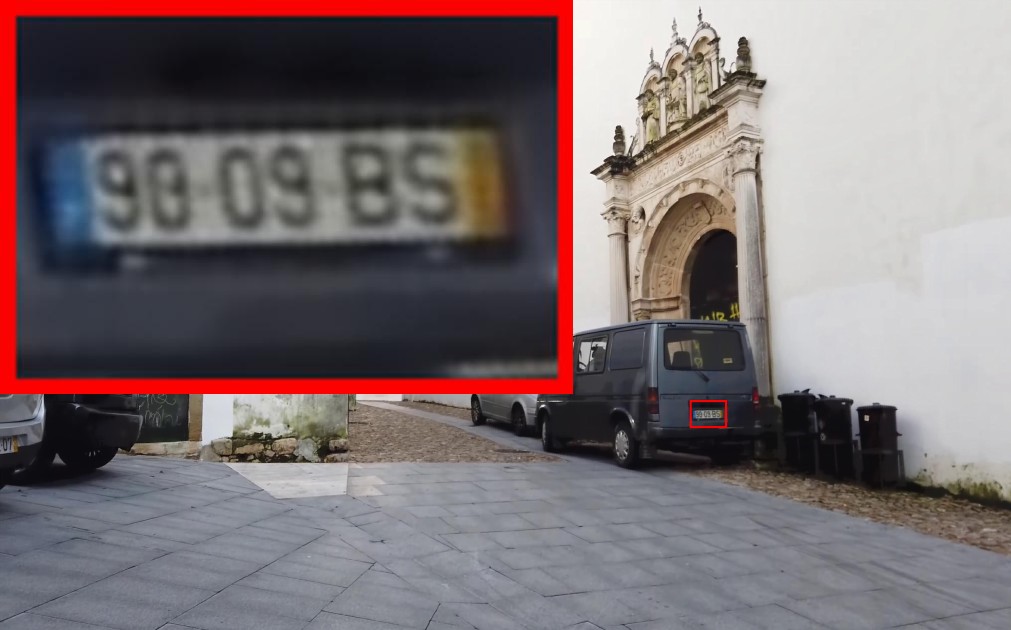}
&
\includegraphics[width=0.193\linewidth]{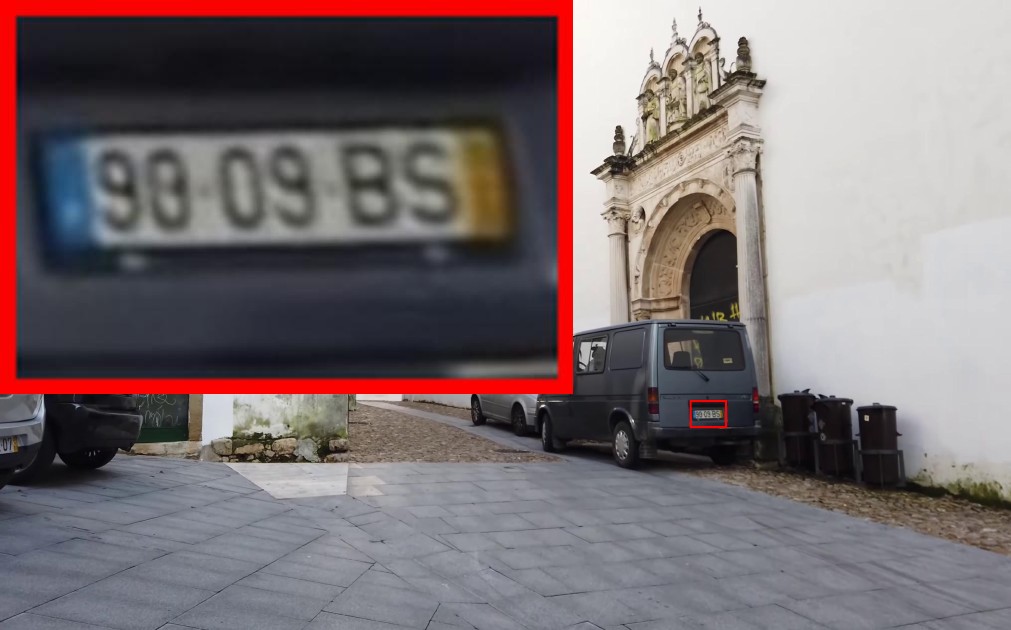}
\\[-4pt]
\parbox[t]{0.193\columnwidth}{\centering
\fontsize{6.2pt}{6pt}\selectfont
(a) Input}
&
\parbox[t]{0.193\columnwidth}{\centering
\fontsize{6.2pt}{6pt}\selectfont
(b)  w/o High-Freq\\Guidance }
&
\parbox[t]{0.193\columnwidth}{\centering
\fontsize{6.2pt}{6pt}\selectfont
(c) w/o Low-Freq\\Guidance }
&
\parbox[t]{0.193\columnwidth}{\centering
\fontsize{6.2pt}{6pt}\selectfont
(d) \textbf{Bidirectional\\Guidance (Ours)} }
&
\parbox[t]{0.193\columnwidth}{\centering
\fontsize{6.2pt}{6pt}\selectfont
(e) GT }
\end{tabular}
\endgroup
\caption{Visual comparison of the bidirectional guidance mechanism, showing the benefit of the ordered high-to-low-to-high interaction.}
\label{fig:bidirectional guidance mechanism}
\end{figure}

For the bidirectional guidance mechanism, we further evaluate the contributions of high-frequency guidance and low-frequency guidance, as reported in Table~\ref{tab:biguidance}. Removing high-frequency guidance decreases PSNR by 0.31~dB, confirming its benefit for low-frequency restoration. Removing low-frequency guidance leads to a 0.22~dB PSNR drop, demonstrating that enhanced low-frequency features effectively guide high-frequency refinement. Our method achieves the best performance with only a marginal increase in complexity. Fig.~\ref{fig:bidirectional guidance mechanism} shows the visual comparison of different cross-frequency interaction strategies, highlighting the advantage of the ordered high-to-low-to-high interaction. Overall, bidirectional guidance provides complementary benefits for cross-frequency restoration.



\section{Limitations}

\begin{figure}[t]
    \centering
    \setlength{\tabcolsep}{2pt}
    \begin{tabular}{@{}ccc@{}}
        \includegraphics[
            width=0.32\linewidth
        ]{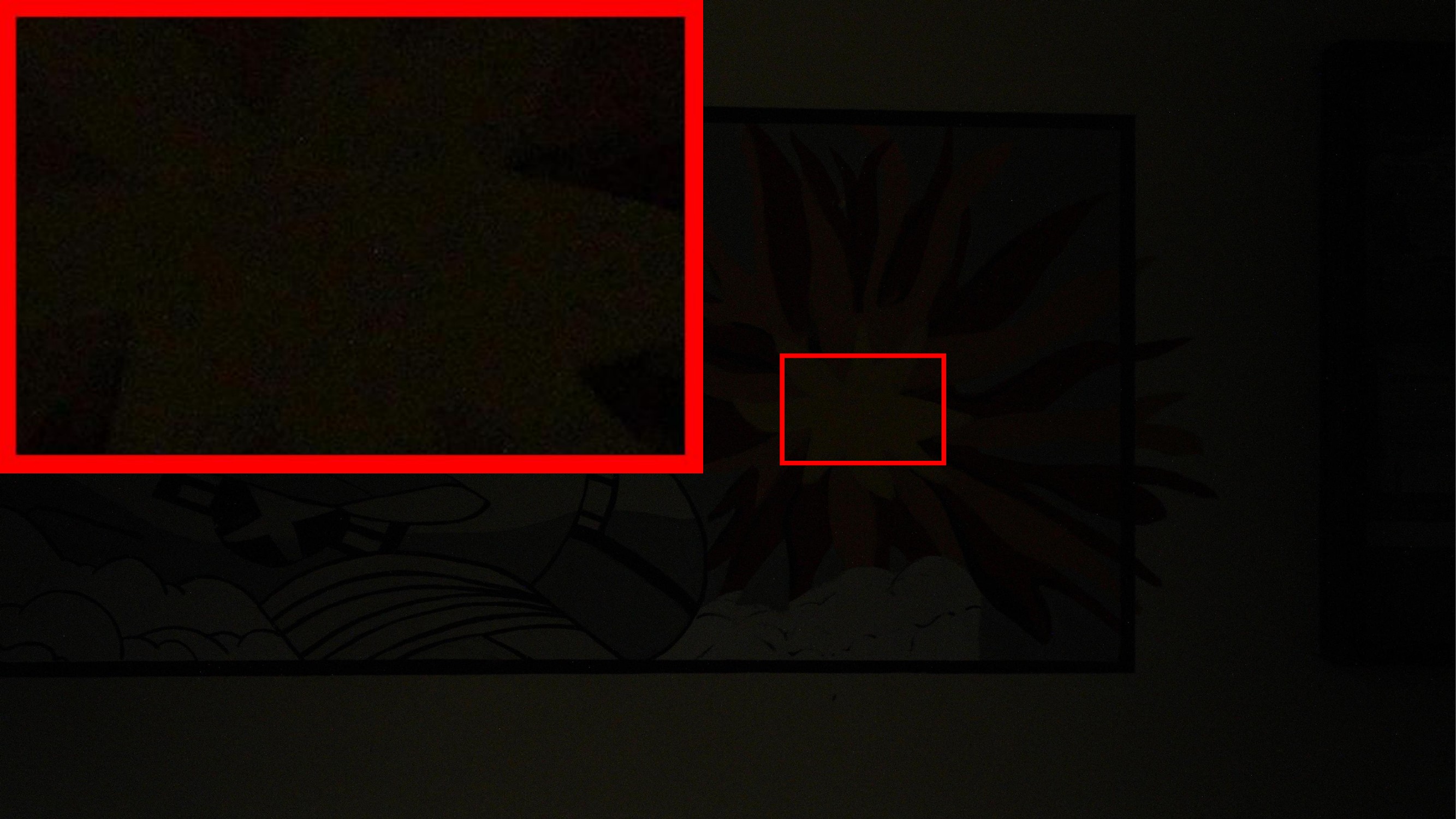} &
        \includegraphics[
            width=0.32\linewidth
        ]{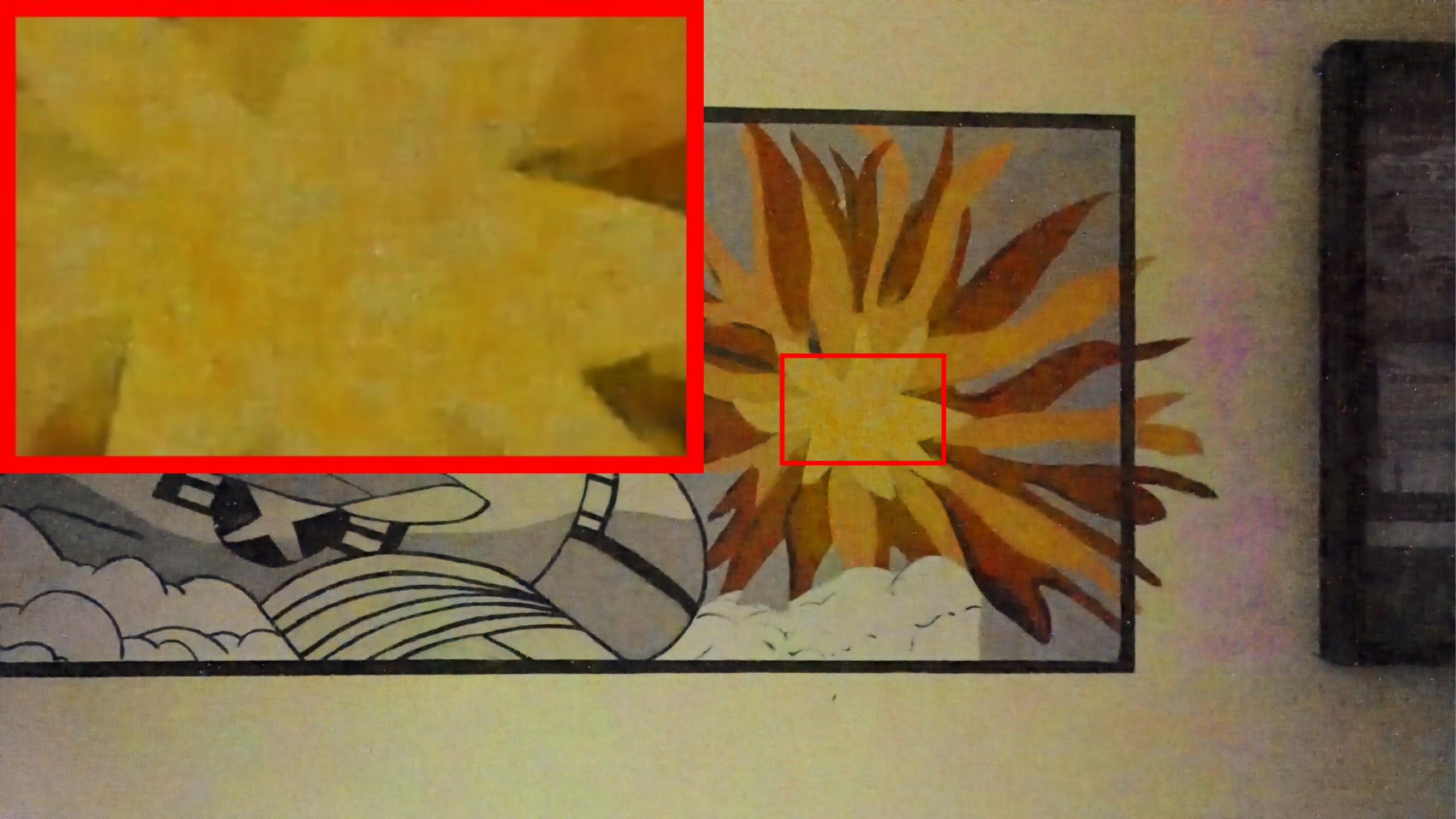} &
        \includegraphics[
            width=0.32\linewidth
        ]{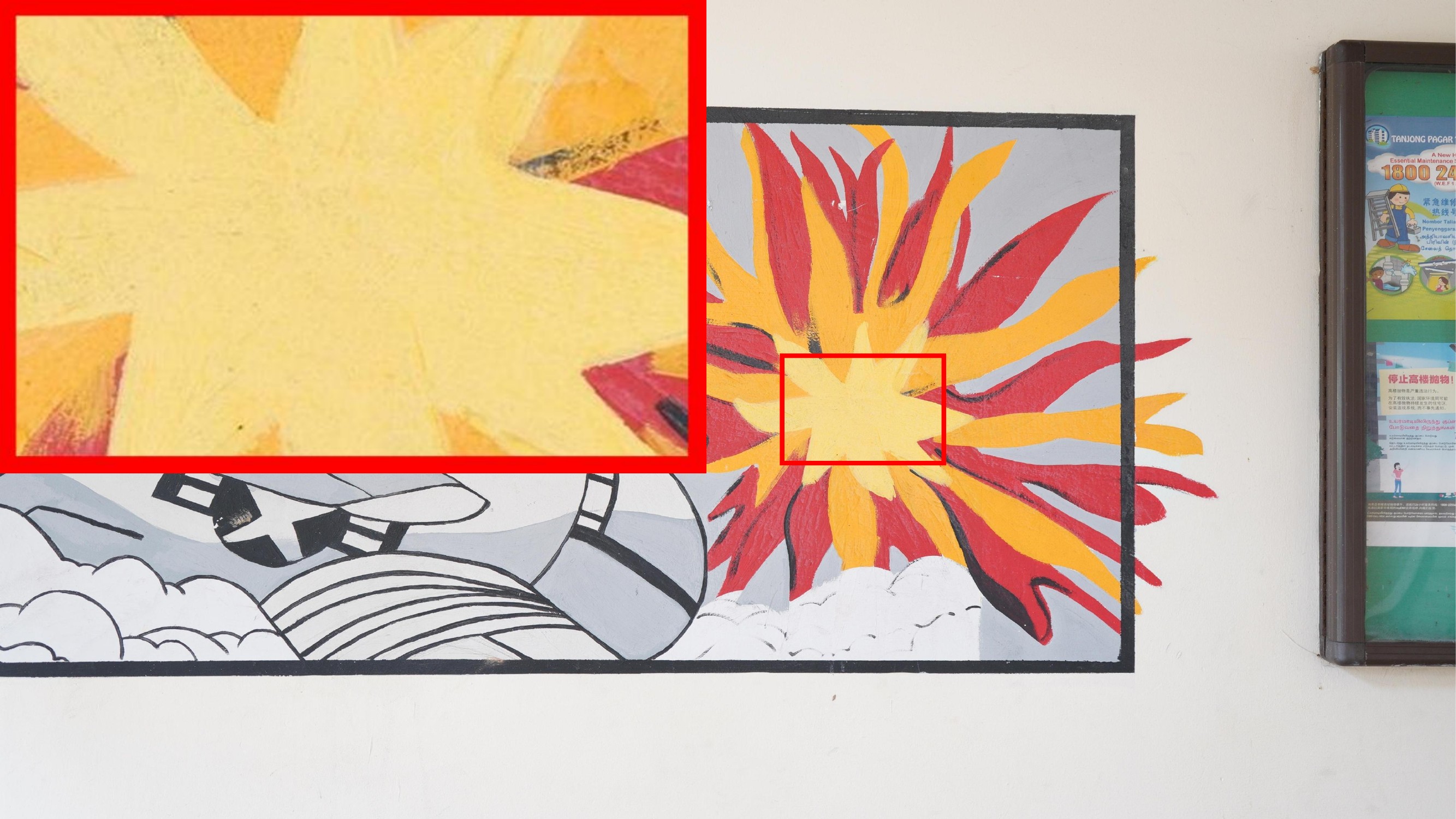}
        \\[-1mm]
        \scalebox{0.8}{(a) Input} &
        \scalebox{0.8}{(b) \textbf{BiFreqNet (Ours)}}&
        \scalebox{0.8}{(c) GT}
    \end{tabular}
    \caption{
        BiFreqNet may fail to restore the white balance and color temperature in extremely low-light scenes due to insufficient color information in low-light RGB input.
    }
    \label{fig:limitations}
    \vspace{-12pt}
\end{figure}

BiFreqNet still has limitations in extremely low-light scenes. As shown in Fig.~\ref{fig:limitations}, severe underexposure results in insufficient color information in the RGB input (a). Although BiFreqNet can recover the illumination and scene content, such insufficient color cues in the RGB input make accurate restoration of white balance and color temperature difficult, leading to a color tone shift (b). Incorporating physics-based image formation priors (\eg, camera spectral sensitivity~\cite{CSSModel} and illuminant chromaticity~\cite{KimKYK24}) into our enhancement framework may help address this limitation, which we leave for future work.

\section{Conclusion}

In this paper, we propose a Bidirectional Cross-frequency Domain Interaction Network (BiFreqNet) for UHD low-light image enhancement. Our key insight is that high-frequency structural cues can benefit low-frequency illumination restoration, while the restored low-frequency context can in turn guide high-frequency detail reconstruction. To realize this bidirectional interaction, we design a Dual-domain Feature Interaction (DFI) module consisting of a Spatial and Spectral-Aware Low-Frequency Enhancement (S$^2$LFE) block and a Cross-Frequency Enhancement (CFE) block. S$^2$LFE leverages high-frequency cues and joint spatial-spectral modeling to restore illumination and suppress noise, while CFE uses the enhanced low-frequency context to enrich structures and fine details. Extensive experiments on four general/UHD LLIE benchmarks demonstrate that BiFreqNet performs favorably against existing state-of-the-art general and UHD LLIE methods.

\bibliographystyle{IEEEtran}
\bibliography{egbib}

\end{document}